\documentclass[10pt,twocolumn,letterpaper]{article}
\usepackage{cvpr}              % To produce the CAMERA-READY version
\usepackage{graphicx}
\usepackage{amsmath}
\usepackage{amssymb}
\usepackage{booktabs}

\usepackage{color}
\usepackage{multirow}
\usepackage{bbding}
\usepackage{mathrsfs}
\usepackage{subcaption}
\usepackage{float}
\usepackage[pagebackref,breaklinks,colorlinks]{hyperref}

\usepackage[capitalize]{cleveref}
\crefname{section}{Sec.}{Secs.}
\Crefname{section}{Section}{Sections}
\Crefname{table}{Table}{Tables}
\crefname{table}{Tab.}{Tabs.}

\def\cvprPaperID{3021} % *** Enter the CVPR Paper ID here
\def\confName{CVPR}
\def\confYear{2022}

\begin{document}

\title{SCS-Co: Self-Consistent Style Contrastive Learning for Image Harmonization}

\author{
Yucheng Hang$^{1,*}$, Bin Xia$^{1,*}$, Wenming Yang$^{1,2,\dagger}$, Qingmin Liao$^{1,2}$\\
$^1$ Shenzhen International Graduate School, Tsinghua University, China\\
$^2$ Department of Electronic Engineering, Tsinghua University, China\\
{\tt\small \{hangyc20, xiab20\}@mails.tsinghua.edu.cn, \{yang.wenming, liaoqm\}@sz.tsinghua.edu.cn}
%\and
}
%\maketitle
\twocolumn[{%
	\renewcommand\twocolumn[1][]{#1}%
	\maketitle
	\begin{figure}[H]
		\vspace{-1.7em}
		\hsize=\textwidth % cvpr 需要
		\centering
		\begin{subfigure}{0.16\textwidth}
			% include first image
			\centering
			\includegraphics[width=\linewidth]{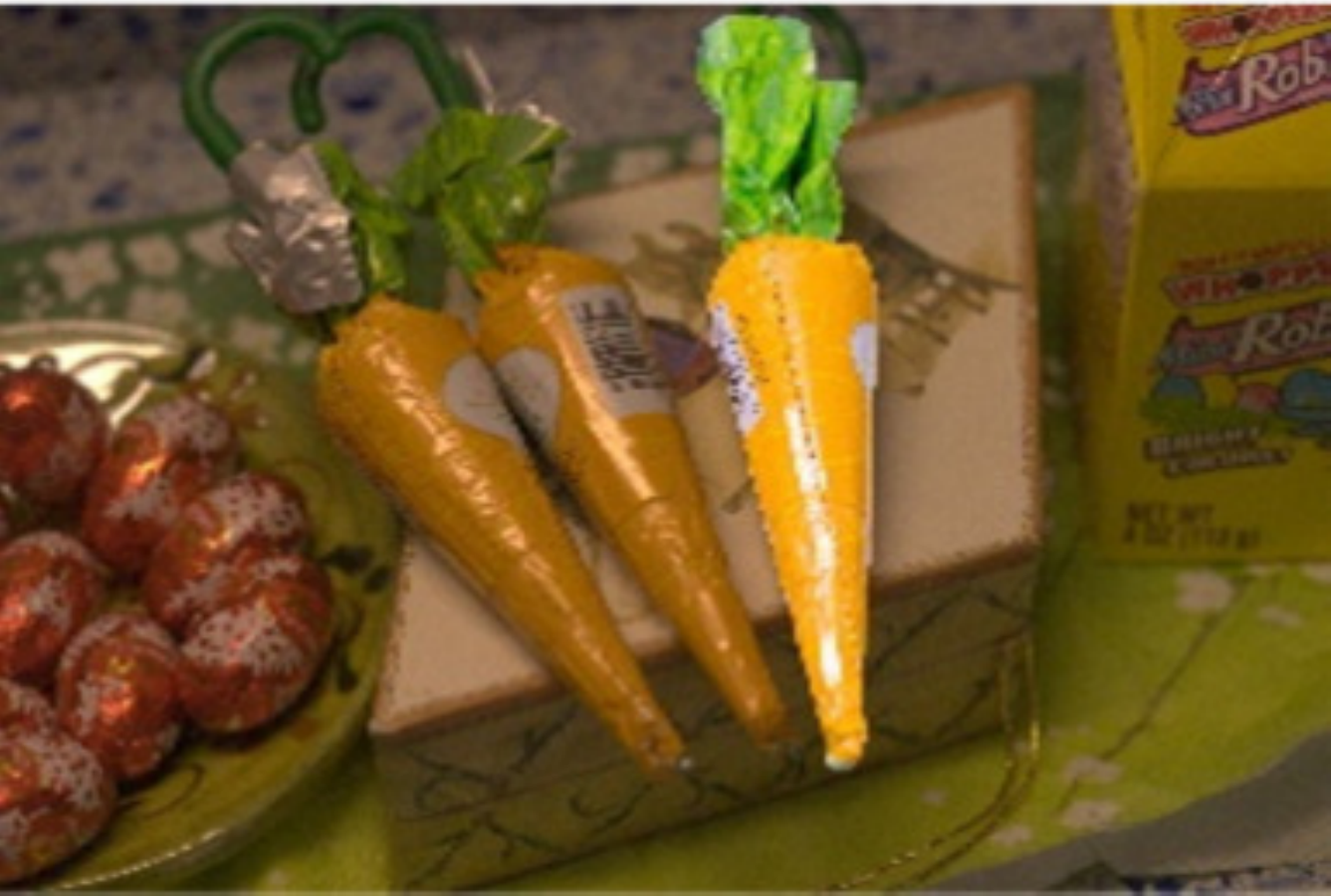}
		\end{subfigure}
		\hfil
		\begin{subfigure}{0.16\textwidth}
			% include second image
			\centering
			\includegraphics[width=\textwidth]{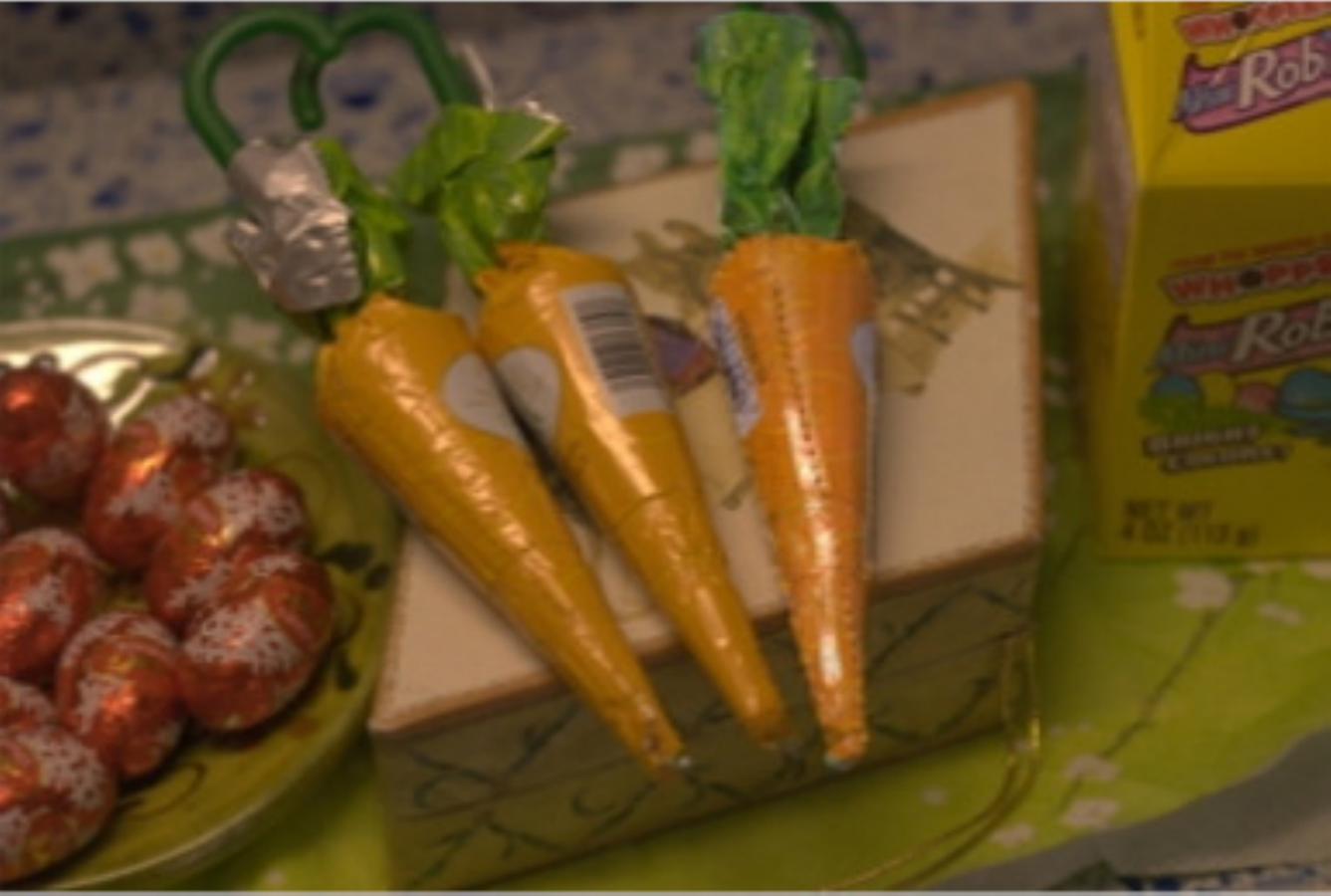}
		\end{subfigure}
		\hfil
		\begin{subfigure}{0.16\textwidth}
			% include second image
			\centering
			\includegraphics[width=\linewidth]{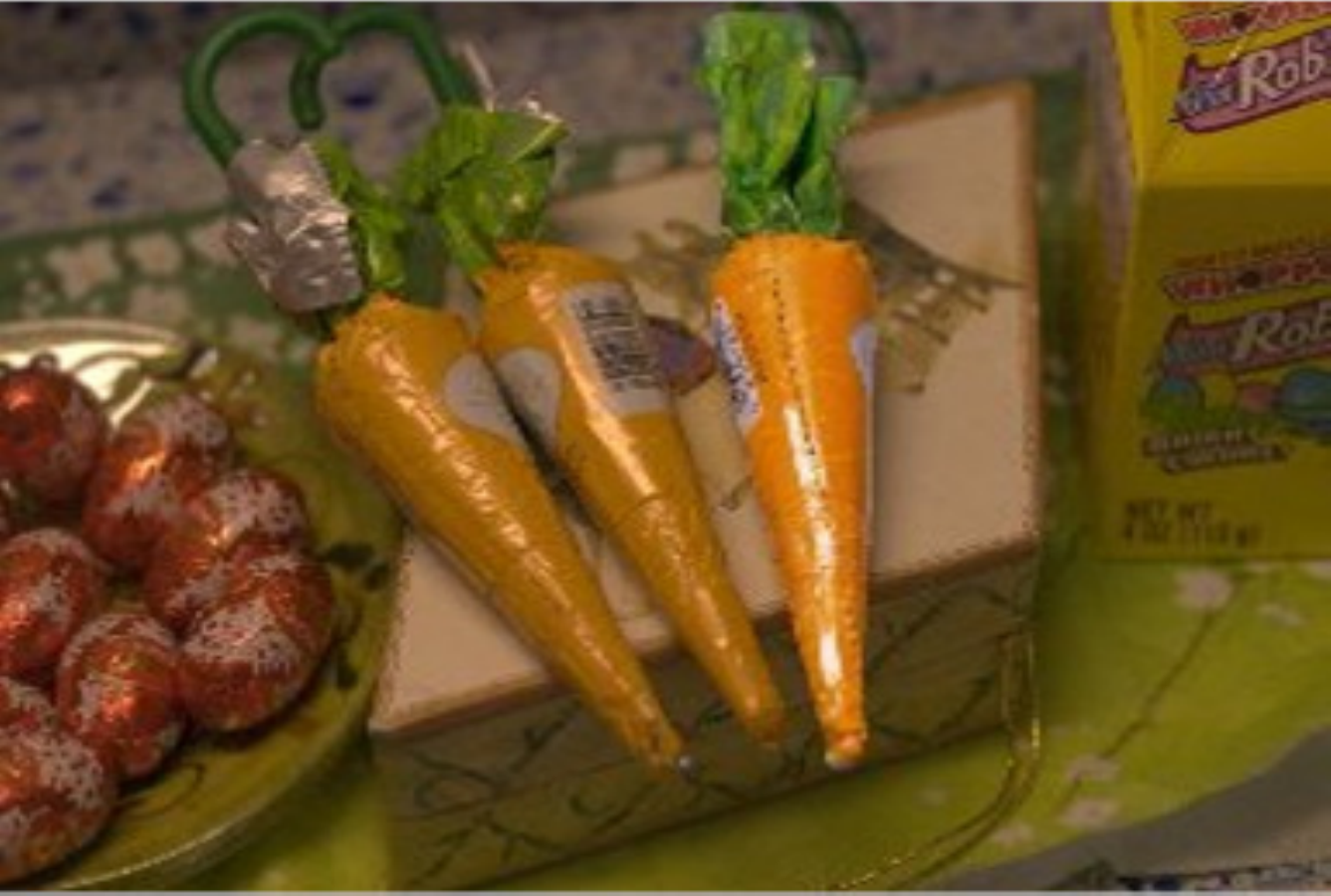}
		\end{subfigure}
		\hfil
		\begin{subfigure}{0.16\textwidth}
			% include first image
			\centering
			\includegraphics[width=\linewidth]{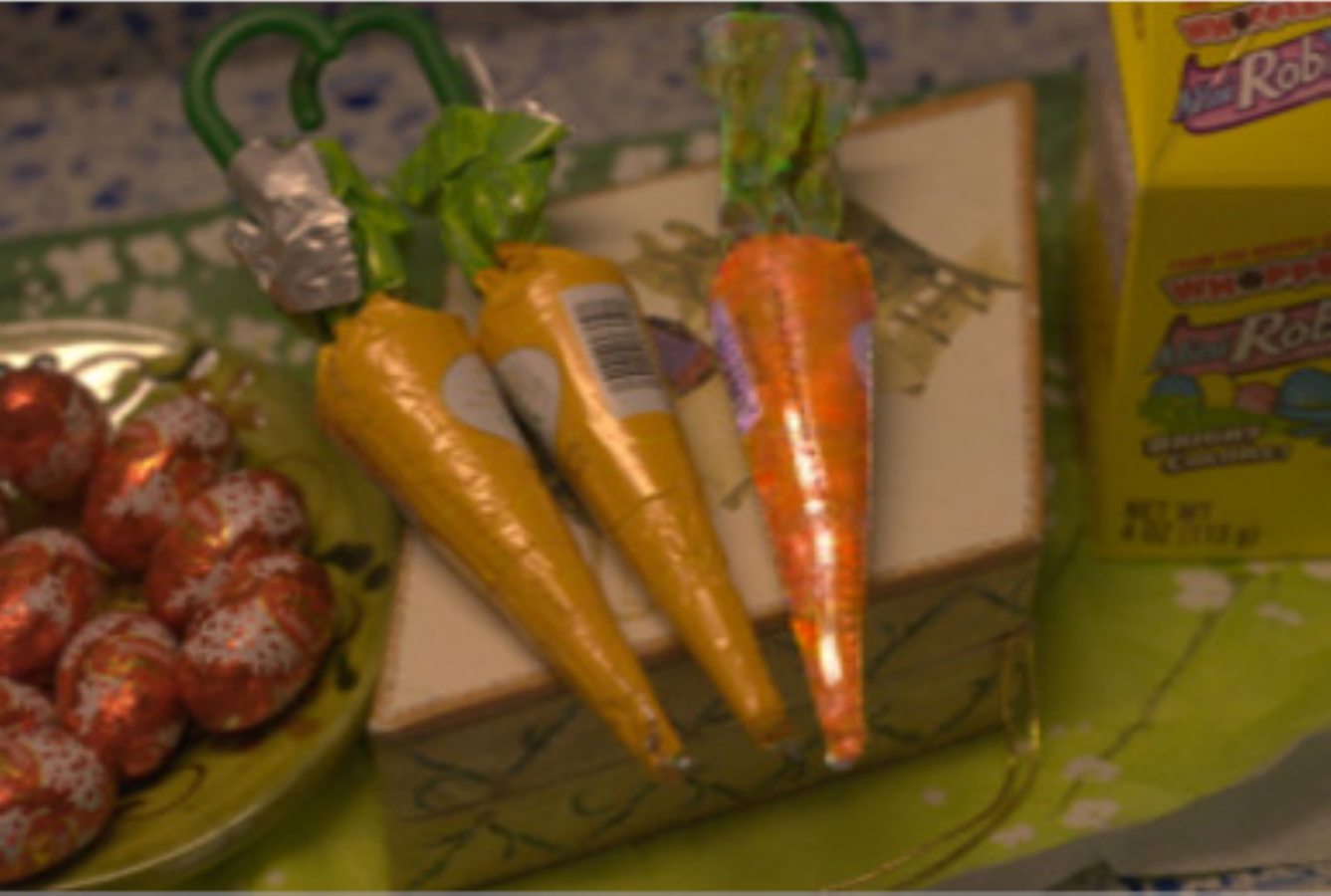}
		\end{subfigure}
		\hfil
		\begin{subfigure}{0.16\textwidth}
			% include first image
			\centering
			\includegraphics[width=\linewidth]{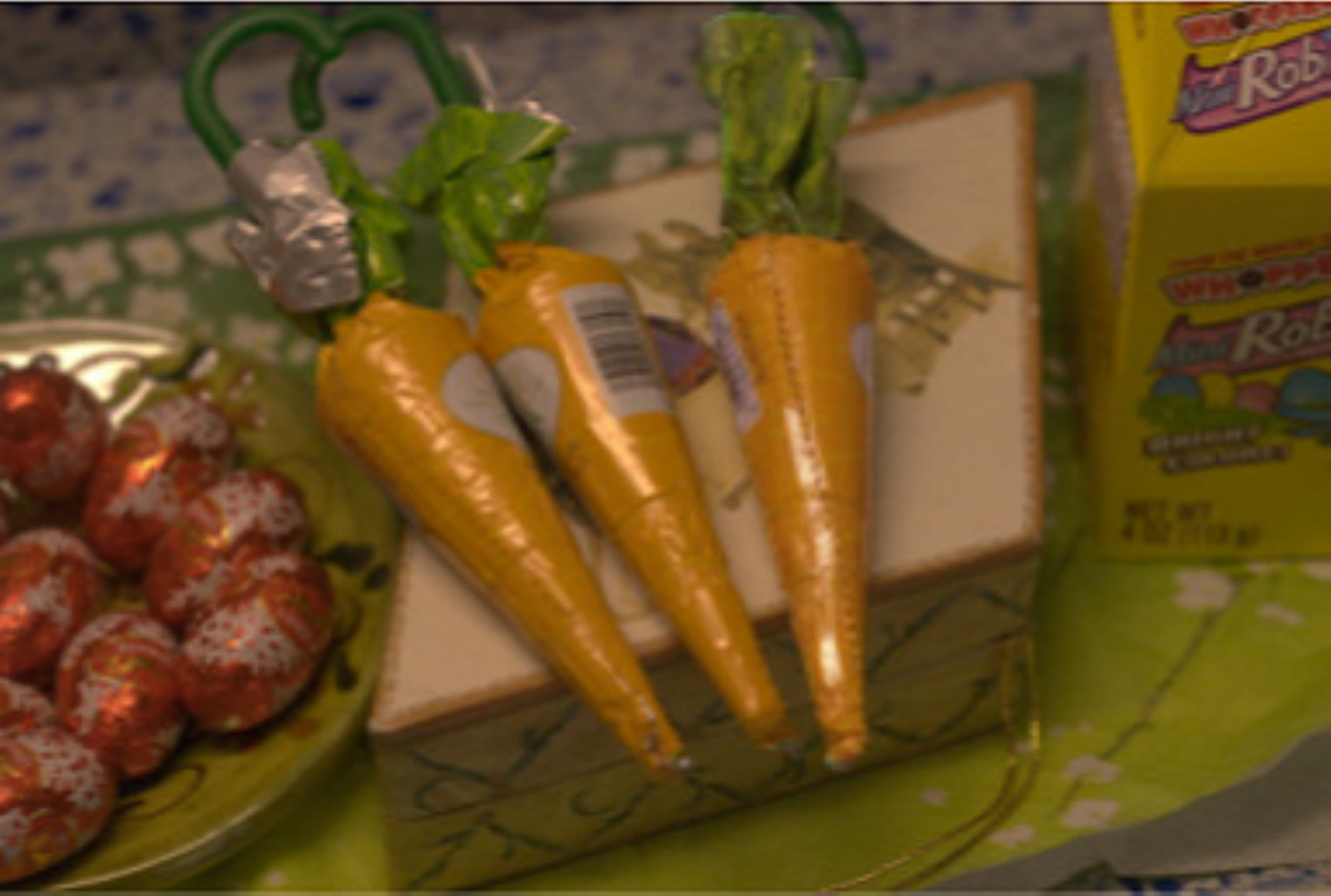}
		\end{subfigure}
		\hfil
		\begin{subfigure}{0.16\textwidth}
			% include first image
			\centering
			\includegraphics[width=\linewidth]{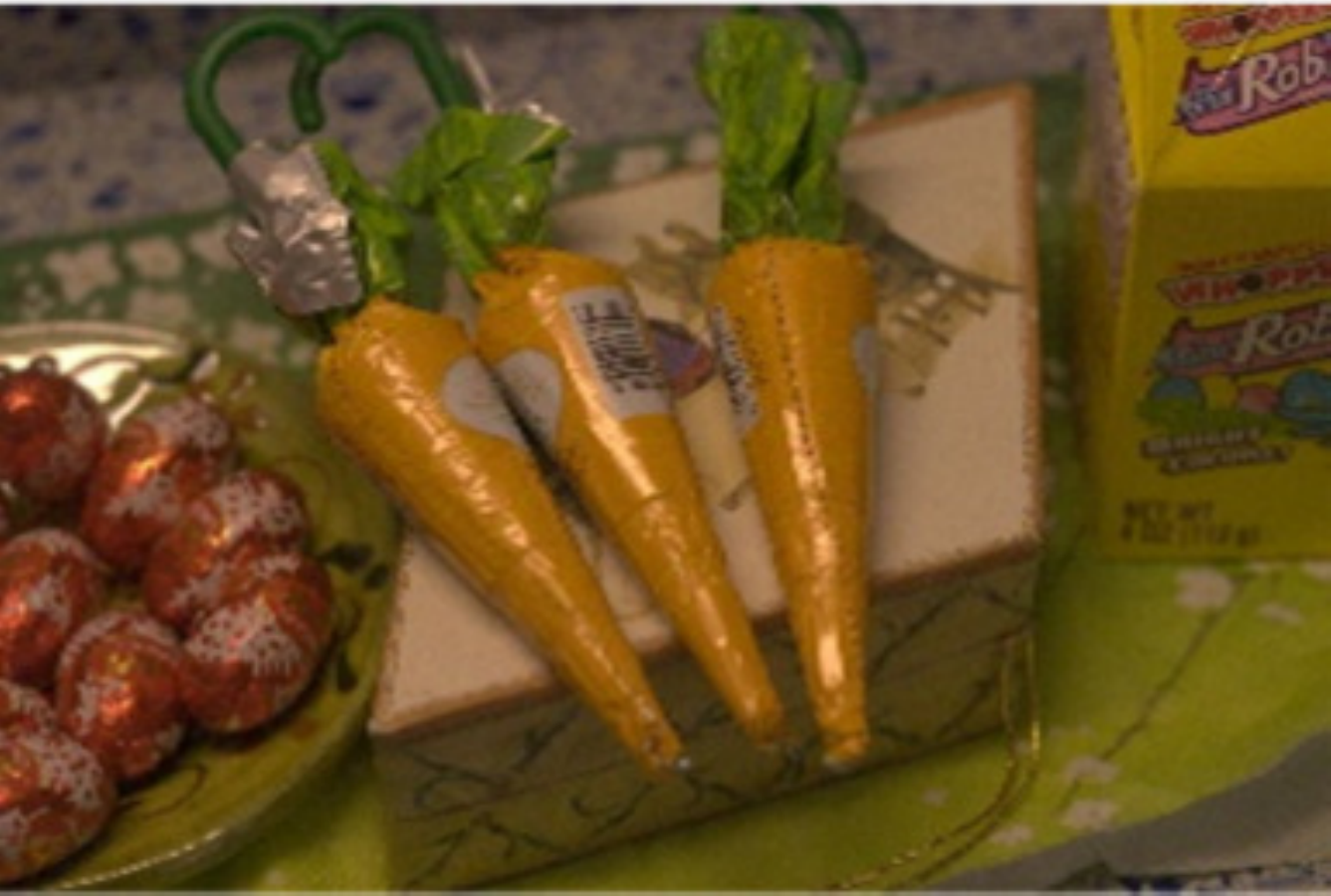}
		\end{subfigure}
		\quad
		\begin{subfigure}{0.16\textwidth}
			% include first image
			\centering
			\includegraphics[width=\linewidth]{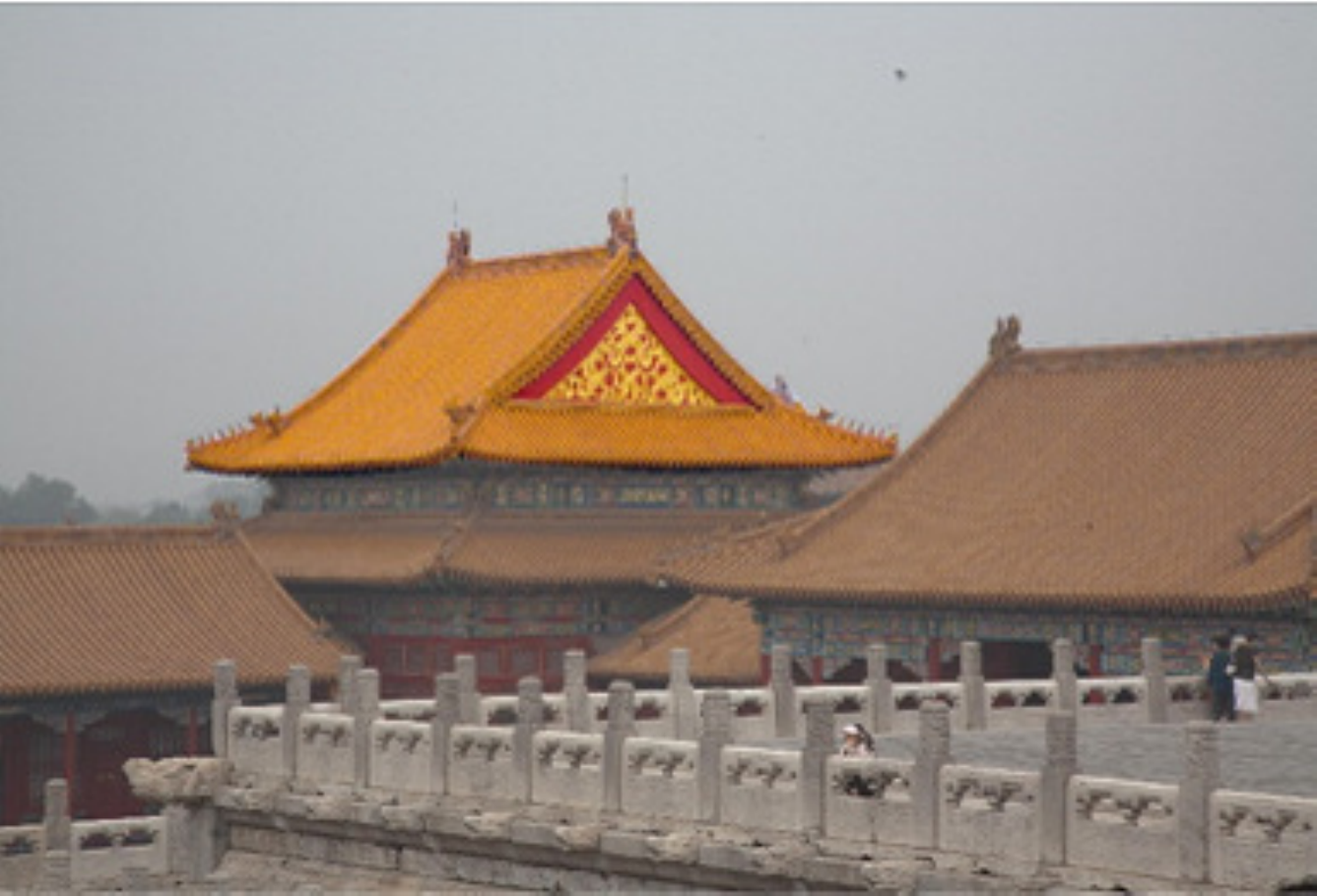}
			\caption{Composite}
		\end{subfigure}
		\hfil
		\begin{subfigure}{0.16\textwidth}
			% include second image
			\centering
			\includegraphics[width=\linewidth]{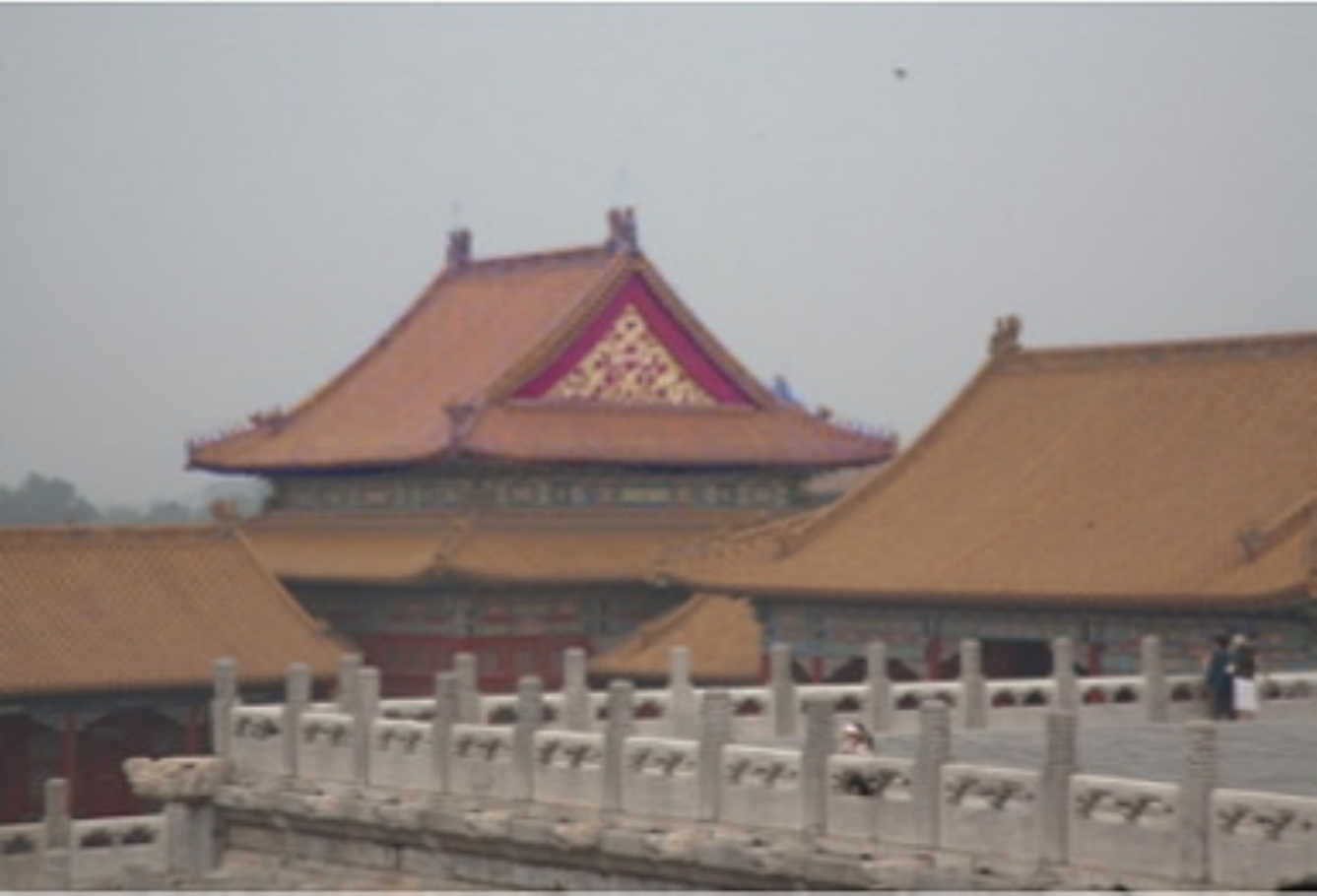}
			\caption{DoveNet \cite{cong2020dovenet}}
		\end{subfigure}
		\hfil
		\begin{subfigure}{0.16\textwidth}
			% include second image
			\centering
			\includegraphics[width=\linewidth]{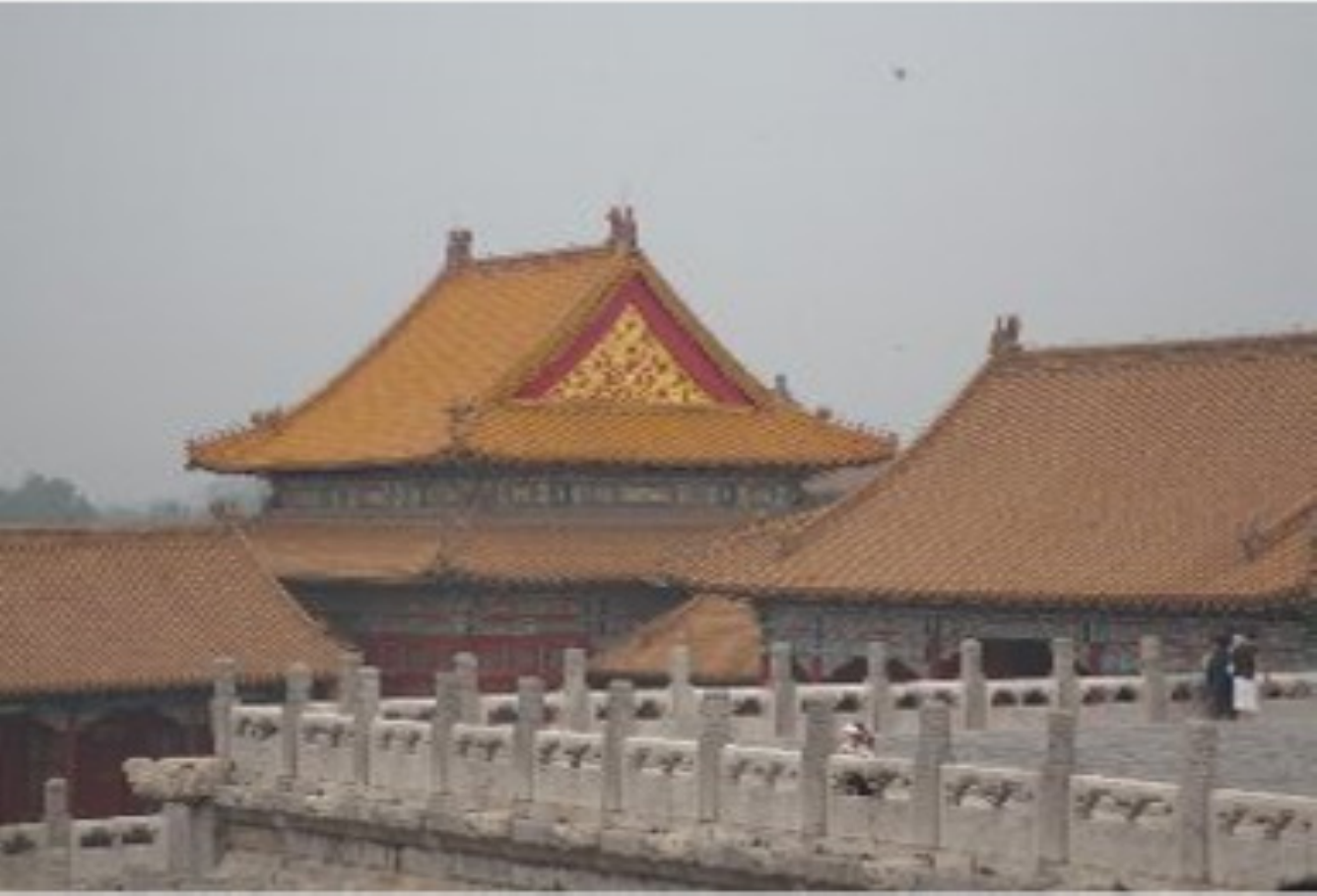}
			\caption{iS$^2$AM \cite{sofiiuk2021foreground}}
		\end{subfigure}
		\hfil
		\begin{subfigure}{0.16\textwidth}
			% include first image
			\centering
			\includegraphics[width=\linewidth]{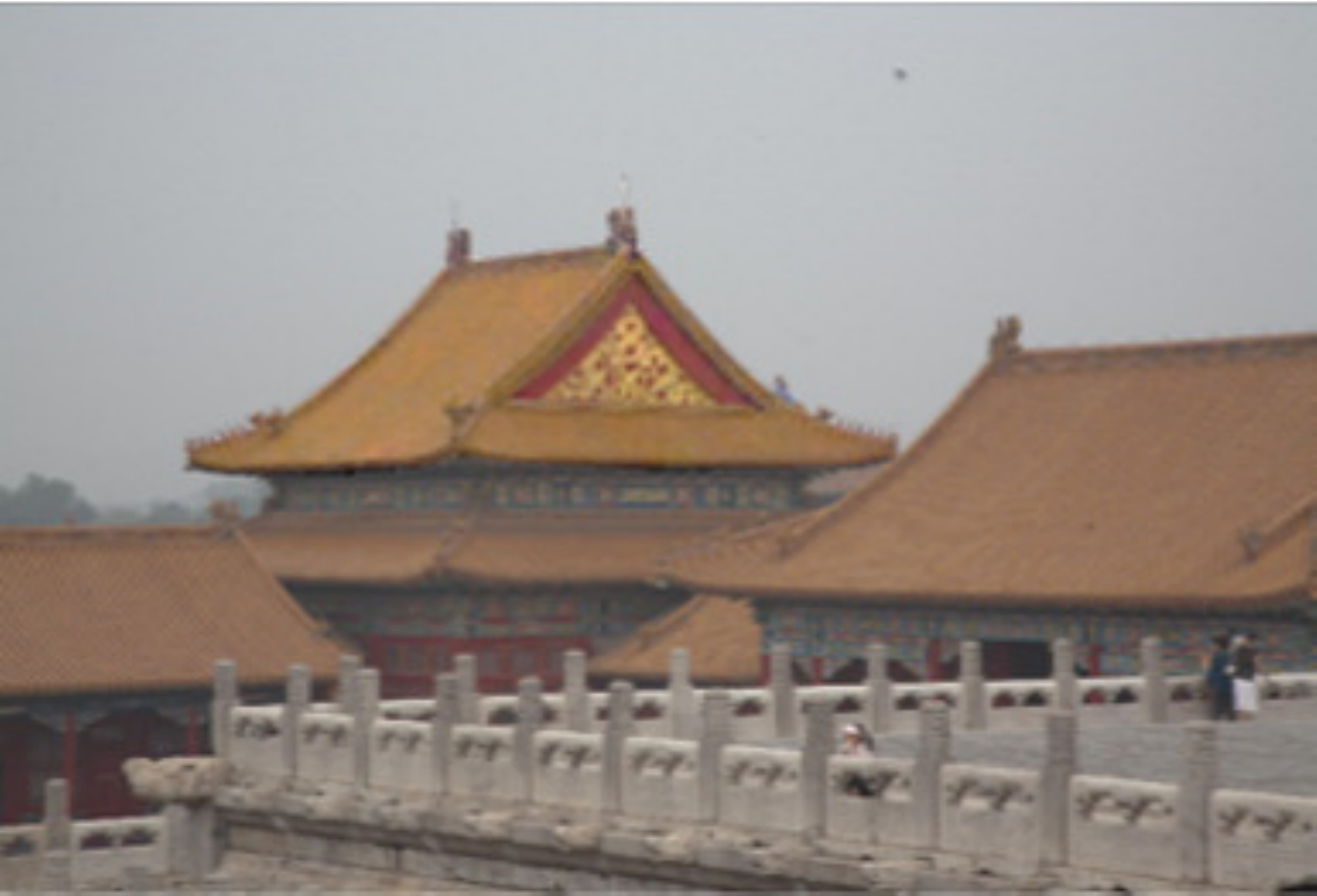}
			\caption{RainNet \cite{ling2021region}}
		\end{subfigure}
		\hfil
		\begin{subfigure}{0.16\textwidth}
			% include first image
			\centering
			\includegraphics[width=\linewidth]{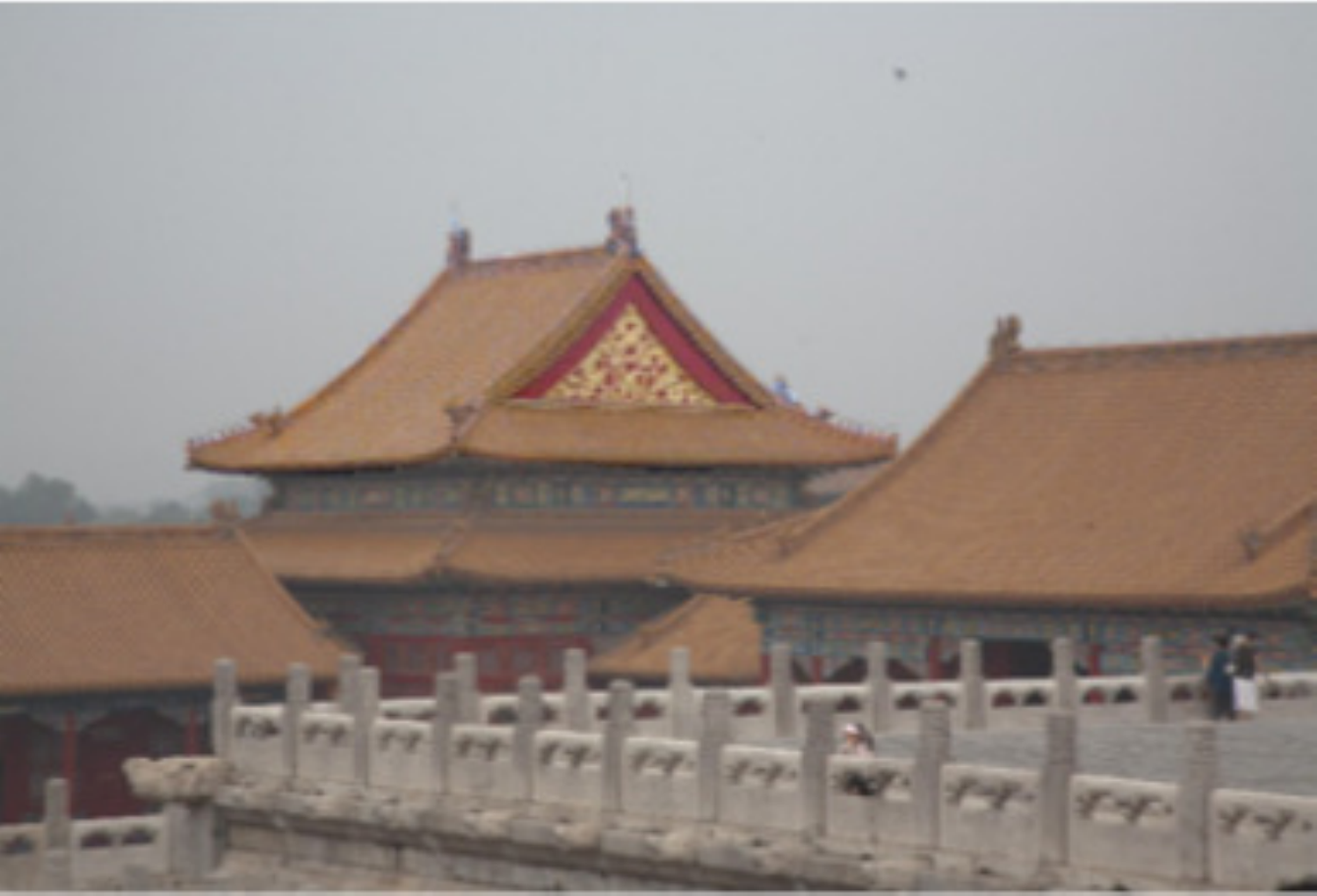}
			\caption{Ours}
		\end{subfigure}
		\hfil
		\begin{subfigure}{0.16\textwidth}
			% include first image
			\centering
			\includegraphics[width=\linewidth]{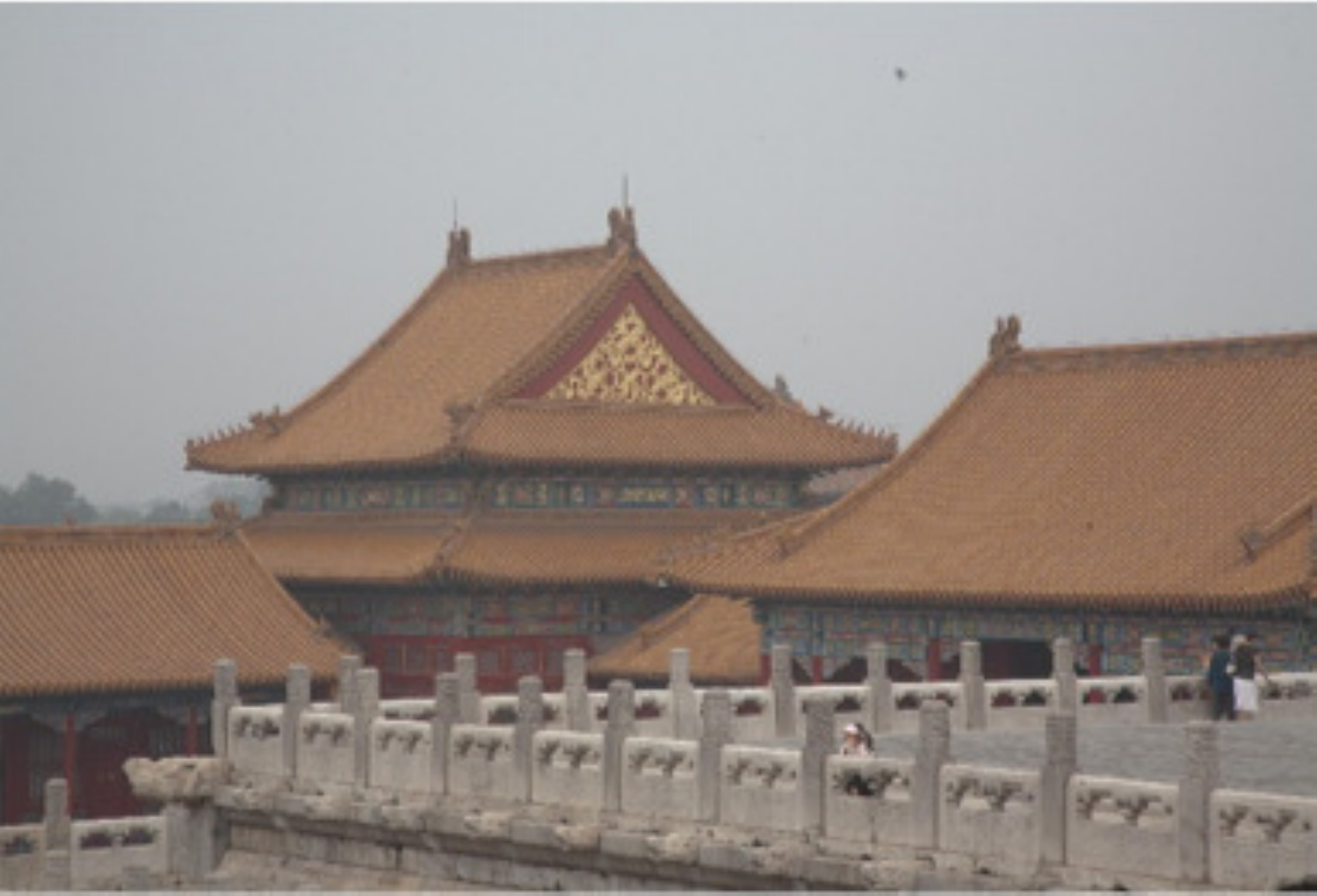}
			\caption{Real (Ground Truth)}
		\end{subfigure}
		\caption{Comparison with other methods. Thanks to the proposed SCS-Co, our method can obtain more explicit distortion knowledge from dynamically generated negative samples, and further jointly constrain the solution space from two aspects of the foreground self-style and foreground-background style consistency. Coupled with BAIN, our method produces a more photorealistic harmonized result.}
		\label{fig-intro}
	\end{figure}
	% \begin{center}
	% \centering
	% \includegraphics[scale=0.68]{first_visual-eps-converted-to}
	% \caption{foo}
	% users can easily change input control variable $\alpha_{in}$ to attain a smooth control of visual quality of restored images.}
	% \label{first_visual}
	% \end{center}%
}]
%%%%%%%%% ABSTRACT
\renewcommand{\thefootnote}{\fnsymbol{footnote}}
\footnotetext[1]{Equal contribution.}
\footnotetext[2]{Corresponding author.}
\renewcommand{\thefootnote}{\arabic{footnote}}
\begin{abstract}
\vspace{-0.5em}
   Image harmonization aims to achieve visual consistency in composite images by adapting a foreground to make it compatible with a background. However, existing methods always only use the real image as the positive sample to guide the training, and at most introduce the corresponding composite image as a single negative sample for an auxiliary constraint, which leads to limited distortion knowledge, and further causes a too large solution space, making the generated harmonized image distorted. Besides, none of them jointly constrain from the foreground self-style and foreground-background style consistency, which exacerbates this problem. Moreover, recent region-aware adaptive instance normalization achieves great success but only considers the global background feature distribution, making the aligned foreground feature distribution biased. To address these issues, we propose a self-consistent style contrastive learning scheme (SCS-Co). By dynamically generating multiple negative samples, our SCS-Co can learn more distortion knowledge and well regularize the generated harmonized image in the style representation space from two aspects of the foreground self-style and foreground-background style consistency, leading to a more photorealistic visual result. In addition, we propose a background-attentional adaptive instance normalization (BAIN) to achieve an attention-weighted background feature distribution according to the foreground-background feature similarity. Experiments demonstrate the superiority of our method over other state-of-the-art methods in both quantitative comparison and visual analysis. Code is available at \href{https://github.com/YCHang686/SCS-Co}{https://github.com/YCHang686/SCS-Co}.
\end{abstract}
%%%%%%%%% BODY TEXT
\vspace{-1.5em}
\section{Introduction}
\label{sec:intro}
Image composition is widely used in image editing \cite{xue2012understanding, cun2020improving} and data augmentation \cite{dwibedi2017cut, zhang2020learning}, which targets synthesizing a composite image by extracting the foreground of one image and pasting it on the background of another image. However, since the foreground and background appearance will be distinct due to different capture conditions, the composite image often looks unrealistic, \ie, suffers from the inharmony problem. Therefore, image harmonization, which aims to adjust the appearance of the foreground to make it compatible with the background in the composite image, is significant and challenging.

Numerous deep learning-based methods have been proposed for image harmonization. However, most methods \cite{guo2021intrinsic,tsai2017deep,cun2020improving,zhu2015learning,Guo_2021_ICCV,sofiiuk2021foreground} do not consider this problem from the perspective of visual style. Hence, they fail to ensure a visual style consistency between the foreground and the background \cite{ling2021region}. Methods based on domain translation \cite{cong2020dovenet,cong2021bargainnet} implicitly consider this problem from the perspective of domain-consistency, but do not directly transform the foreground feature in the generator.

Recently, Ling \etal~\cite{ling2021region} explicitly introduce the concept of visual style and first regard image harmonization as a background-to-foreground style transfer problem\footnote{In fact, for a similar task, namely painterly harmonization, Luan \etal~\cite{luan2018deep} introduce the concept of visual style earlier.}. Inspired by AdaIN~\cite{huang2017arbitrary}, they propose a region-aware adaptive instance normalization (RAIN) for image harmonization and achieve great success. However, as shown in Figure \ref{fig-intro}(d), the distortion still exists or even is very severe in some cases.

We argue that two issues lead to the above dilemma: (1) Just like the problem with AdaIN, RAIN only considers the global style distribution in the background and aligns the foreground feature distribution with it. However, as a common intuition, areas in the background that feature-similar to the foreground need more attention. For example, in the first row of Figure \ref{fig-intro}, the foreground object reappears twice in the background. The model should pay more attention to the local style distributions of these two areas. (2) The second is a general issue, not limited to the style-based method, and is the core issue we want to solve. Most existing methods \cite{guo2021intrinsic,tsai2017deep,cun2020improving,zhu2015learning,Guo_2021_ICCV,sofiiuk2021foreground} only use real images to guide the training via an $\mathcal{L}_{1}$ loss, which is too simple and cannot constrain the solution space well \cite{wu2021contrastive}. Toward this end, DoveNet \cite{cong2020dovenet} and RainNet \cite{ling2021region} adopt a domain verification loss. However, it only regards the foreground-background feature similarity of the real/harmonized image as positive/negative, and the input composite image is not used, which contains important distortion knowledge. In other words, it is just a positive-orient constraint. In addition, since image harmonization aims to adjust the foreground, why not directly constrain the foreground feature? Considering the above problems, Cong \etal propose a triplet loss \cite{cong2021bargainnet}. However, it directly pulls the foreground domain code to the background domain code, which is too strong and will be interfered by content information. One more important problem is that only using the input composite image as the negative sample, leading to limited external distortion knowledge \cite{wu2021contrastive,wang2021towards}, and the learned feature distribution easily becomes biased \cite{liu2021divco,wang2021towards}. \emph{In summary, why not dynamically generate multiple negative samples and jointly constrain from the foreground self-style and foreground-background style consistency to obtain more distortion knowledge and reduce the solution space?}
\begin{figure}[htbp]
	\centering
	\includegraphics[width=\linewidth]{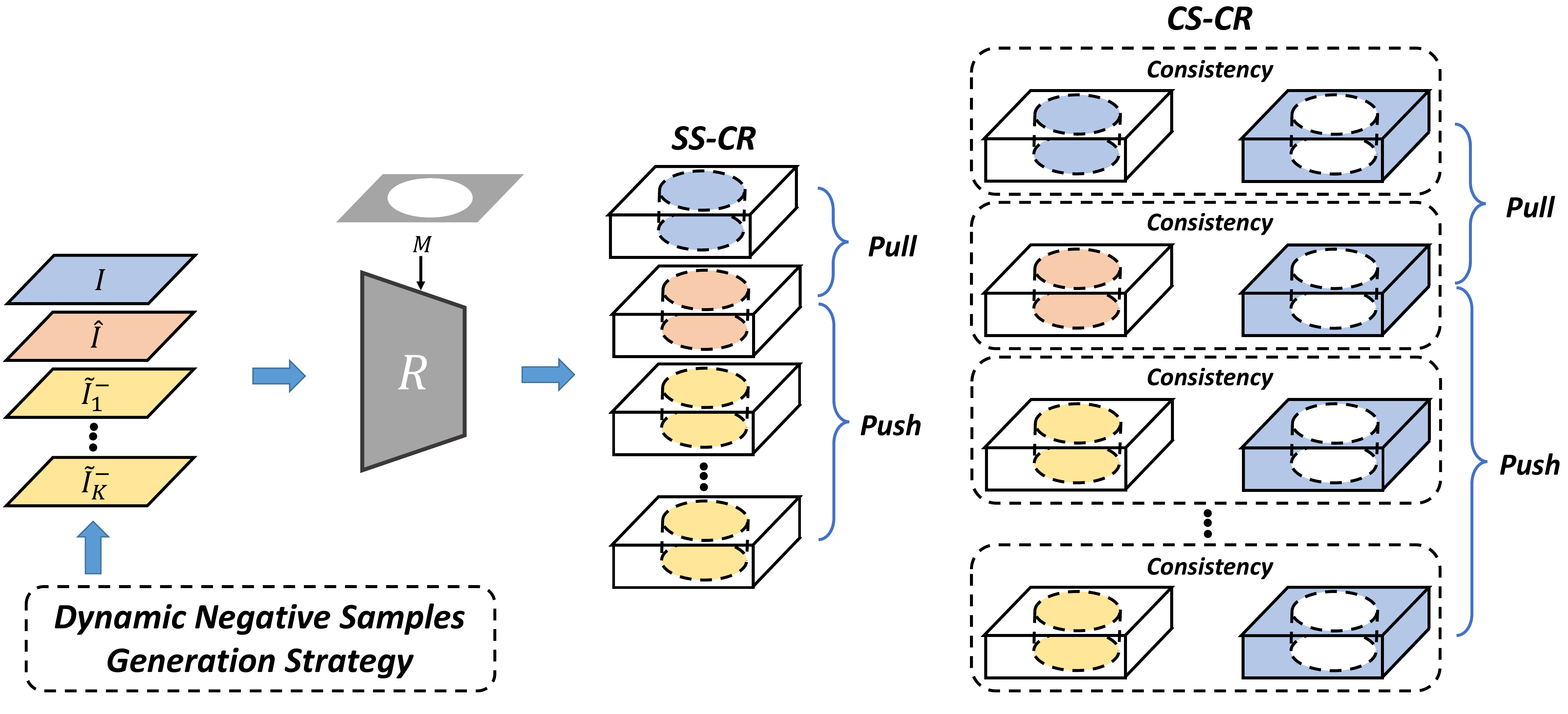} % Reduce the figure size so that it is slightly narrower than the column.
	\caption{The illustration of our SCS-Co, including SCS-CR and the dynamic negative samples generation strategy. The detail of this strategy is shown in Figure \ref{fig:online-strategy}.}
	\label{fig:scs-cr}
\vspace{-1.0em}
\end{figure}

Motivated by the observations and analyses above, we try to address these two issues. For the first issue, inspired by \cite{Liu_2021_ICCV,park2019arbitrary}, we propose a Background-attentional Adaptive Instance Normalization (BAIN). It can learn the feature similarity between the foreground and background, and calculate an attention-weighted style distribution of the background according to this feature similarity. Finally, the foreground feature distribution is aligned with this distribution.

For the second issue, we attempt to solve it by considering the positive and negative relations simultaneously in the form of contrastive learning. Specifically, we propose a novel Self-Consistent Style Contrastive Learning Scheme (SCS-Co) (see Figure \ref{fig:scs-cr}), including a Self-Consistent Style Contrastive Regularization (SCS-CR) and a Dynamic Negative Samples Generation Strategy (see Figure \ref{fig:online-strategy}). For a composite image $\tilde{I}$, we denote its corresponding harmonized image $\hat{I}$ and its ground truth real image $I$ as the anchor and positive sample, respectively. We also denote this composite image $\tilde{I}$ as the first negative sample $\tilde{I}_{1}^{-}$. More negative samples with the same content but different distortions are achieved via our dynamic negative samples generation strategy. Then we try to pull the anchor sample closer to the positive sample and push the anchor sample away from negative samples in the style representation space. In detail, for more powerful constraint, we not only constrain from the foreground self-style representation, but also use the background style representation as guidance to constrain from the foreground-background style consistency.

Our contributions are summarized as three-fold:
\begin{itemize}
	\item For the first time, we introduce contrastive learning to image harmonization. Our self-consistent style contrastive learning scheme (SCS-Co) can further improve the performance of existing image harmonization networks without any increase in model parameters.
	\item We develop a background-attentional adaptive instance normalization (BAIN). It learns a foreground-background feature similarity attention map and properly normalizes the foreground feature by the per-point attention-weighted background feature statistics.
	\item Extensive experiments prove that our method is powerful for image harmonization. Compared with other state-of-the-art methods, our method obtains superior results in both quantitative metrics and visual quality.
\end{itemize}
%-------------------------------------------------------------------------
\begin{figure*}[htbp]
	\centering
	\includegraphics[width=0.9\textwidth]{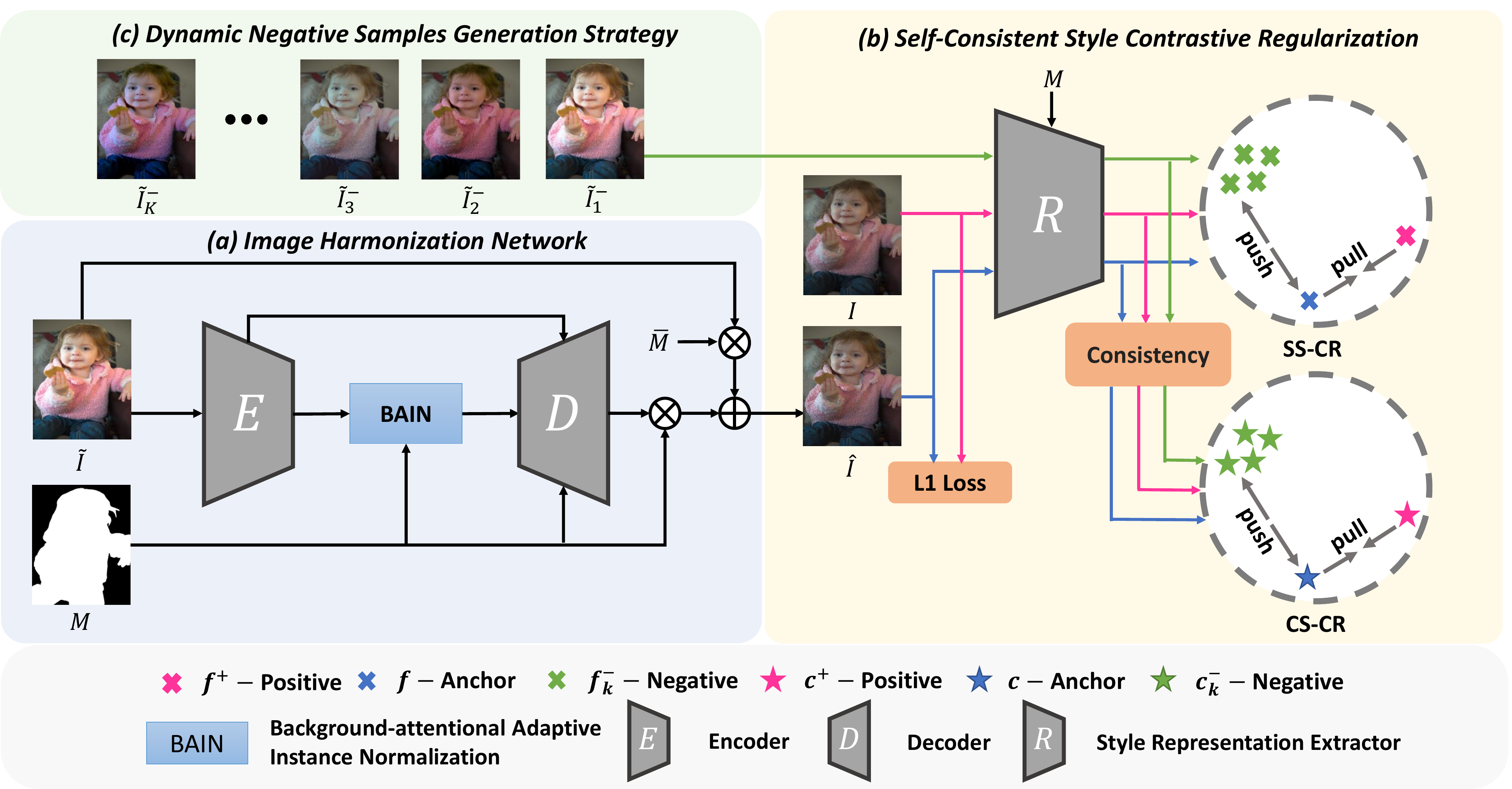} % Reduce the figure size so that it is slightly narrower than the column.
	\caption{The architecture of our method, which consists of (a) image harmonization network, (b) self-consistent style contrastive regularization and (c) dynamic negative samples generation strategy. Please note that the first negative sample $\tilde{I}_{1}^{-}$ is the input composite image $\tilde{I}$. (b) and (c) make up our self-consistent style contrastive learning scheme.}
	\label{fig3}
\vspace{-1.0em}
\end{figure*}
\section{Related Work}
\vspace{-0.2em}
\paragraph{Image Harmonization.}
Various approaches have been proposed for image harmonization. Traditional methods focus on better transferring hand-crafted low-level appearance statistics, such as color statistics \cite{pitie2005n, reinhard2001color, xue2012understanding}, gradient information \cite{perez2003poisson, jia2006drag, tao2010error}, multi-scale various statistics~\cite{sunkavalli2010multi} between foreground and background regions. However, they could not address complex cases where the foreground image has a large appearance gap with the background image. With the advances of deep learning, more deep learning-based methods were proposed. To learn the differences between various low-level features in the composite images, Cun and Pun~\cite{cun2020improving} design an additional spatial-separated attention module. In~\cite{tsai2017deep}, they present an end-to-end CNN network for image harmonization and incorporate an auxiliary segmentation branch to use semantic information. Guo \etal~\cite{guo2021intrinsic} first model image harmonization based on intrinsic image theory and adopt an autoencoder to disentangle composite image into reflectance and illumination for separate harmonization. In~\cite{sofiiuk2021foreground}, they combine pre-trained semantic segmentation models with encoder-decoder architectures for image harmonization. With the rise of Transformer, Guo \etal~\cite{Guo_2021_ICCV} design the first harmonization Transformer frameworks without and with disentangled representation. In~\cite{Jiang_2021_ICCV}, they propose the first self-supervised harmonization framework that needs neither human-annotated masks nor
professionally created images for training.
\vspace{-1.0em}
\paragraph{Arbitrary Style Transfer.}
Arbitrary style transfer is a technique used to render a photo with a particular visual style by synthesizing global and local style patterns from a given style image evenly over a content image while maintaining its original structure. Originating from non-realistic rendering \cite{kyprianidis2012state}, earlier image style transfer methods are closely related to texture synthesis \cite{efros1999texture, elad2017style, gatys2015texture}. Adopting the success of deep learning, Gatys \etal first formulate style transfer as the matching of multi-level deep features extracted from a pre-trained deep neural network and achieve surprising performance \cite{gatys2016image}. Huang \etal create a novel way for real-time style transfer by matching the mean-variance statistics between content and style features (AdaIN) \cite{huang2017arbitrary}. Afterwards, many methods are proposed \cite{gu2018arbitrary,zhang2019multimodal,an2021artflow,Liu_2021_ICCV,Wu_2021_ICCV}. However, as stressed in \cite{ling2021region}, these style transfer methods are not practical for our task because the style defined in our work is consistent with image realism instead of texture, and our task is region-aware, which otherwise will introduce new problems of feature shift.
\vspace{-1.0em}
\paragraph{Contrastive Learning.}
Contrastive learning has demonstrated its effectiveness in self-supervised representation learning \cite{he2020momentum, sermanet2018time,chen2020simple,henaff2020data,tian2020contrastive,wu2018unsupervised,oord2018representation}. Instead of using a pre-defined and fixed target, contrastive learning aims to pull positive samples close to the anchor and push negative samples away in a representation space, increasing mutual information. However, different from high-level vision tasks \cite{he2020momentum, grill2020bootstrap, chen2020simple, henaff2020data}, which inherently suit for modeling the contrast between positive and negative samples, there are still few works applying contrastive learning to low-level vision tasks due to their difficulty in constructing negative samples and contrastive loss \cite{wu2021contrastive,wang2021unsupervised,wang2021towards}. In this paper, specifically for image harmonization, we design a self-consistent style contrastive learning scheme.
\section{Our Method}
\subsection{Problem Formulation}
Given a foreground image ${I}_{f}$ and a background image $I_{b}$, the object composition process of the composition image can be formulated as  $\tilde{I}=M \cdot {I}_{f}+(1-M) \cdot I_{b}$, where $\cdot$ is element-wise multiplication, $M$ is the foreground mask, which indicates the region to be harmonized, and therefore the background mask is $\bar{M}=1-M$. Our goal is to learn a harmonization network $G$, whose output is the harmonized image as $\hat{I}=G(\tilde{I}, M)$ and should be close to the ground truth real image $I$ by ${\mathcal{L}}_{rec}=\|I-\tilde{I}\|_{1}$.

\subsection{Image Harmonization Network}
As shown in Figure \ref{fig3}(a), our network $G$ is based on U-Net with skip links from the encoder to the decoder. Their details can be found in supplementary. In addition, we propose a background-attentional adaptive instance normalization (BAIN) inserted between the encoder and decoder, which will be explained in detail in Section \ref{sec:BAIN}. 

\subsection{Self-Consisitent Style Contrastive Learning Scheme (SCS-Co)}
\begin{figure}[htbp]
	\centering
	\includegraphics[width=\linewidth]{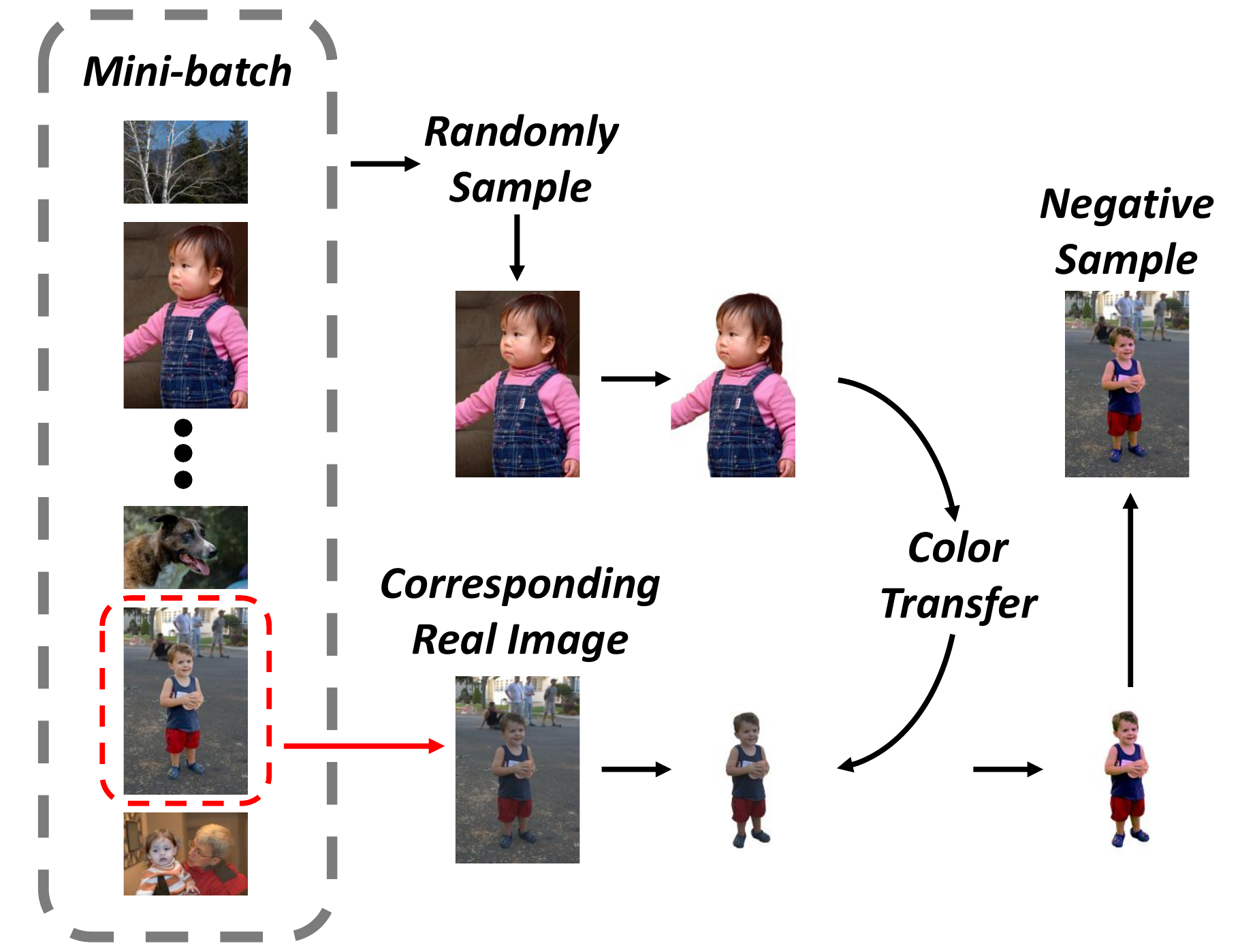} % Reduce the figure size so that it is slightly narrower than the column.
	\caption{The illustration of our dynamic negative samples generation strategy. Red box indicates the input composite image. Through this strategy, more negative samples with the same content but different distortions are obtained, which provide much distortion knowledge.}
	\label{fig:online-strategy}
	\vspace{-1.0em}
\end{figure}

As shown in Figure \ref{fig:scs-cr}, our SCS-Co contains SCS-CR and the dynamic negative samples generation strategy.
In detail, SCS-CR consists of self-style contrastive regularization (SS-CR) and consistent style contrastive regularization (CS-CR). We make it clear that self-style refers to the style of the foreground, and consistent style refers to the foreground-background style consistency.
\vspace{-1.0em}
\paragraph{Formulation.}
For our SCS-Co, we need to resolve two key issues. One is to construct positive and negative samples. In our SCS-Co, we choose the harmonized image $\hat{I}$ generated by the image harmonization network $G$ and the corresponding real image $I$ as the anchor and the positive sample, respectively. The most important task is to construct negative samples. We can simply use the input composite image $\tilde{I}$ as the only negative sample. However, as emphasized in existing contrastive learning methods \cite{he2020momentum,chen2020simple}, a large dictionary covering a rich set of negative samples is critical for good representation learning. Therefore, during the training process, for each input composite image $\tilde{I}$, we generate $K$ negative samples online. Specifically, we propose a dynamic negative samples generation strategy. As shown in Figure \ref{fig:online-strategy}, given an input composite image $\tilde{I}$ (\textcolor[rgb]{ 1,  0,  0}{Red box}), we use it as the first negative sample $\tilde{I}_{1}^{-}$. 
Then we get its corresponding real image $I$ and segment the foreground region $R_{f}$ according to the foreground mask $M$. Afterwards, we sample $K-1$ images (other than $\tilde{I}_{1}^{-}$) from the same mini-batch and segment their foreground regions. As suggested in \cite{tsai2017deep,cong2020dovenet}, we transfer the color information of these $K-1$ foreground regions to $R_{f}$ respectively, leading to $K-1$ negative samples. Finally, we successfully obtain $K$ negative samples, \ie, $\tilde{I}_{k}^{-}, k=1,2,3,...,K$.

The other is to find the style representation space of these samples for contrast. We use a fixed pre-trained style representation extractor $R$ and introduce the foreground mask $M$ and the background mask $\bar{M}=1-M$ to obtain style representations for different regions. Specifically, we input $\hat{I}$, $I$ and $\tilde{I}_{k}^{-}, k=1,2,3,...,K$. Then we can obtain the anchor foreground style representation $f=R(\hat{I}, M)$, the positive foreground style representation ${f}^{+}=R(I, M)$, the positive background style representation ${b}^{+}=R(I, \bar{M})$, and negative foreground style representations ${f}^{-}_{k}=R(\tilde{I}_{1}^{-}, M), k=1,2,3,...,K$.

Thus, SS-CR can be formulated as:
\begin{equation}
\label{equ:ss-cr}
{\mathcal{L}}_{ss-cr}= \frac{D\left(f, f^{+}\right)}{D\left(f, f^{+}\right)+\sum_{k=1}^{K} D\left(f, f_{k}^{-}\right)},
\end{equation}
where $D(x, y)=\|x-y\|_{1}$ denotes the $\mathcal{L}_{1}$ distance between $x$ and $y$. As shown in Figure \ref{fig:scs-cr}, our SS-CR focuses on the foreground self-style, pulling $f$ closer to ${f}^{+}$ and pushing $f$ away from $\{f_{k}^{-}\}_{k=1}^{K}$.

However, so far we have not used the background style representation, which is a powerful guidance for image harmonization \cite{cong2021bargainnet}. Therefore, we further make contrastive constraints from the perspective of foreground-background style consistency. Specifically, we calculate the style consistency between $f$ and ${b}^{+}$ as $c=\operatorname{Gram}(f, {b}^{+})$, where $\operatorname{Gram}(\cdot)$ means Gram Matrix \cite{gatys2016image}. Similarly, we can obtain ${c}^{+}=\operatorname{Gram}({f}^{+}, {b}^{+})$, and ${c}^{-}_{k}=\operatorname{Gram}(f_{k}^{-}, {b}^{+}), k=1,2,3,...,K$.

Thus, CS-CR can be formulated as:
\begin{equation}
\label{equ:cs-cr}
{\mathcal{L}}_{cs-cr}= \frac{D\left(c, c^{+}\right)}{D\left(c, c^{+}\right)+\sum_{k=1}^{K} D\left(c, c_{k}^{-}\right)},
\end{equation}
As shown in Figure \ref{fig:scs-cr}, our CS-CR focuses on the foreground-background style consistency, pulling $c$ closer to ${c}^{+}$ and pushing $c$ away from $\{c_{k}^{-}\}_{k=1}^{K}$.

Finally, the total loss function for training is:
\begin{equation}
\begin{aligned}
&{\mathcal{L}}_{scs-cr}={\mathcal{L}}_{ss-cr}+{\mathcal{L}}_{cs-cr},\\
&\mathcal{L}={\mathcal{L}}_{rec}+\lambda \cdot {\mathcal{L}}_{scs-cr}.
\label{equ:loss}
\end{aligned}
\end{equation}
where $\lambda$ is a hyperparameter for balancing the reconstruction loss and SCS-CR.
\vspace{-1.0em}
\paragraph{Difference with the triplet loss.}
\label{cr-discussion}
Compared with the triplet loss~\cite{cong2021bargainnet}, as shown in Figure \ref{fig:scs-cr}, we use a contrastive learning framework. Our SCS-Co dynamically generates $K$ negative samples online and pushes the output harmonized image away from them. Through such multiple push operations, more powerful constraints can be performed in the representation space. In addition, our SCS-Co does not simply pull $f$ to $b^{+}$, but constrains from the perspective of foreground-background style consistency. More experiments demonstrate our SCS-Co outperforms the triplet loss for image harmonization (see Section \ref{sec:Ablation}).
\begin{table*}[htbp]
	\centering
	\caption{Quantitative comparison across four sub-datasets of iHarmony4 \cite{cong2020dovenet}. $\uparrow$ means the higher the better, and $\downarrow$ means the lower the better. \textcolor[rgb]{ 1,  0,  0}{\textbf{Red}} and \textcolor[rgb]{ 0,  0,  1}{\underline{blue}} indicate the best and second best performance, respectively.}
	\resizebox{\textwidth}{!}{
		\begin{tabular}{c|r|rrrrrrrrrr}
			\toprule
			Dataset & Metric & Composite & DIH \cite{tsai2017deep}   & S$^2$AM \cite{cun2020improving}  & DoveNet \cite{cong2020dovenet} & BargainNet \cite{cong2021bargainnet} & Guo \etal \cite{guo2021intrinsic} & RainNet \cite{ling2021region} & iS$^2$AM \cite{sofiiuk2021foreground} & D-HT \cite{Guo_2021_ICCV}  & Ours \\
			\midrule
			\multirow{3}[2]{*}{HCOCO} & PSNR$\uparrow$  & 33.94 & 34.69 & 35.47 & 35.83 & 37.03 & 37.16 & 37.08 & \textcolor[rgb]{ 0,  0,  1}{\underline{39.16}} & 38.76 & \textcolor[rgb]{ 1,  0,  0}{\textbf{39.88}} \\
			& MSE$\downarrow$   & 69.37 & 51.85 & 41.07 & 36.72 & 24.84 & 24.92 & 29.52     & \textcolor[rgb]{ 0,  0,  1}{\underline{16.48}} & 16.89 & \textcolor[rgb]{ 1,  0,  0}{\textbf{13.58}} \\
			& fMSE$\downarrow$  & 996.59 & 798.99 & 542.06 & 551.01 & 397.85     & 416.38 & 501.17     & \textcolor[rgb]{ 0,  0,  1}{\underline{266.19}} & 299.30 & \textcolor[rgb]{ 1,  0,  0}{\textbf{245.54}} \\
			\midrule
			\multirow{3}[2]{*}{HAdobe5K} & PSNR$\uparrow$  & 28.16 & 32.28 & 33.77 & 34.34 & 35.34 & 35.20  & 36.22 & \textcolor[rgb]{ 0,  0,  1}{\underline{38.08}} & 36.88 & \textcolor[rgb]{ 1,  0,  0}{\textbf{38.29}} \\
			& MSE$\downarrow$   & 345.54 & 92.65 & 63.40  & 52.32 & 39.94 & 43.02 & 43.35     & \textcolor[rgb]{ 0,  0,  1}{\underline{21.88}} & 38.53 & \textcolor[rgb]{ 1,  0,  0}{\textbf{21.01}} \\
			& fMSE$\downarrow$  & 2051.61 & 593.03 & 404.62 & 380.39 & 279.66     & 284.21 & 317.55     & \textcolor[rgb]{ 0,  0,  1}{\underline{173.96}} & 265.11 & \textcolor[rgb]{ 1,  0,  0}{\textbf{165.48}} \\
			\midrule
			\multirow{3}[2]{*}{HFlickr} & PSNR$\uparrow$  & 28.32 & 29.55 & 30.03 & 30.21 & 31.34 & 31.34 & 31.64 & \textcolor[rgb]{ 0,  0,  1}{\underline{33.56}} & 33.13 & \textcolor[rgb]{ 1,  0,  0}{\textbf{34.22}} \\
			& MSE$\downarrow$   & 264.35 & 163.38 & 143.45 & 133.14 & 97.32 & 105.13 & 110.59     & \textcolor[rgb]{ 0,  0,  1}{\underline{69.67}} & 74.51 & \textcolor[rgb]{ 1,  0,  0}{\textbf{55.83}} \\
			& fMSE$\downarrow$  & 1574.37 & 1099.13 & 785.65 & 827.03 & 698.40     & 716.6 & 688.40     & \textcolor[rgb]{ 0,  0,  1}{\underline{443.65}} & 515.45 & \textcolor[rgb]{ 1,  0,  0}{\textbf{393.72}} \\
			\midrule
			\multirow{3}[2]{*}{Hday2night} & PSNR$\uparrow$  & 34.01 & 34.62 & 34.50  & 35.27 & 35.67 & 35.96 & 34.83 & \textcolor[rgb]{ 0,  0,  1}{\underline{37.72}} & 37.10  & \textcolor[rgb]{ 1,  0,  0}{\textbf{37.83}} \\
			& MSE$\downarrow$   & 109.65 & 82.34 & 76.61 & 51.95 & 50.98 & 55.53 & 57.40     & \textcolor[rgb]{ 1,  0,  0}{\textbf{40.59}} & 53.01 & \textcolor[rgb]{ 0,  0,  1}{\underline{41.75}} \\
			& fMSE$\downarrow$  & 1409.98 & 1129.40 & 989.07 & 1075.71 & 835.63     & 797.04 & 916.48     & \textcolor[rgb]{ 1,  0,  0}{\textbf{590.97}} & 704.42 & \textcolor[rgb]{ 0,  0,  1}{\underline{606.80}} \\
			\midrule
			\multirow{3}[2]{*}{Average} & PSNR  & 31.63 & 33.41 & 34.35 & 34.76 & 35.88 & 35.90  & 36.12 & \textcolor[rgb]{ 0,  0,  1}{\underline{38.19}} & 37.55 & \textcolor[rgb]{ 1,  0,  0}{\textbf{38.75}} \\
			& MSE$\downarrow$   & 172.47 & 76.77 & 59.67 & 52.33 & 37.82 & 38.71 & 40.29 & \textcolor[rgb]{ 0,  0,  1}{\underline{24.44}} & 30.30  & \textcolor[rgb]{ 1,  0,  0}{\textbf{21.33}} \\
			& fMSE$\downarrow$  & 1376.42 & 773.18 & 594.67 & 532.62 & 405.23 & 400.29 & 469.60 & \textcolor[rgb]{ 0,  0,  1}{\underline{264.96}} & 320.78 & \textcolor[rgb]{ 1,  0,  0}{\textbf{248.86}} \\
			\bottomrule
	\end{tabular}}%
	\label{tab:metric-result}%
\end{table*}%
\subsection{Background-attentional Adaptive Instance Normalization (BAIN)}
\label{sec:BAIN}
\begin{figure}[]
	\centering
	\includegraphics[width=\linewidth]{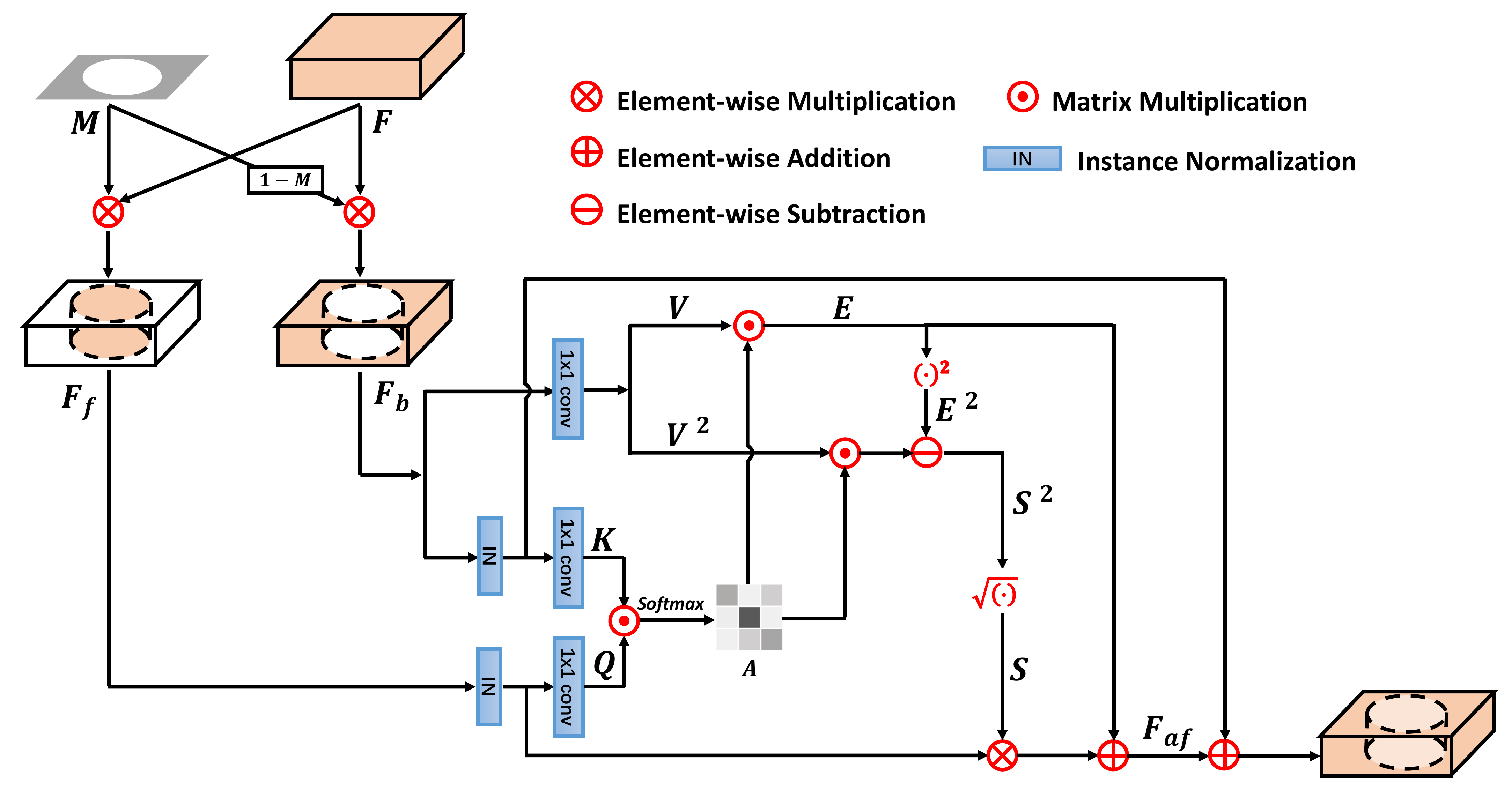} % Reduce the figure size so that it is slightly narrower than the column.
	\caption{Background-attentional Adaptive Instance Normalization (BAIN).}
	\label{fig:BAIN}
	\vspace{-1.0em}
\end{figure}

\paragraph{Formulation.}
We illustrate the structure of BAIN in Figure \ref{fig:BAIN}. Let $F \in \mathbb{R}^{C \times H \times W}$ be the feature map produced by the encoder and $M \in \mathbb{R}^{1 \times H \times W}$ be the resized foreground mask, where $C$, $H$, $W$ indicate the number of channels, height, and width of $F$, respectively.

Specifically, in order to learn the foreground feature and background feature individually, we first separate the foreground feature map and background feature map with the corresponding mask:
\begin{equation}
\begin{aligned}
&{F}_{b}={F} \cdot \bar{M},\\
&{F}_{f}={F} \cdot {M},
\end{aligned}
\end{equation}
where ${F}_{b}$ and ${F}_{f}$ are the background feature map and foreground feature map. Then, we adopt instance normalization to normalize ${F}_{b}$ and ${F}_{f}$. The normalized foreground feature map $\bar{F}_{f}$ at site $(c, h, w)$ is computed by:
\begin{equation}
\bar{F}_{f}^{c, h, w}=\frac{{F}_{f}^{c, h, w}-\mu_{f}^{c}}{\sigma_{f}^{c}},
\end{equation}
where $\mu_{f}^{c}$ and $\sigma_{f}^{c}$ denote the channel-wise mean and standard variance of the foreground feature map. Similarly, we can obtain the normalized background feature map $\bar{F}_{b}$. Further, we transform $\bar{F}_{f}$, $\bar{F}_{b}$ and ${F}_{b}$ into $Q$ (query), $K$ (key) and $V$ (value) as:
\begin{equation}
Q=f(\bar{F}_{f}),
K=b(\bar{F}_{b}),
V=k({F}_{b}),
\end{equation}
where $f(\cdot)$, $b(\cdot)$ and $k(\cdot)$ are $1 \times 1$ convolutions. Thus, the attention map $A \in \mathbb{R}^{HW \times HW}$ can be calculated as:
\begin{equation}
A=\operatorname{Softmax}\left(Q^{\top} \odot K\right),
\end{equation}
where $\odot$ indicates matrix multiplication. 

Then we calculate attention-weighted background expectation and standard variance respectively. The attention-weighted background expectation $E \in \mathbb{R}^{C \times HW}$ can be calculated as:
\begin{equation}
E=V \odot A^{\top},
\end{equation}
Since the variance of a variable equals to the expectation of its square
minus the square of its expectation, we can obtain the attention-weighted background standard variance $S \in \mathbb{R}^{C \times HW}$ as:
\begin{equation}
S=\sqrt{(V \cdot V) \odot A^{\top}-E \cdot E},
\end{equation}

Finally, we reshape $E$ and $S$ to $\mathbb{R}^{C \times H \times W}$, and align $\bar{F}_{f}$ with $E$ and $S$. The aligned foreground feature map ${F}_{af}$ at site $(c, h, w)$ is computed by:
\begin{equation}
F_{af}^{c, h, w}=S^{c, h, w} \cdot \bar{F}_{f}^{c, h, w}+E^{c, h, w}.
\end{equation}

\paragraph{Difference with RAIN.}
Inspired by AdaIN \cite{huang2017arbitrary}, Ling \etal propose RAIN~\cite{ling2021region} for image harmonization and achieve great success. However, just like the problem with AdaIN, RAIN only considers the holistic style distribution in the background and globally aligns the foreground feature distribution with that of the background feature. Different from RAIN, inspired by \cite{Liu_2021_ICCV,park2019arbitrary}, our BAIN can pay more attention to those areas in the background that feature-similar to the foreground, and based on this attention map, the attention-weighted expectation and standard variance of the background feature are calculated to locally align the foreground feature distribution.

\section{Experiments}
\begin{figure*}[htbp]
	\centering
	\begin{subfigure}{0.16\textwidth}
		% include first image
		\centering
		\includegraphics[width=\linewidth]{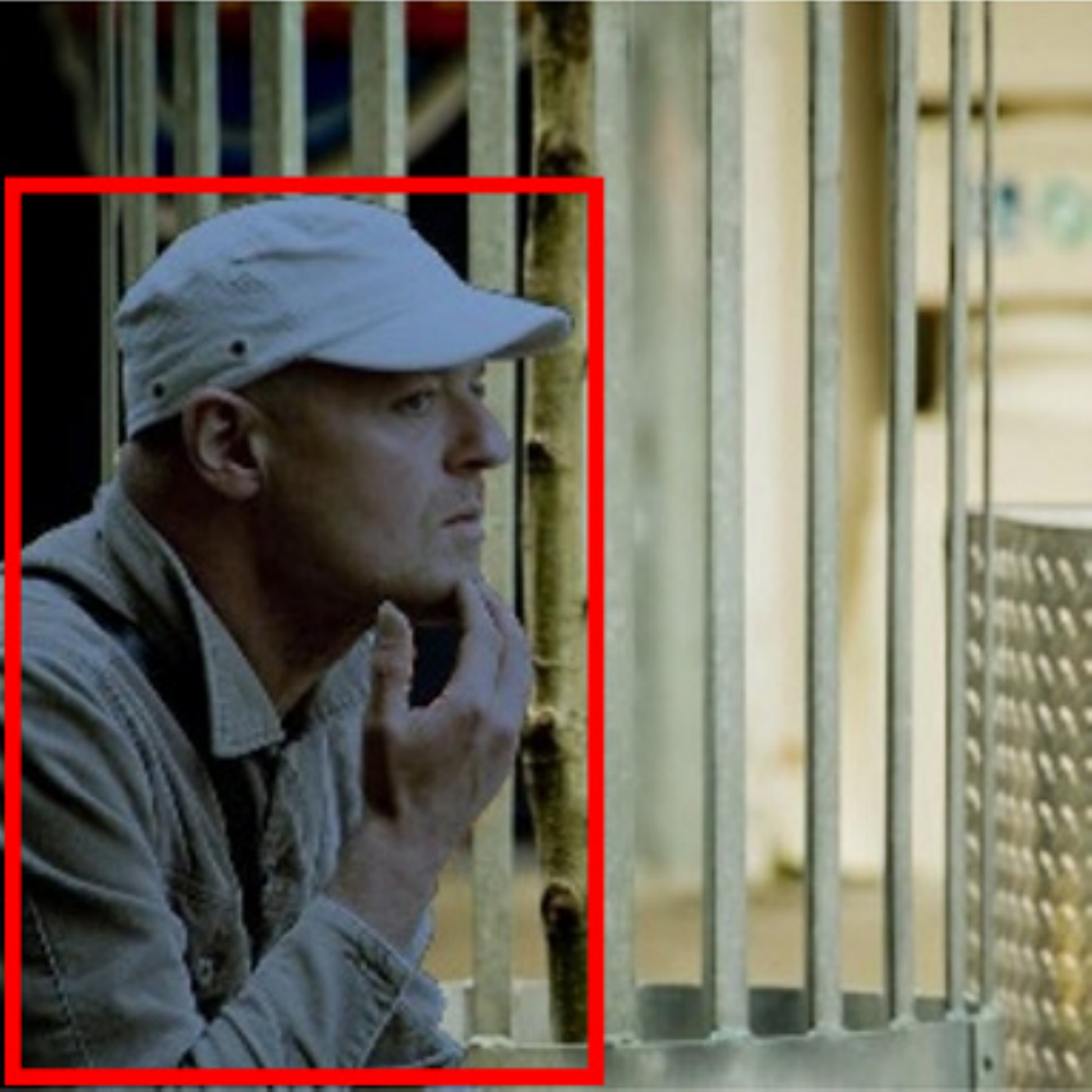}
	\end{subfigure}
	\hfil
	\begin{subfigure}{0.16\textwidth}
		% include second image
		\centering
		\includegraphics[width=\linewidth]{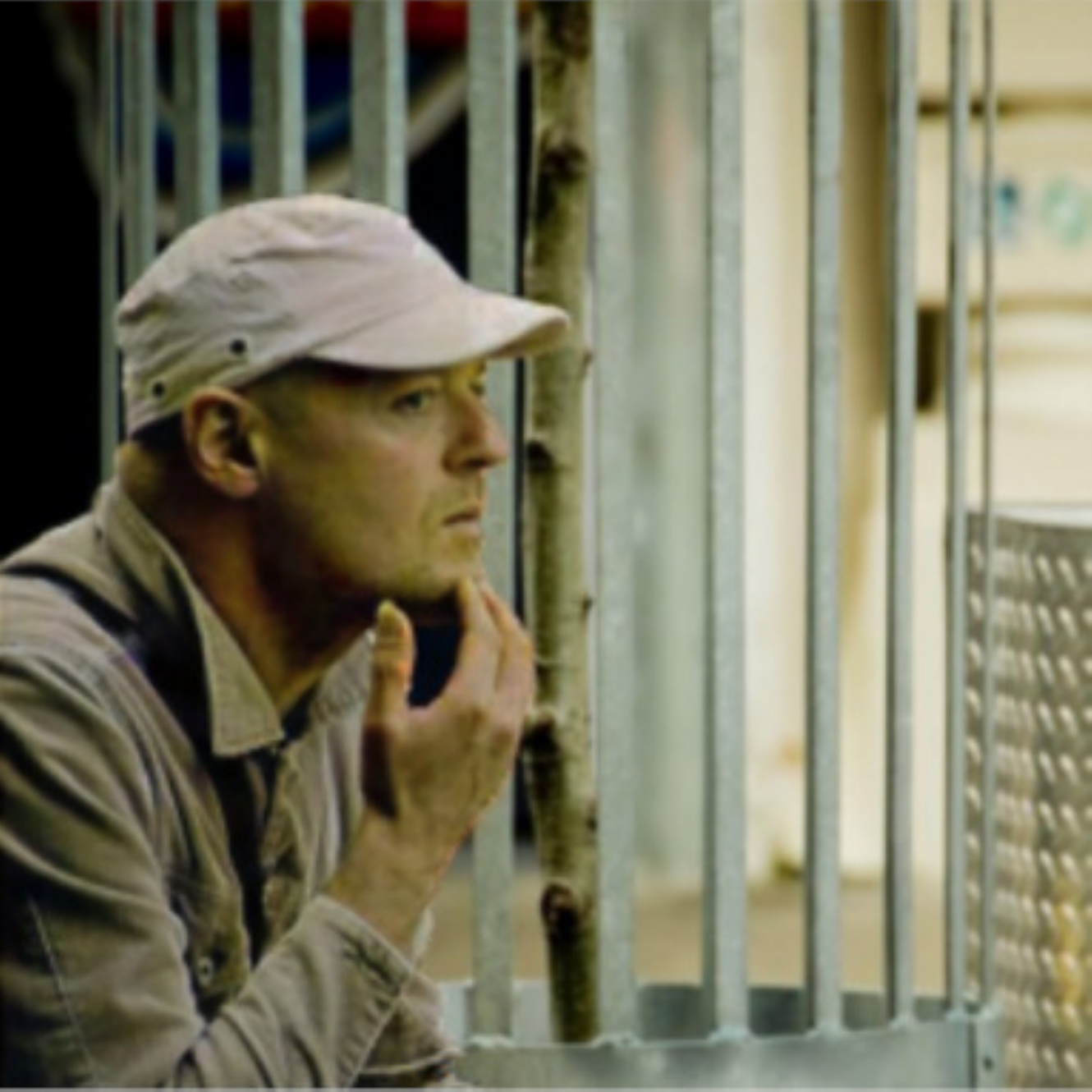}
	\end{subfigure}
	\hfil
	\begin{subfigure}{0.16\textwidth}
		% include second image
		\centering
		\includegraphics[width=\linewidth]{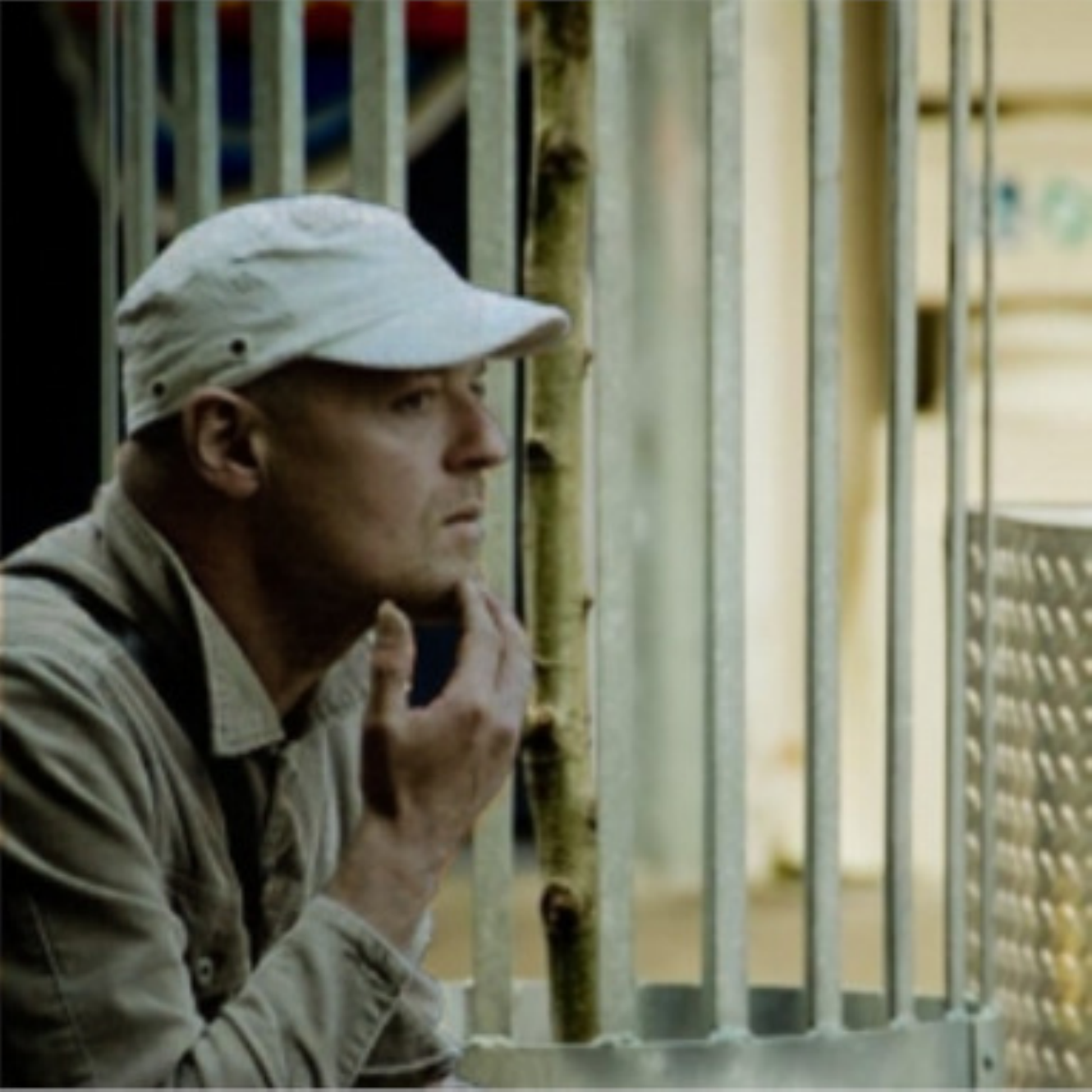}
	\end{subfigure}
	\hfil
	\begin{subfigure}{0.16\textwidth}
		% include first image
		\centering
		\includegraphics[width=\linewidth]{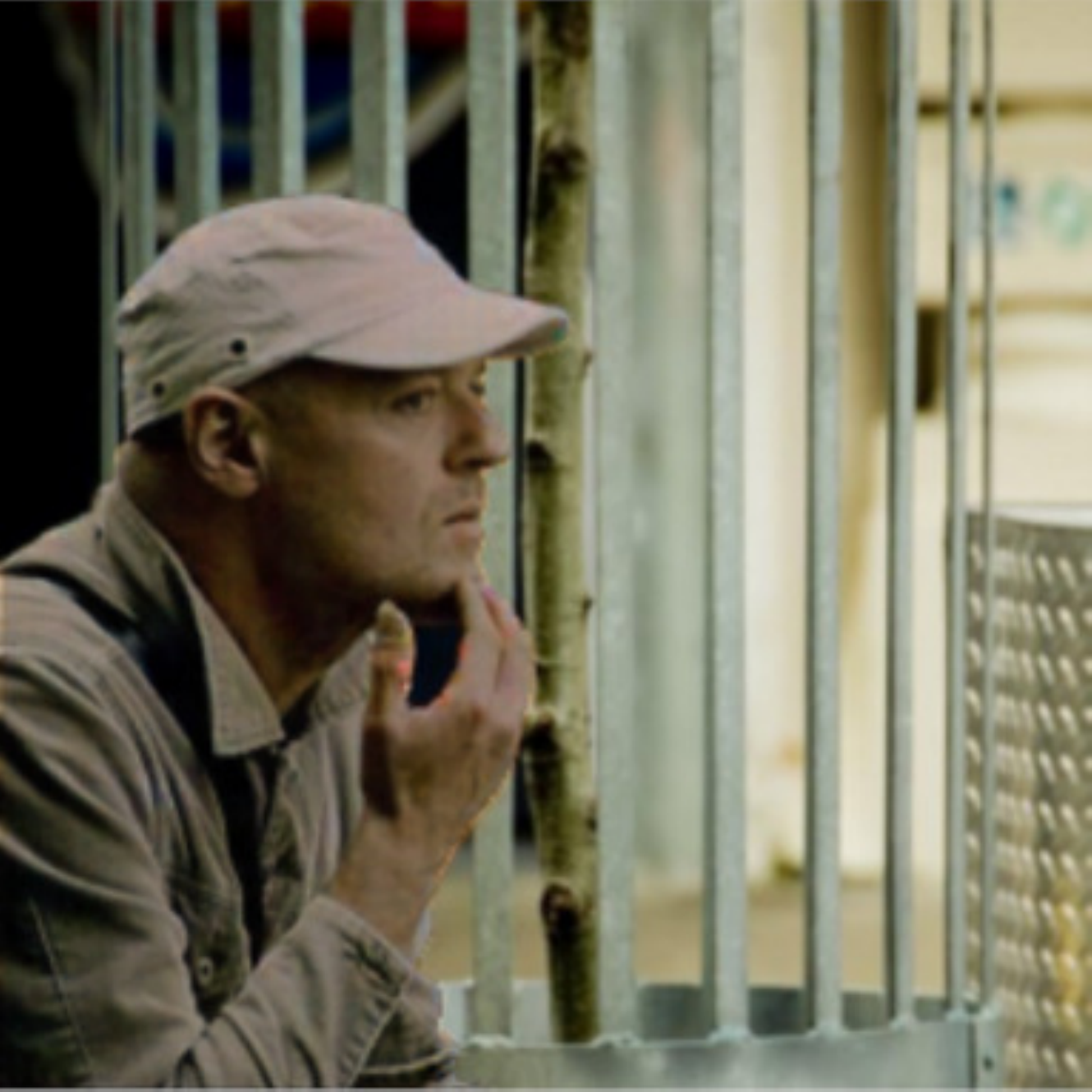}
	\end{subfigure}
	\hfil
	\begin{subfigure}{0.16\textwidth}
		% include second image
		\centering
		\includegraphics[width=\linewidth]{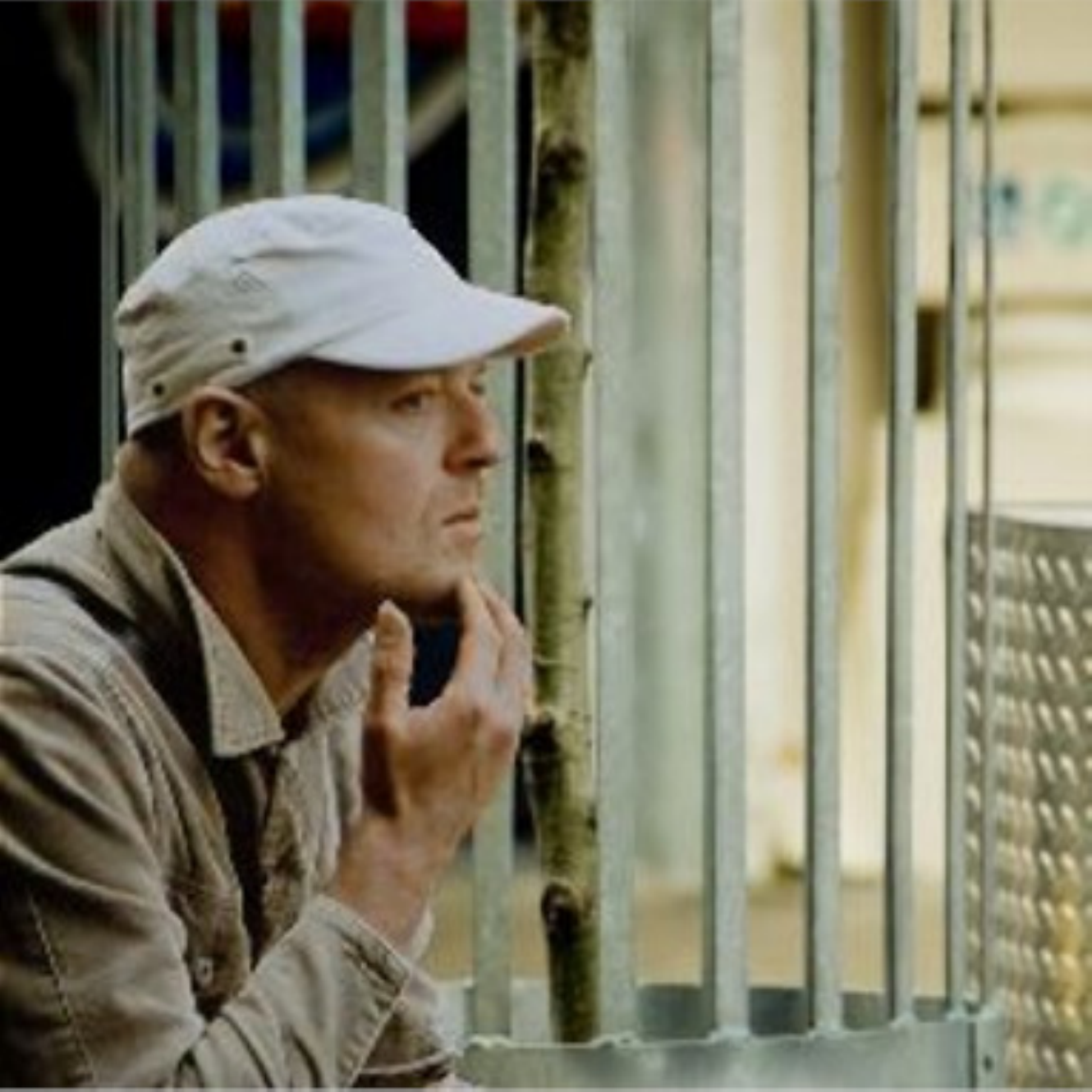}
	\end{subfigure}
	\hfil
	\begin{subfigure}{0.16\textwidth}
		% include second image
		\centering
		\includegraphics[width=\linewidth]{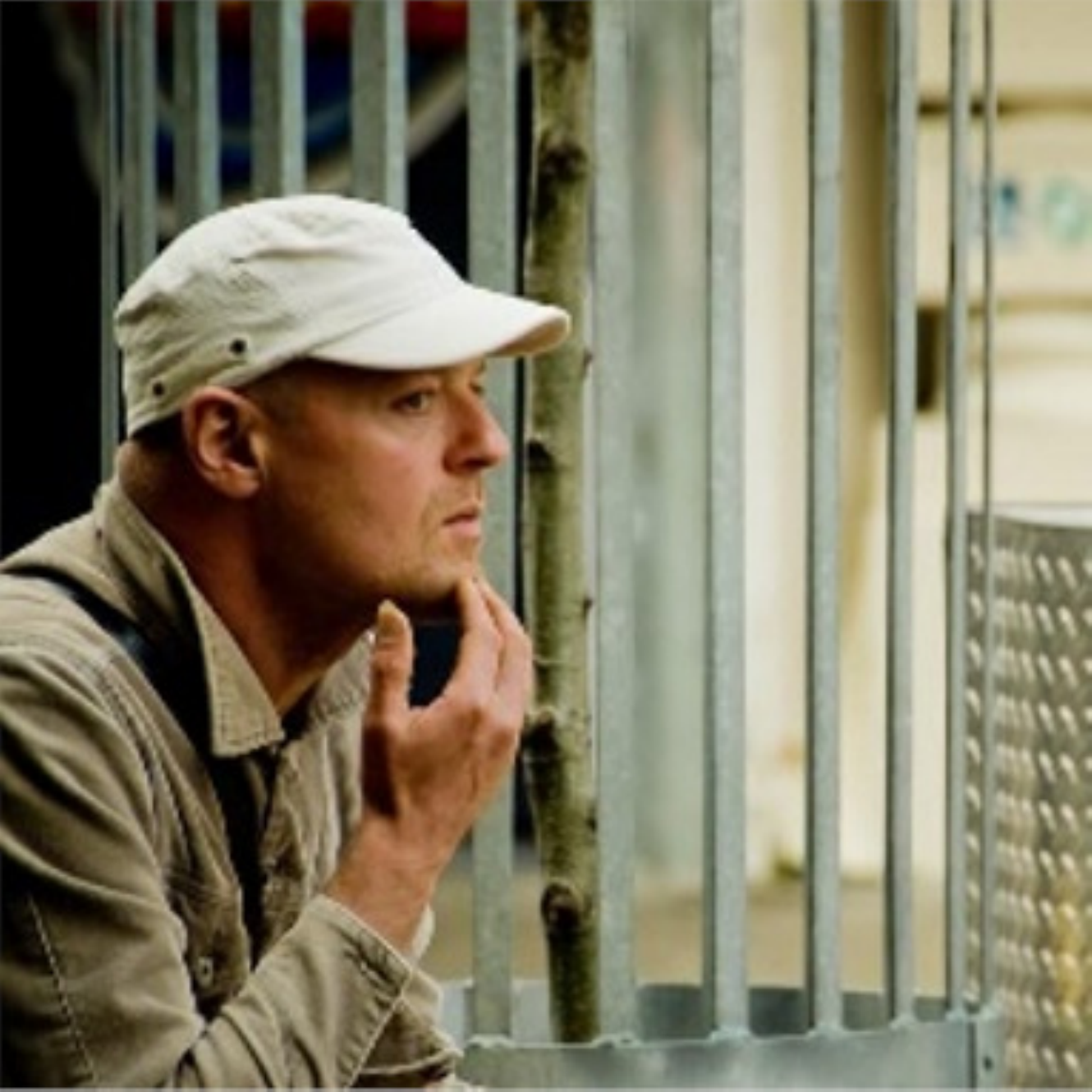}
	\end{subfigure}
	\quad
	\begin{subfigure}{0.16\textwidth}
		% include first image
		\centering
		\includegraphics[width=\linewidth]{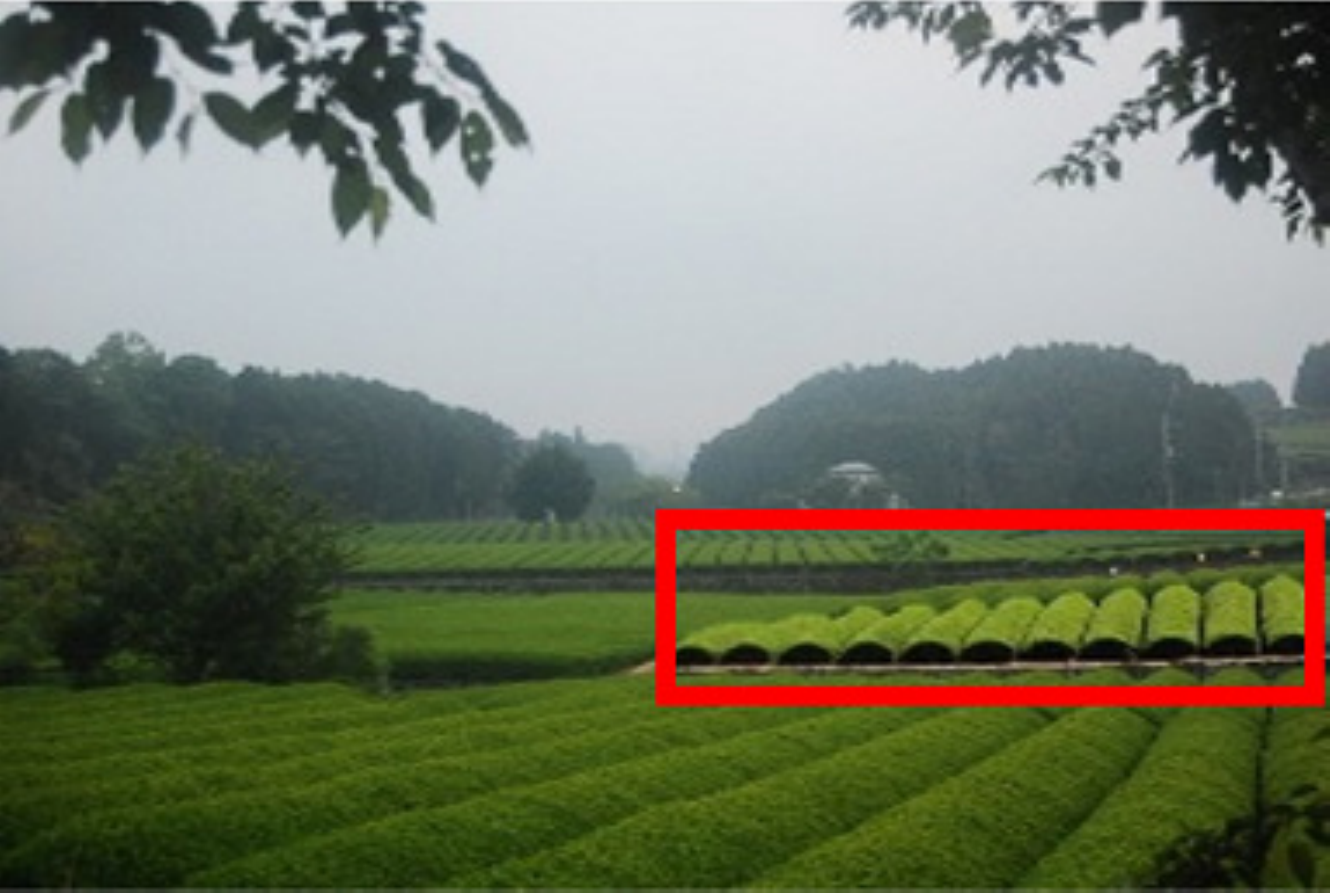}
	\end{subfigure}
	\hfil
	\begin{subfigure}{0.16\textwidth}
		% include second image
		\centering
		\includegraphics[width=\linewidth]{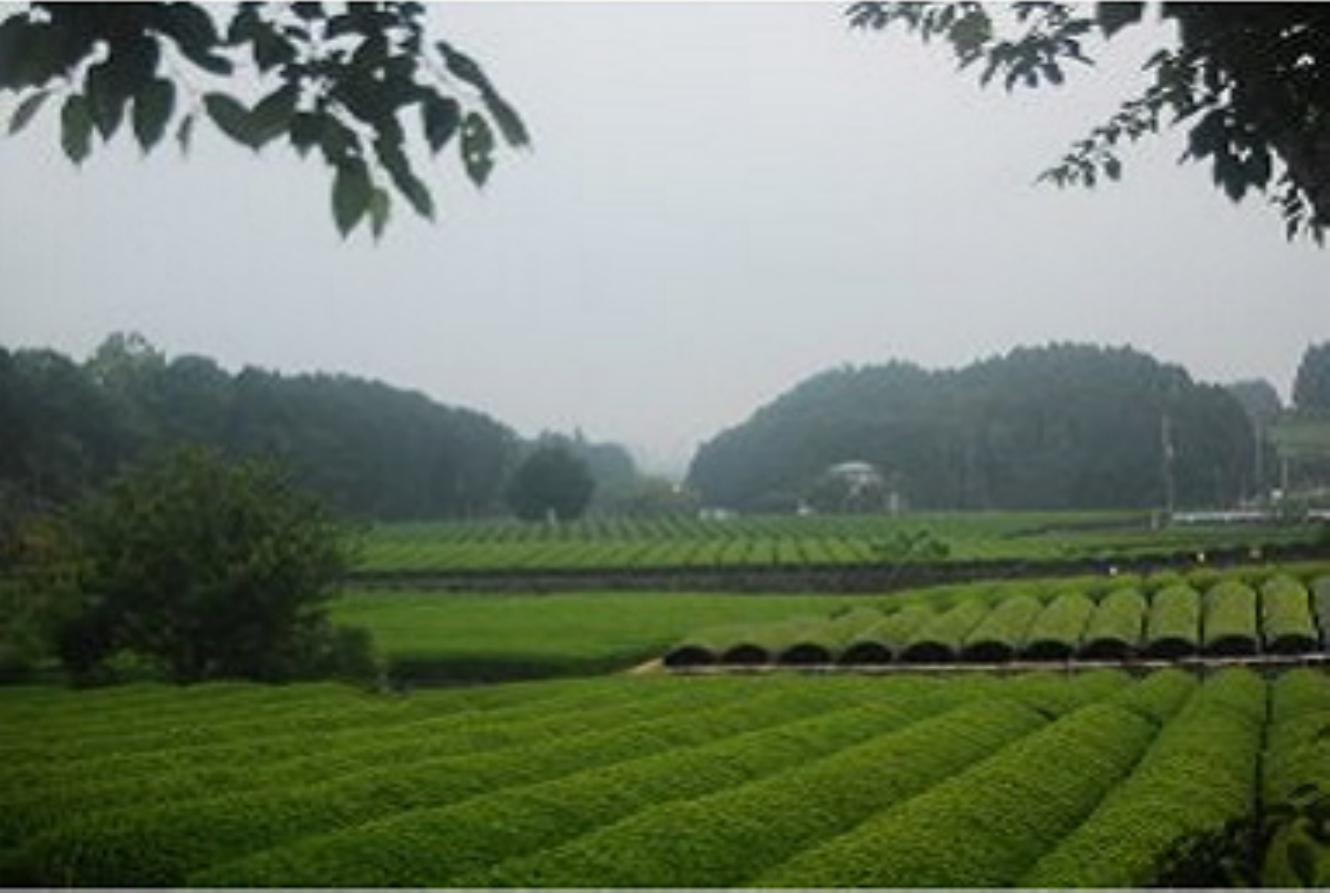}
	\end{subfigure}
	\hfil
	\begin{subfigure}{0.16\textwidth}
		% include second image
		\centering
		\includegraphics[width=\linewidth]{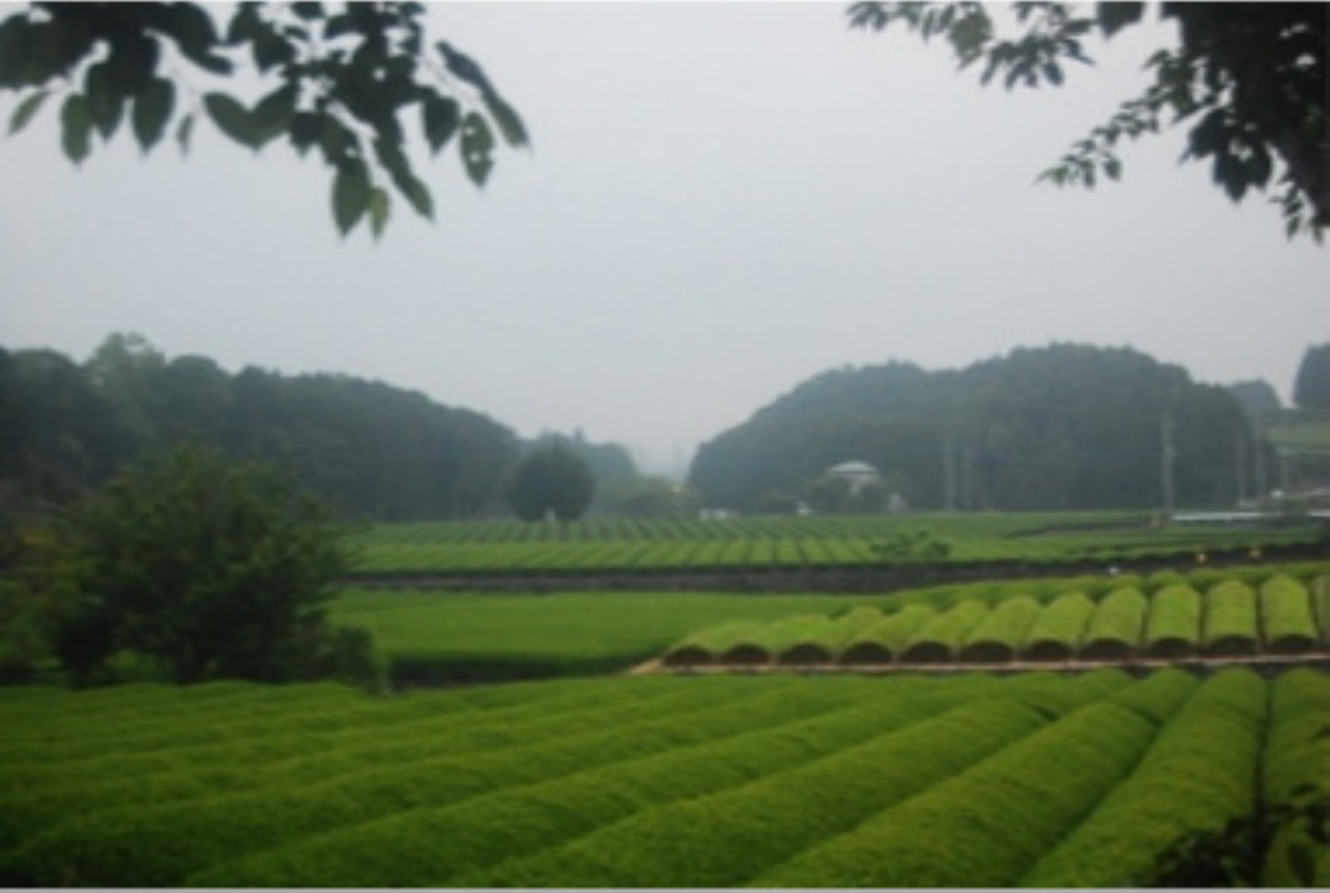}
	\end{subfigure}
	\hfil
	\begin{subfigure}{0.16\textwidth}
		% include first image
		\centering
		\includegraphics[width=\linewidth]{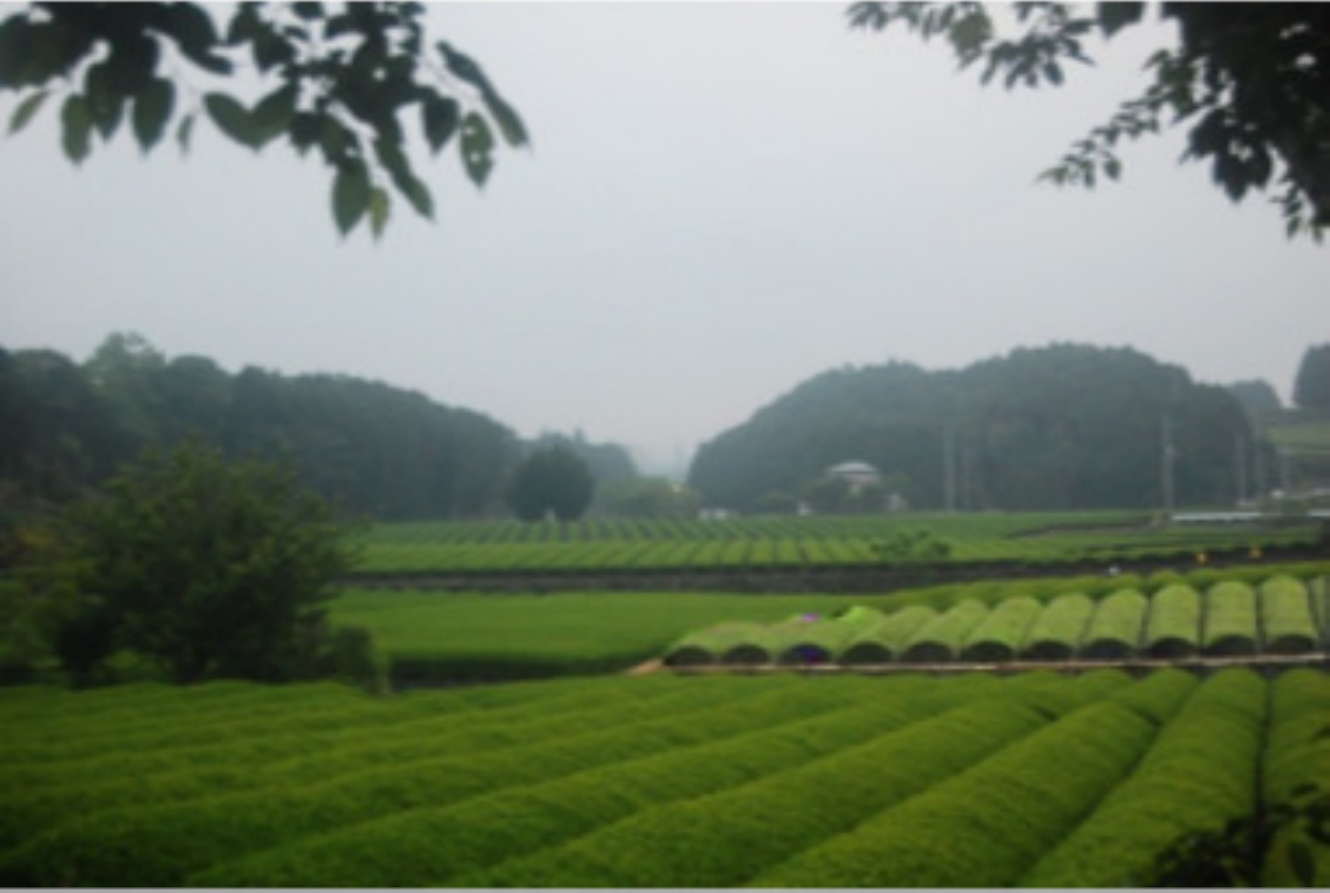}
	\end{subfigure}
	\hfil
	\begin{subfigure}{0.16\textwidth}
		% include second image
		\centering
		\includegraphics[width=\linewidth]{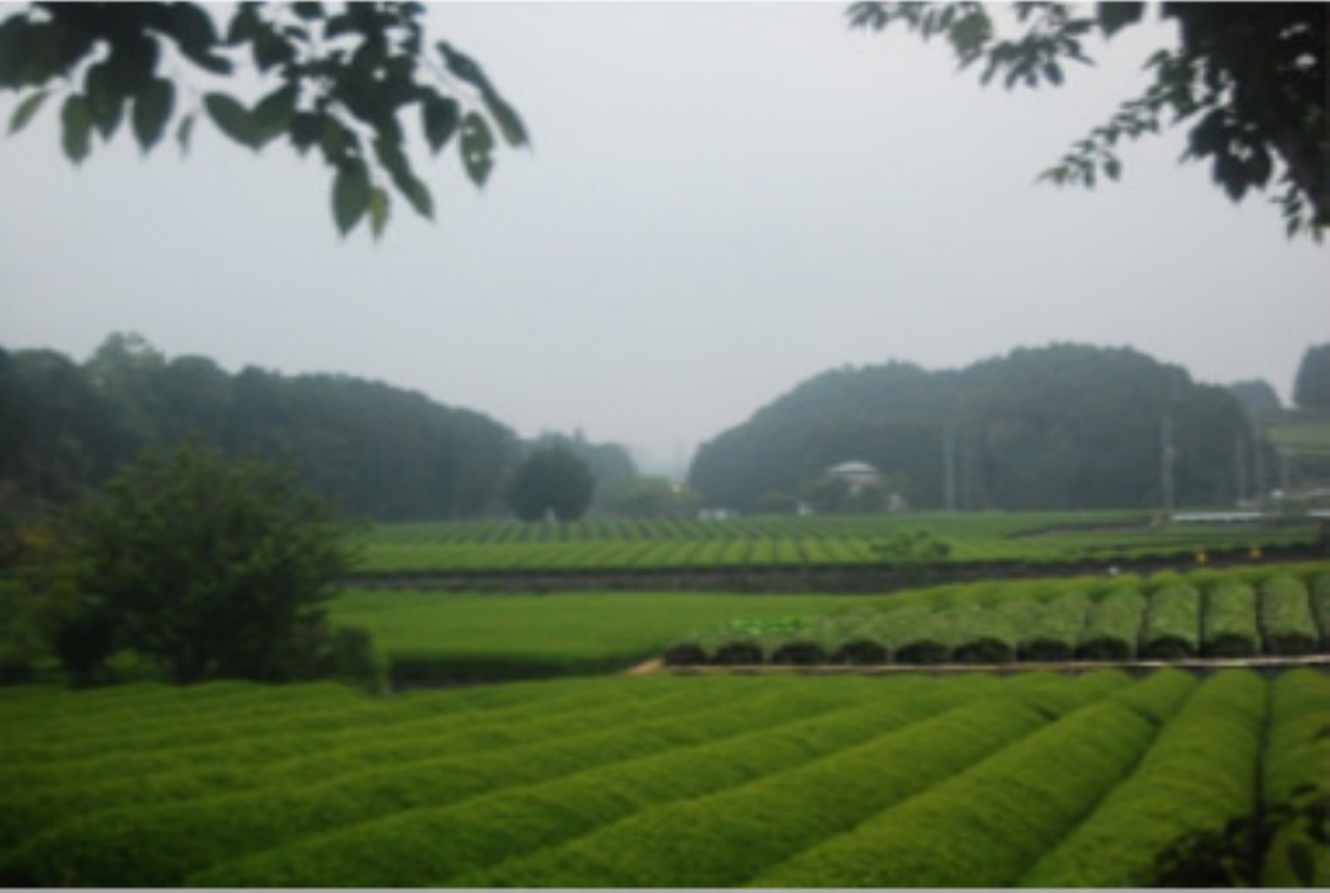}
	\end{subfigure}
	\hfil
	\begin{subfigure}{0.16\textwidth}
		% include second image
		\centering
		\includegraphics[width=\linewidth]{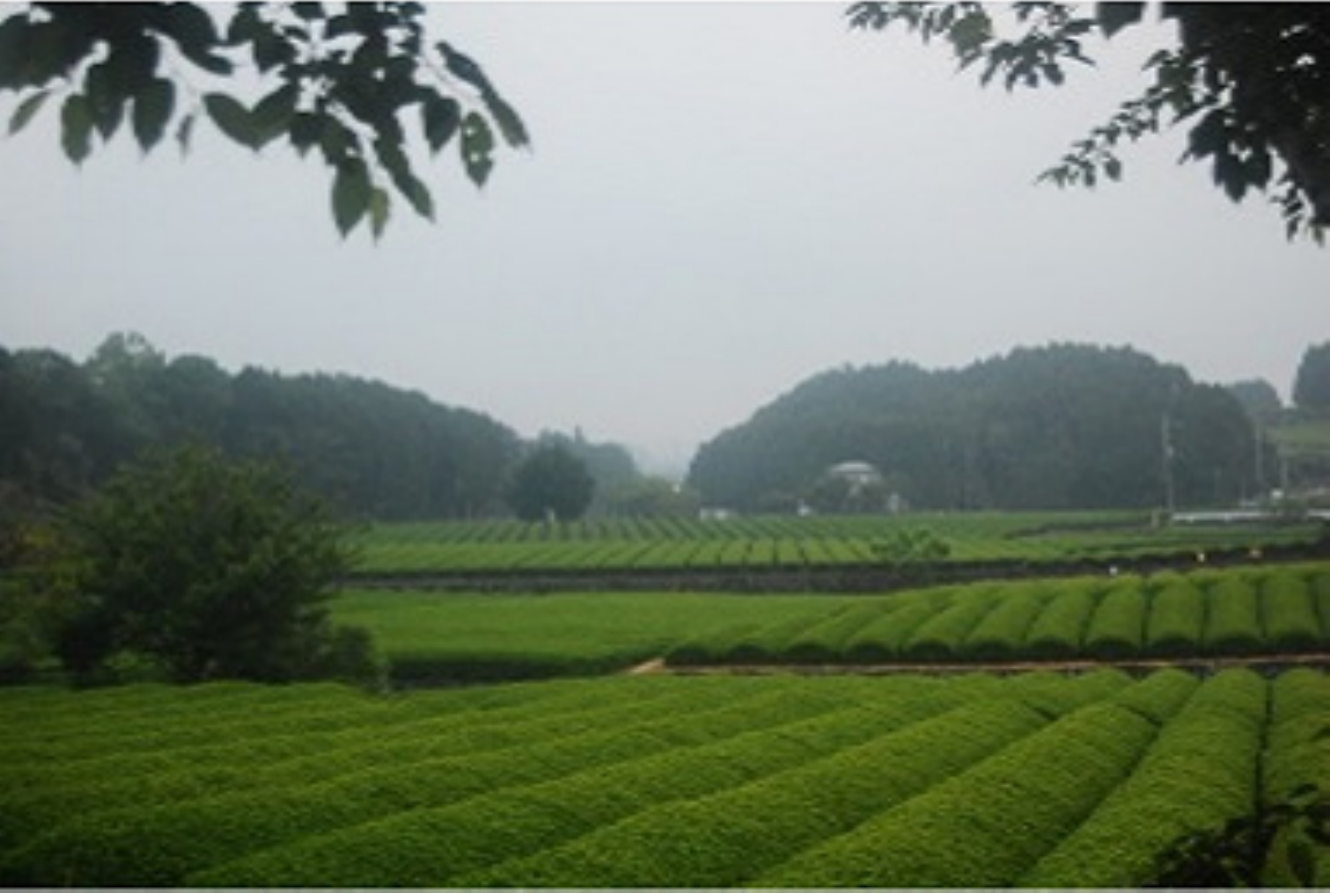}
	\end{subfigure}
	\quad
	\begin{subfigure}{0.16\textwidth}
		% include first image
		\centering
		\includegraphics[width=\linewidth]{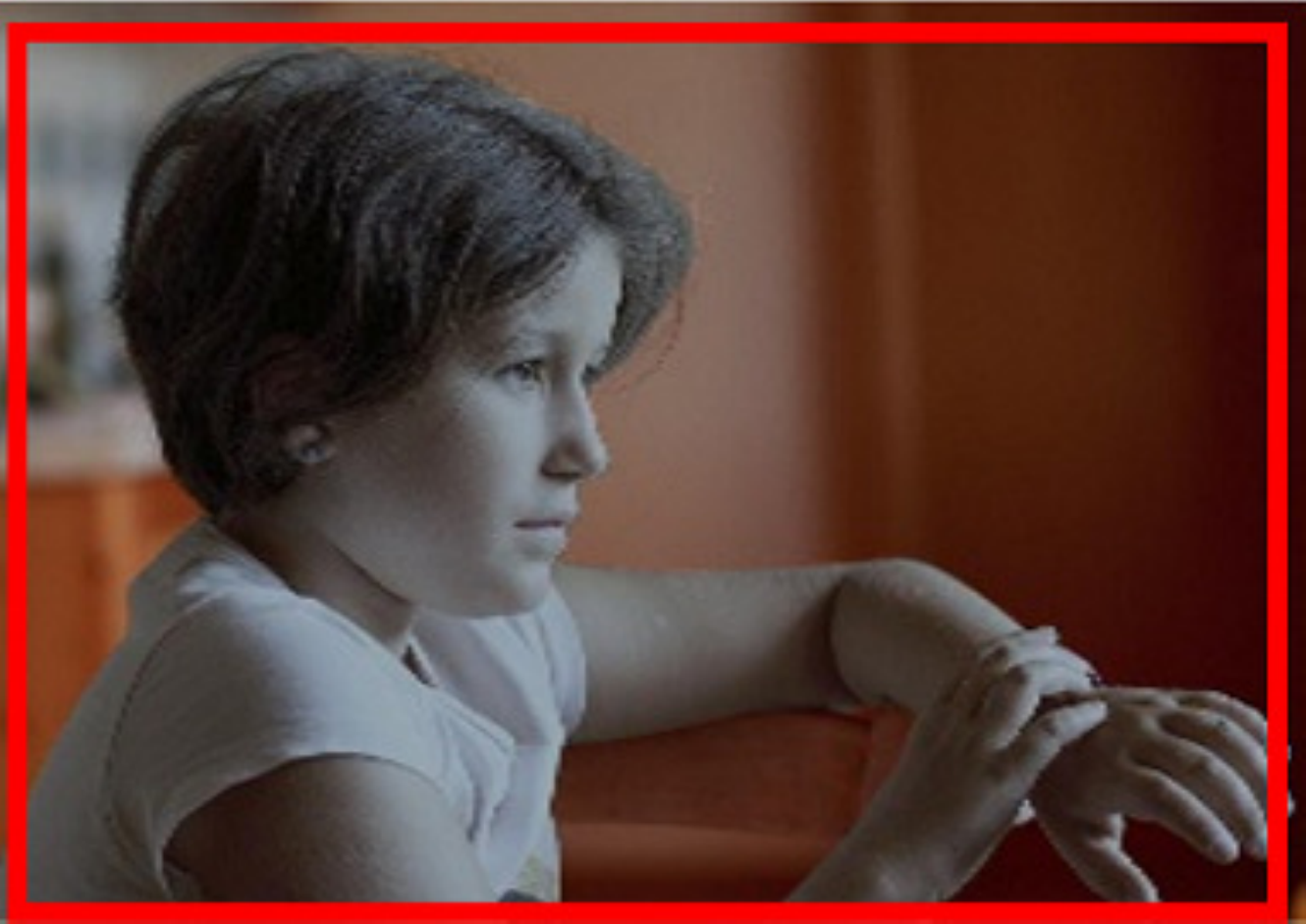}
	\end{subfigure}
	\hfil
	\begin{subfigure}{0.16\textwidth}
		% include second image
		\centering
		\includegraphics[width=\linewidth]{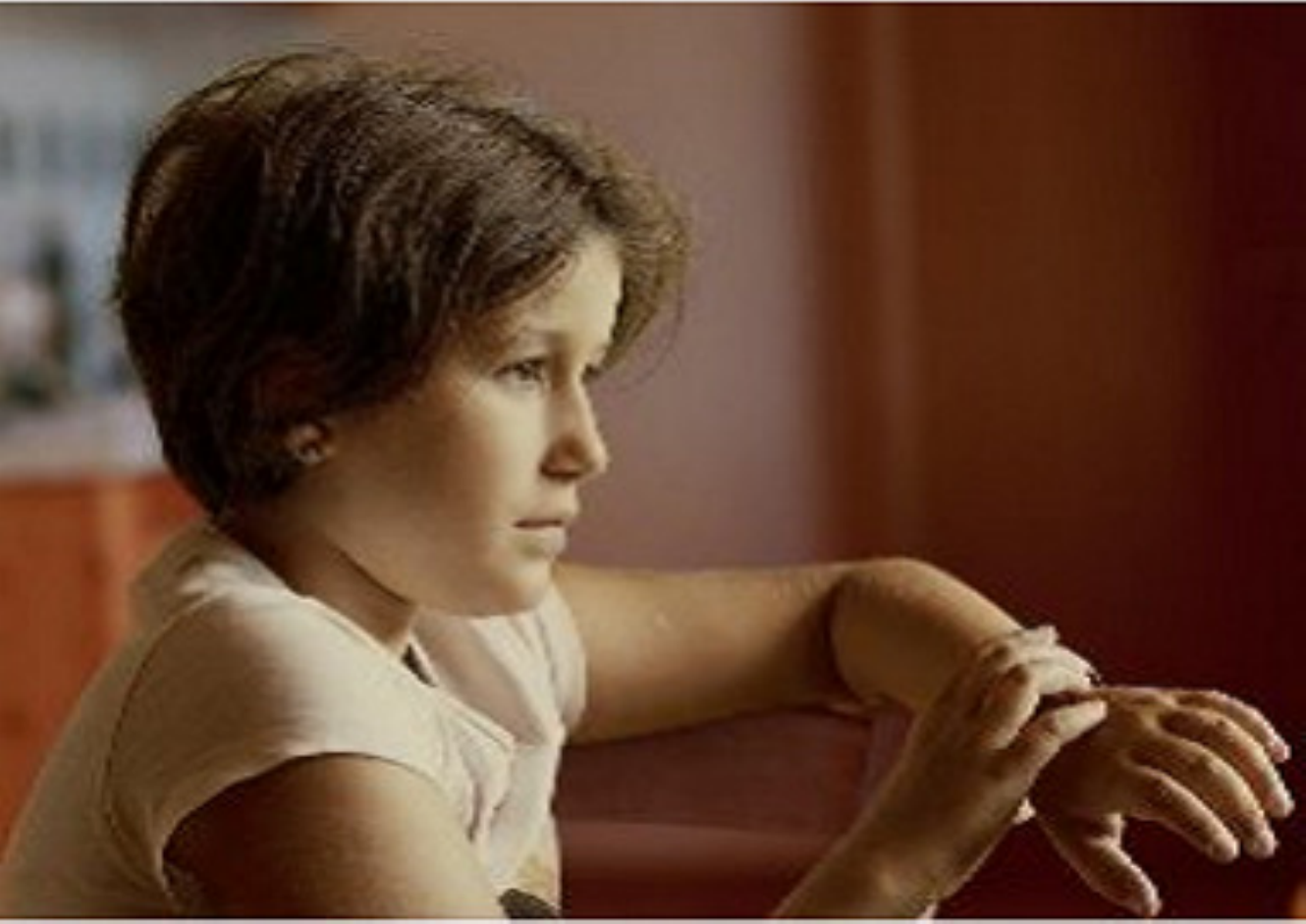}
	\end{subfigure}
	\hfil
	\begin{subfigure}{0.16\textwidth}
		% include second image
		\centering
		\includegraphics[width=\linewidth]{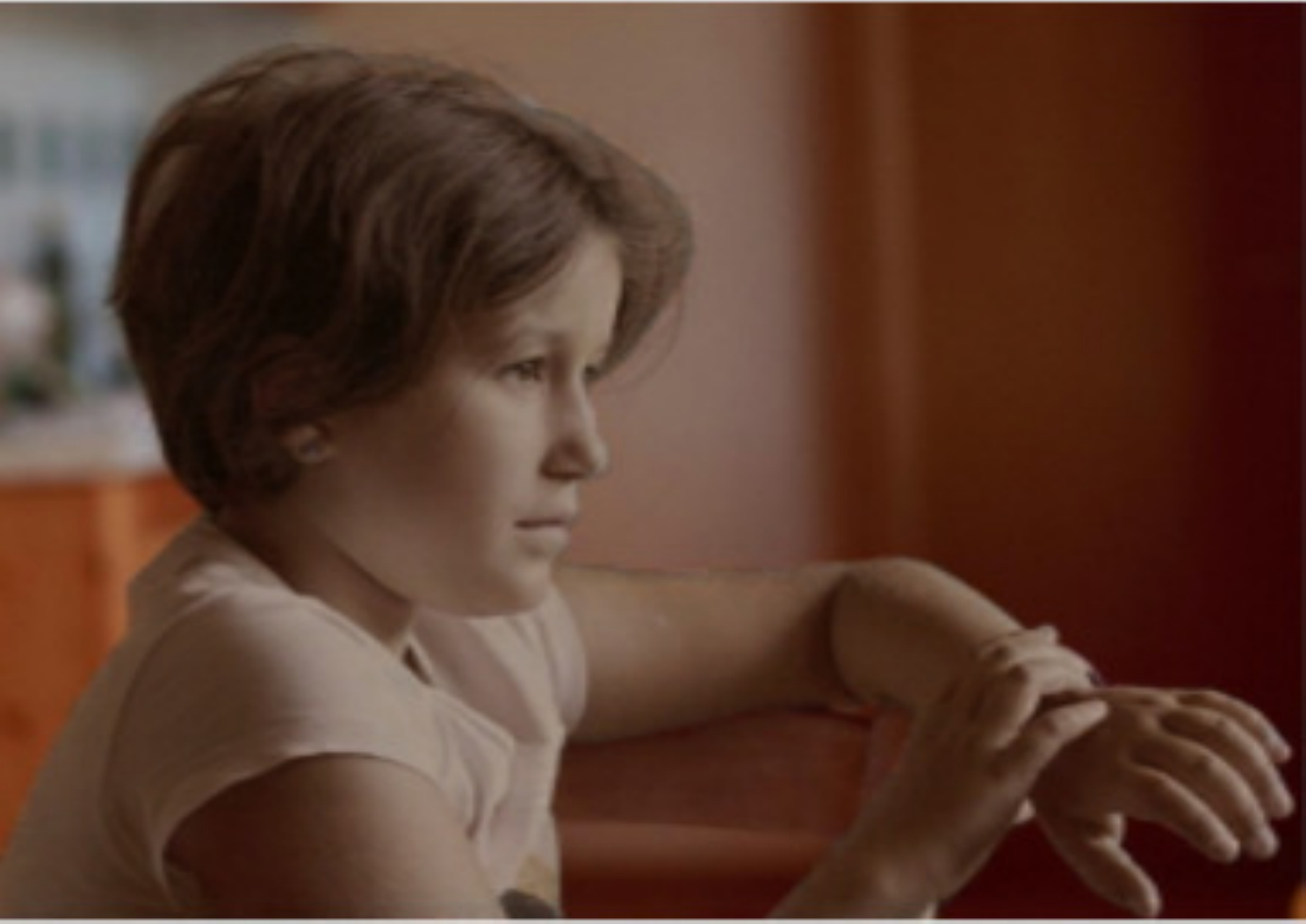}
	\end{subfigure}
	\hfil
	\begin{subfigure}{0.16\textwidth}
		% include first image
		\centering
		\includegraphics[width=\linewidth]{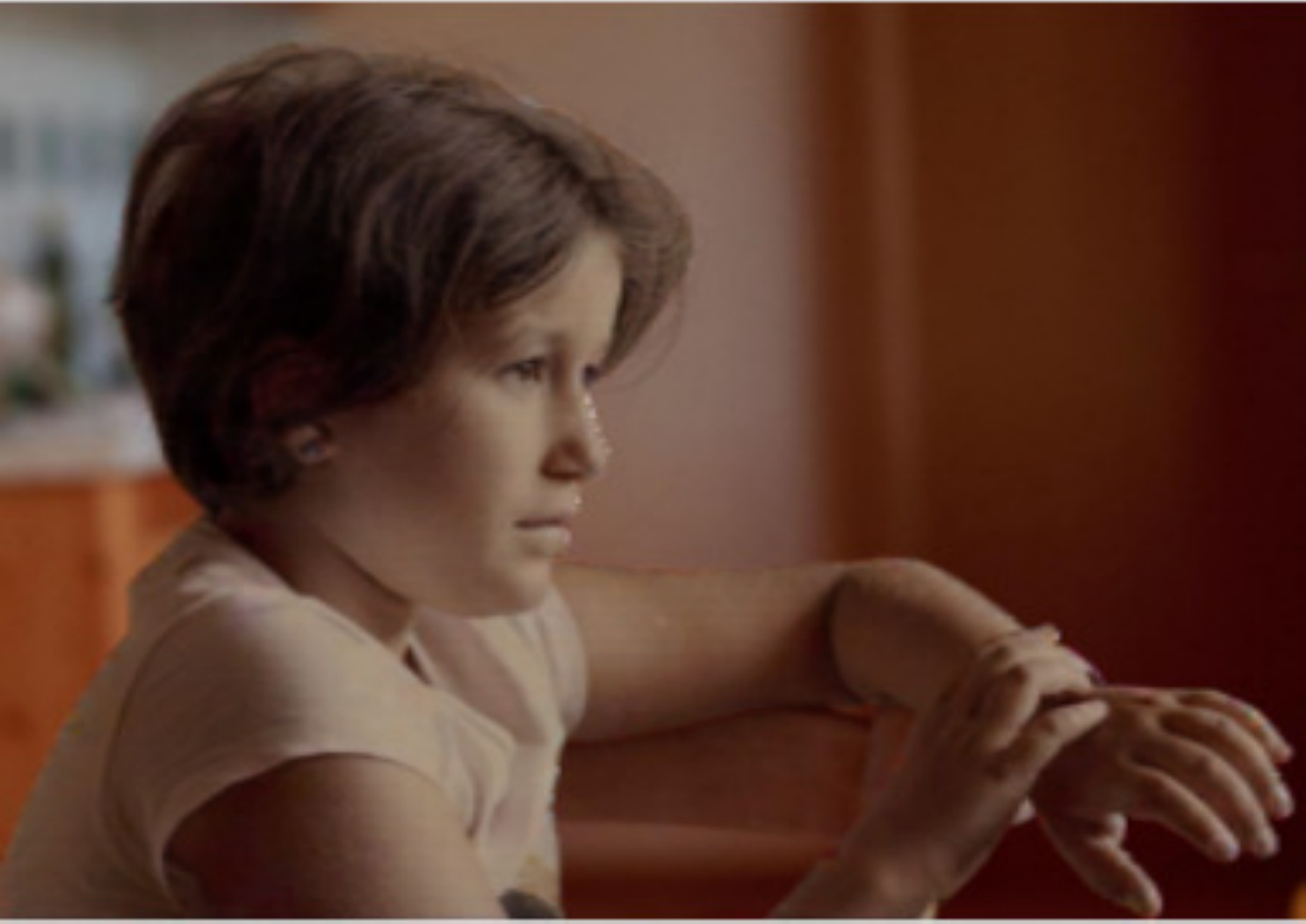}
	\end{subfigure}
	\hfil
	\begin{subfigure}{0.16\textwidth}
		% include second image
		\centering
		\includegraphics[width=\linewidth]{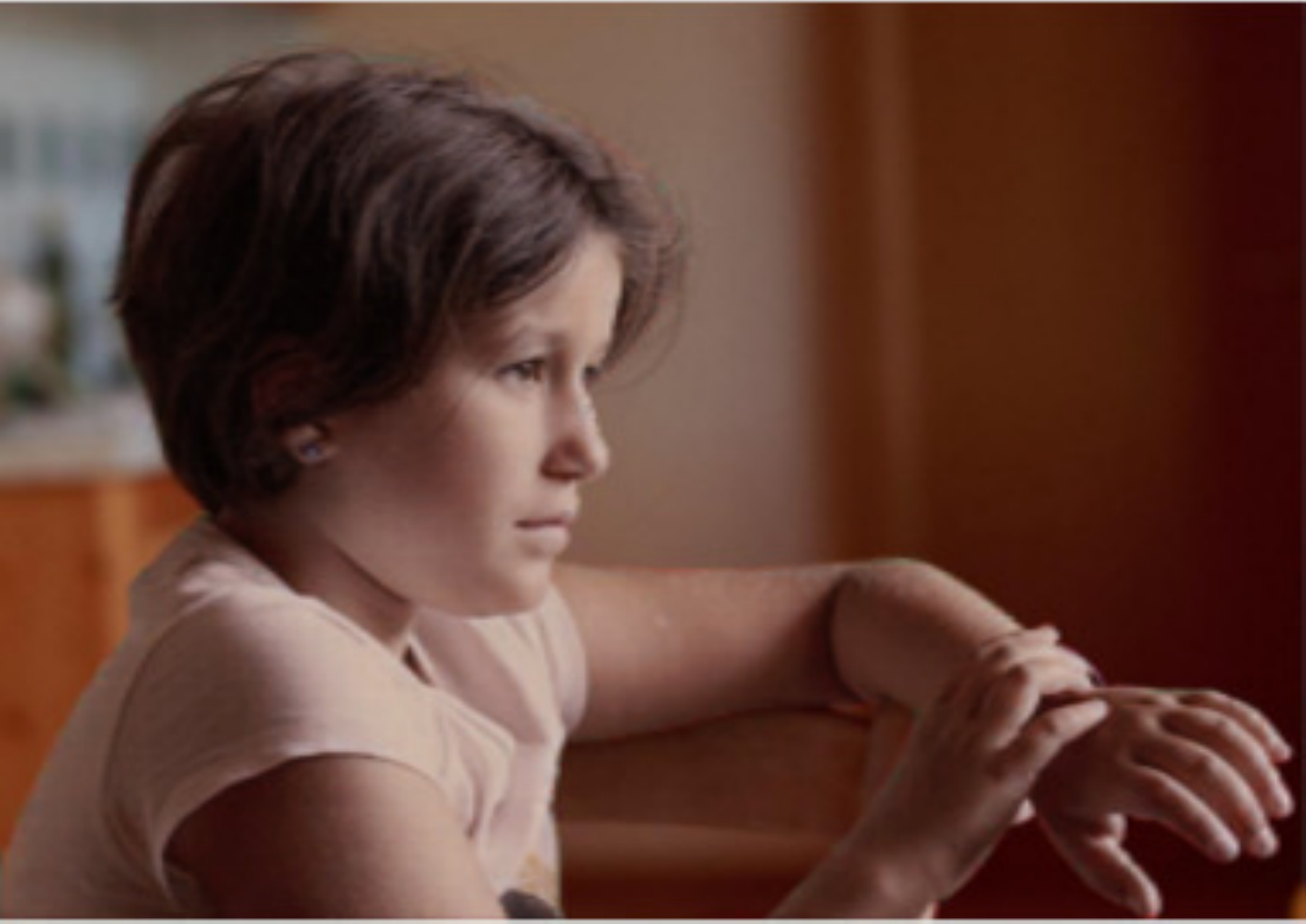}
	\end{subfigure}
	\hfil
	\begin{subfigure}{0.16\textwidth}
		% include second image
		\centering
		\includegraphics[width=\linewidth]{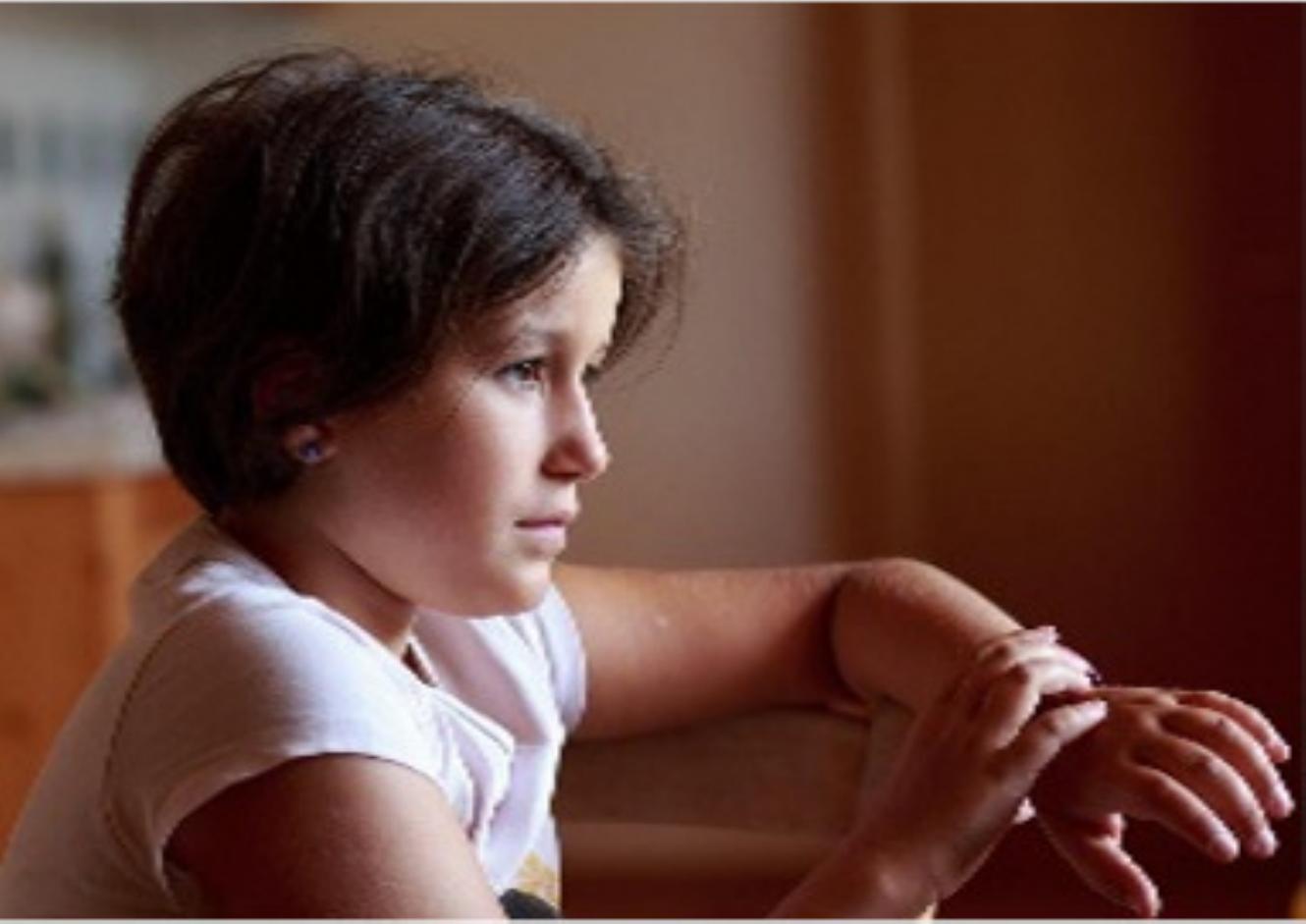}
	\end{subfigure}
	\quad
	\begin{subfigure}{0.16\textwidth}
		% include first image
		\centering
		\includegraphics[width=\linewidth]{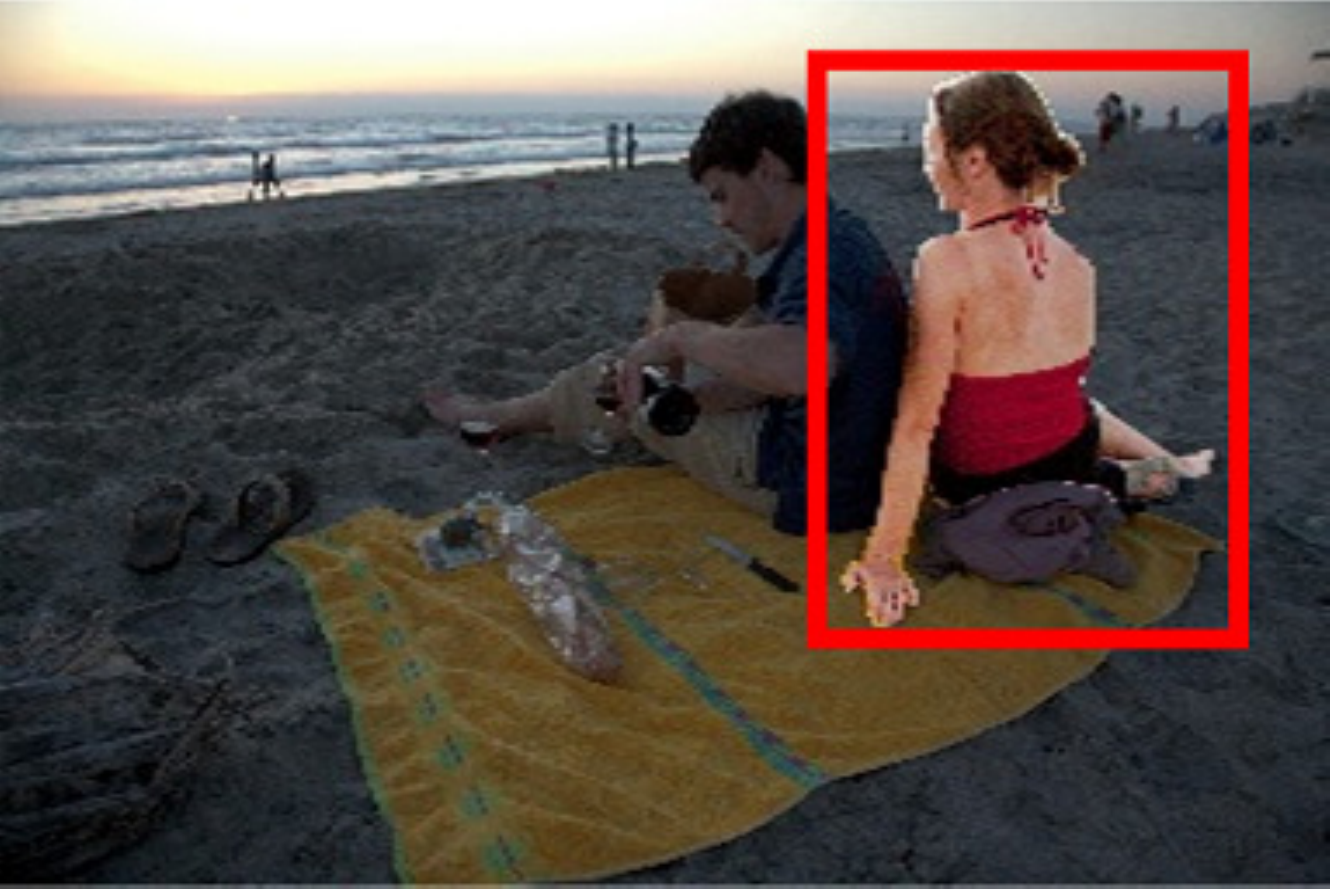}
		\caption*{Input}
	\end{subfigure}
	\hfil
	\begin{subfigure}{0.16\textwidth}
		% include second image
		\centering
		\includegraphics[width=\linewidth]{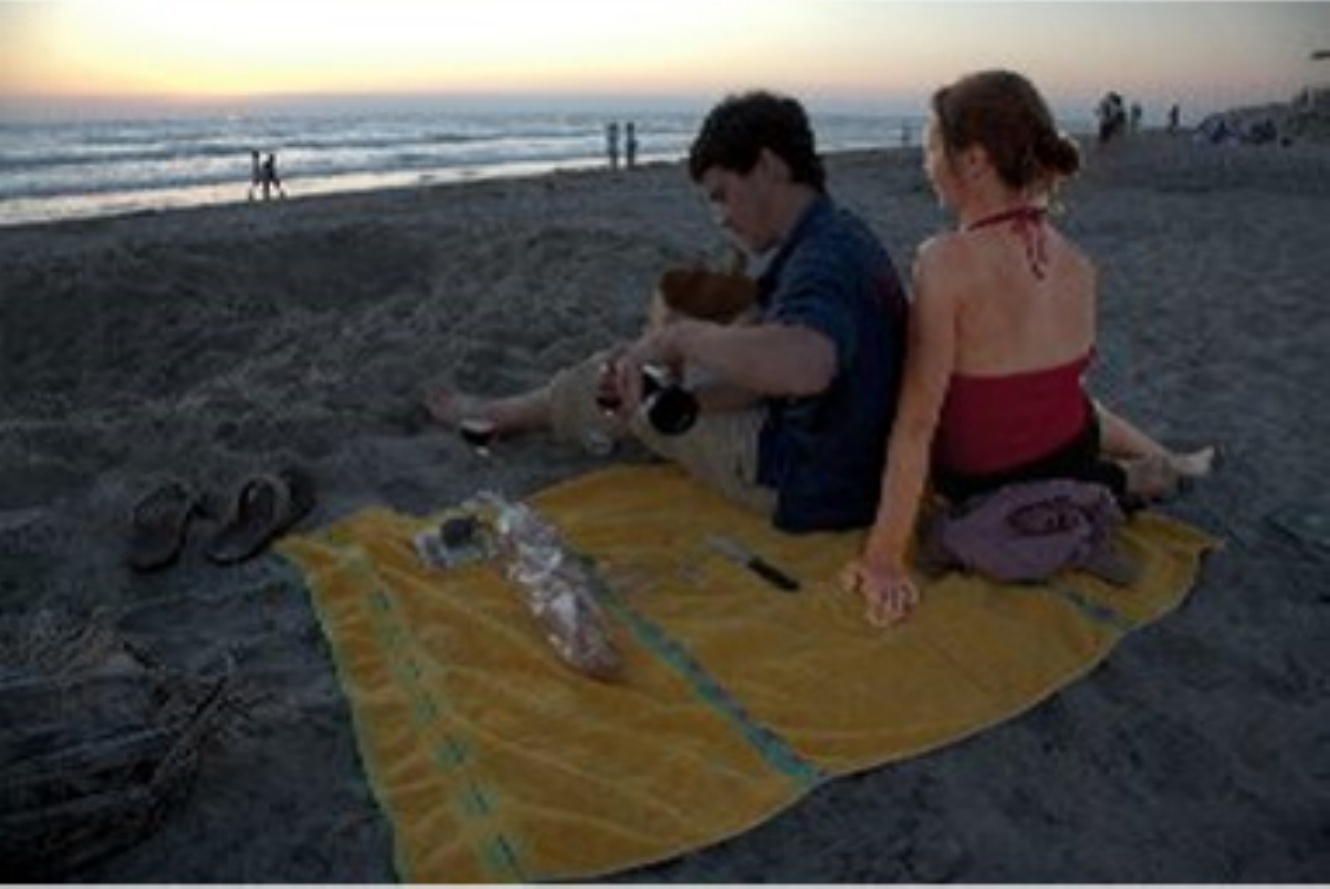}
		\caption*{iS$^2$AM \cite{sofiiuk2021foreground}}
	\end{subfigure}
	\hfil
	\begin{subfigure}{0.16\textwidth}
		% include second image
		\centering
		\includegraphics[width=\linewidth]{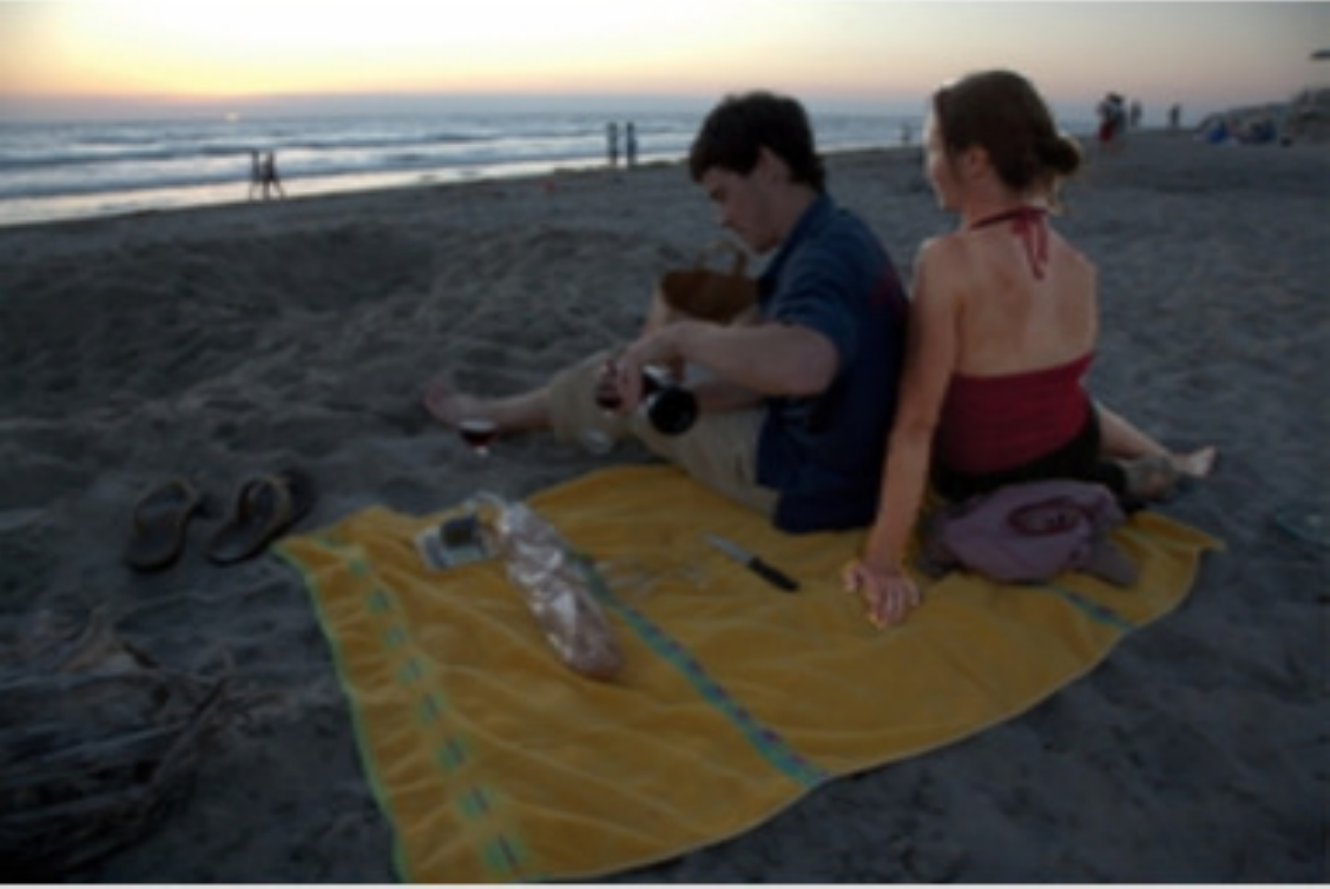}
		\caption*{DoveNet \cite{cong2020dovenet}}
	\end{subfigure}
	\hfil
	\begin{subfigure}{0.16\textwidth}
		% include first image
		\centering
		\includegraphics[width=\linewidth]{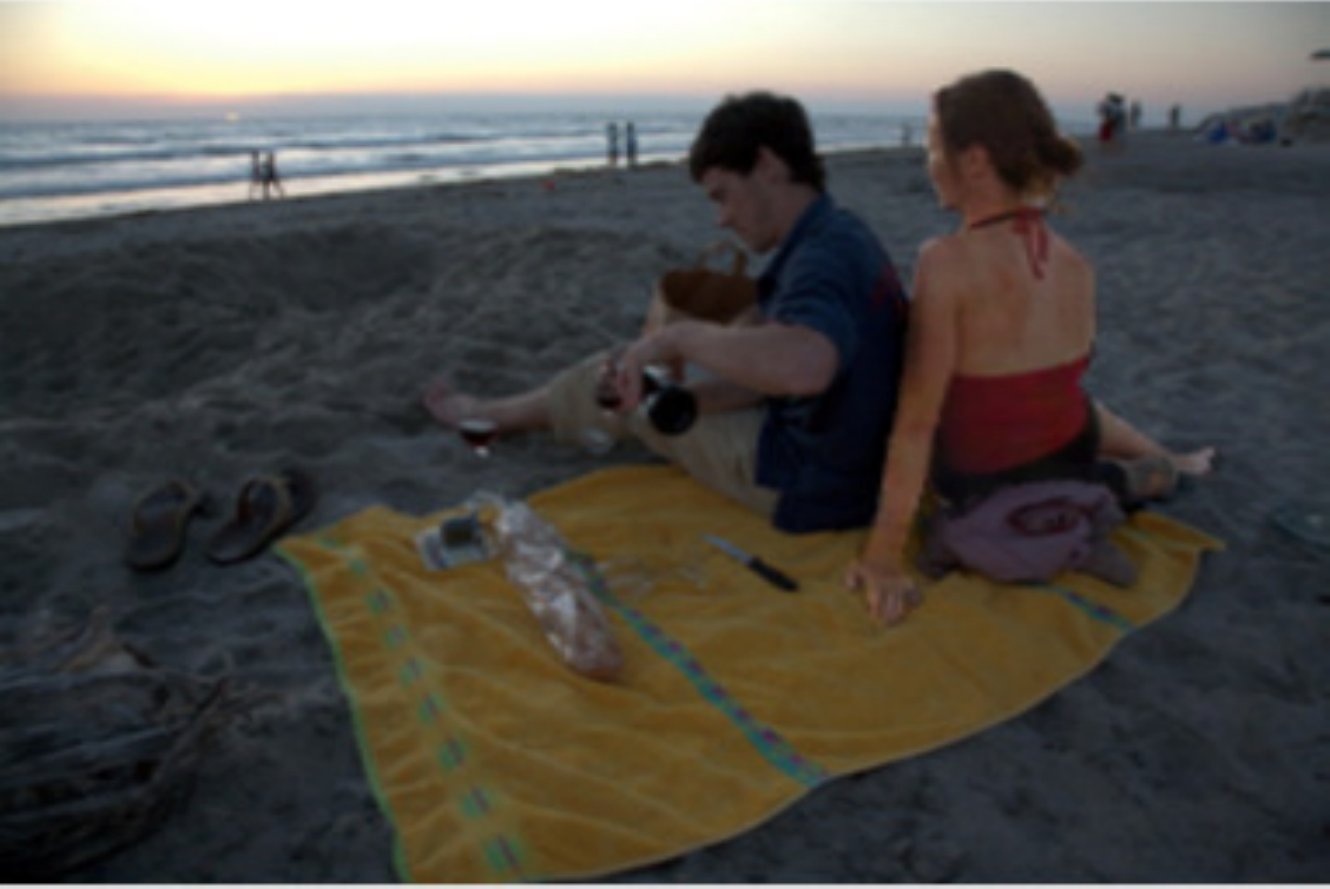}
		\caption*{RainNet \cite{ling2021region}}
	\end{subfigure}
	\hfil
	\begin{subfigure}{0.16\textwidth}
		% include second image
		\centering
		\includegraphics[width=\linewidth]{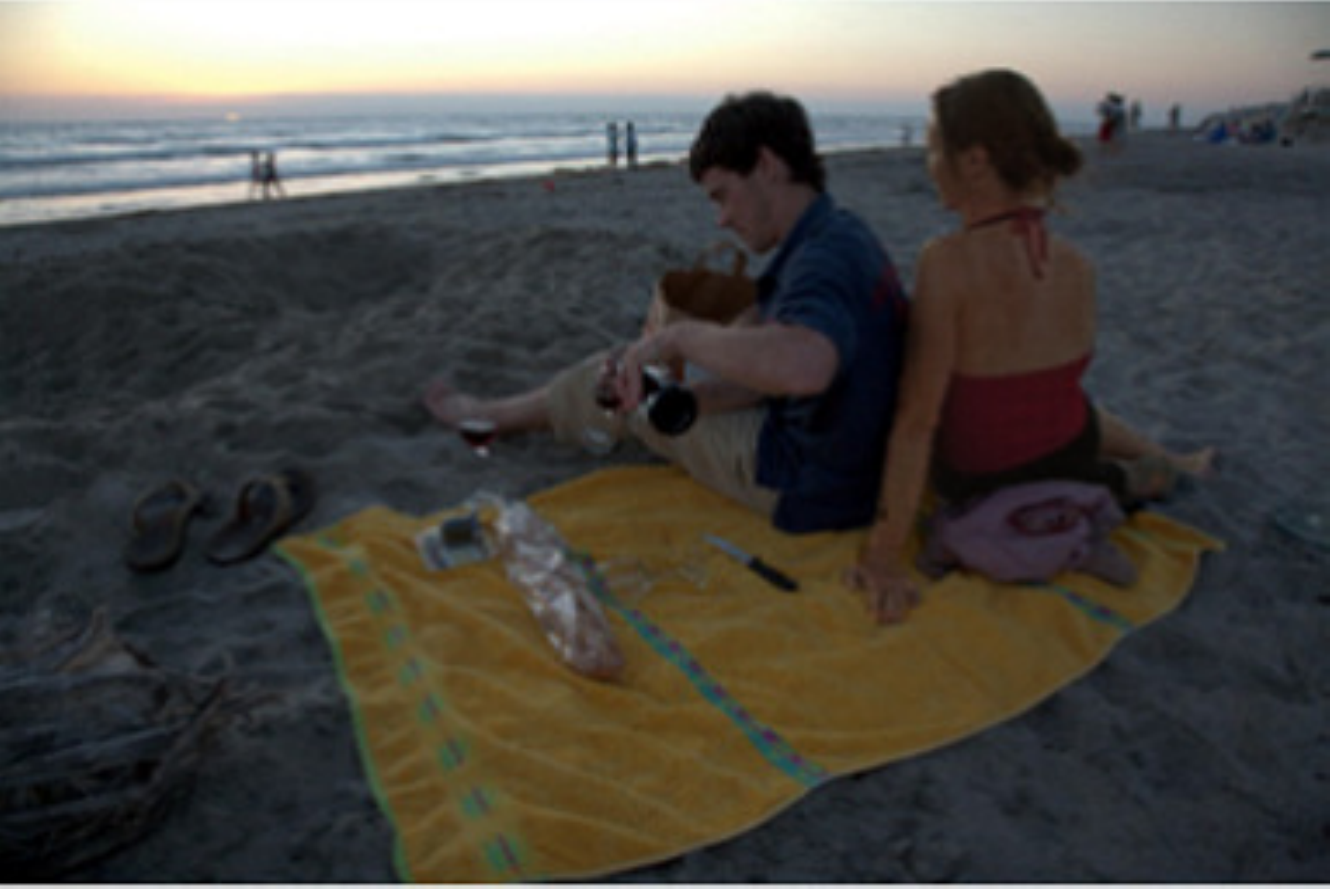}
		\caption*{Ours}
	\end{subfigure}
	\hfil
	\begin{subfigure}{0.16\textwidth}
		% include second image
		\centering
		\includegraphics[width=\linewidth]{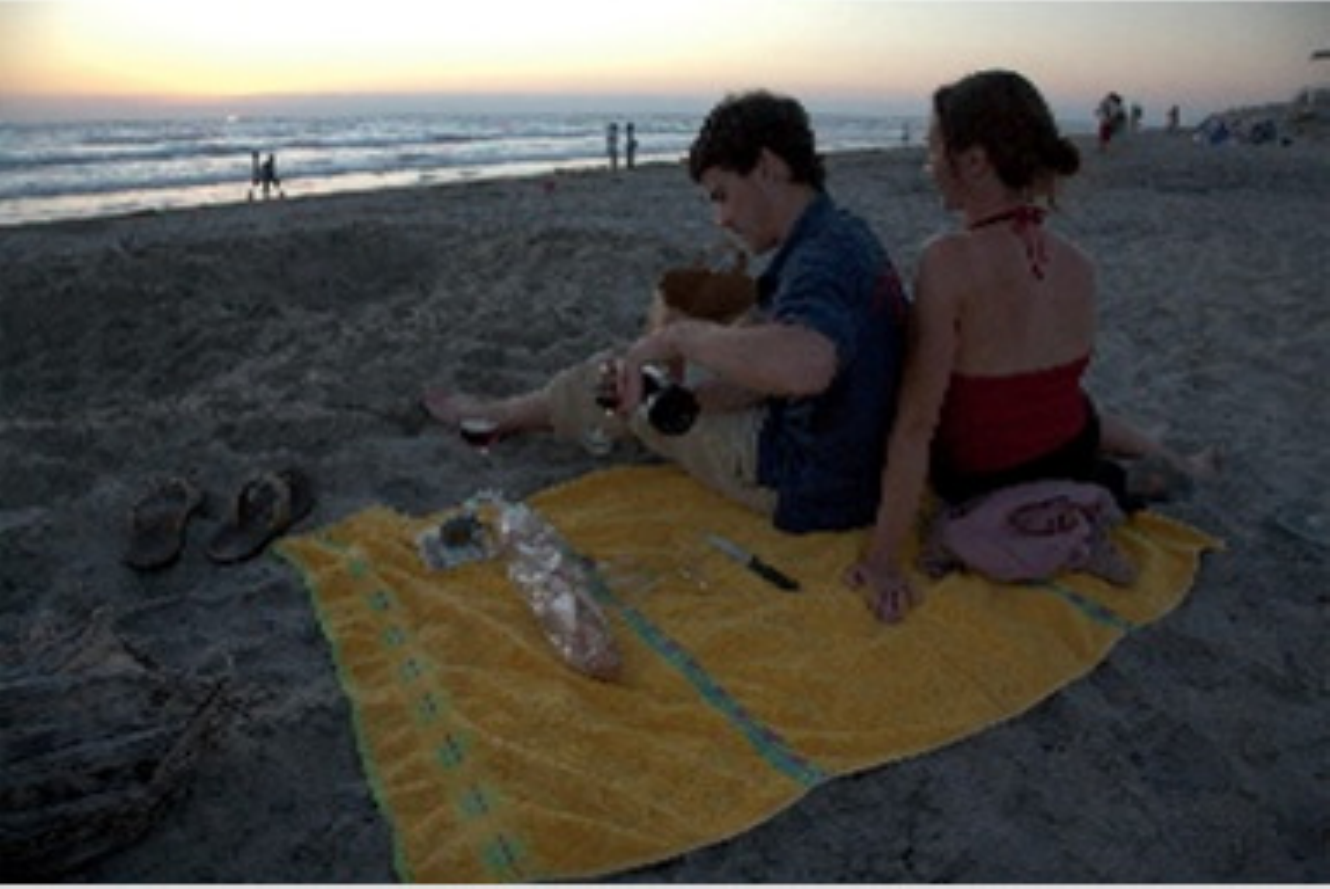}
		\caption*{Real (Ground Truth)}
	\end{subfigure}
	\caption{Qualitative comparison on samples from the testing dataset of iHarmony4 \cite{cong2020dovenet}. Red boxes in composite images mark foreground.}
	\label{result1}
\vspace{-1.0em}
\end{figure*}
\subsection{Experimental Settings}
\paragraph{Datasets.}
Following the same setting as previous methods \cite{cong2020dovenet,ling2021region}, we use the benchmark dataset iHarmony4 \cite{cong2020dovenet} to train and evaluate, which consists of four sub-datasets: HCOCO, HAdobe5k, HFlickr, and Hday2night. We follow the same settings of iHarmony4 as DoveNet \cite{cong2020dovenet}. We also evaluate our method on 99 real composite images released by \cite{tsai2017deep}.
\vspace{-1.0em}
\paragraph{Evaluation Metrics.}
Following \cite{cong2020dovenet,ling2021region,sofiiuk2021foreground,cong2021bargainnet}, the harmonized results are evaluated with Peak Signal-to-Noise
Ratio (PSNR), Mean Squared Error (MSE) and foreground MSE (fMSE) on RGB channels. fMSE is an evaluation metric that only calculates MSE in the foreground region, measuring how well the foreground is harmonized. 
\vspace{-1.0em}
\paragraph{Compared Methods.}
We compare with numerous SOTA image harmonization methods: DIH \cite{tsai2017deep}, DoveNet \cite{cong2020dovenet}, RainNet \cite{ling2021region}, iS$^2$AM \cite{sofiiuk2021foreground}, D-HT \cite{Guo_2021_ICCV}, etc. We do not compare with traditional image harmonization methods since they have
been proven to perform worse than deep learning methods \cite{cong2021bargainnet,ling2021region,cong2020dovenet}. All the results are either provided by the authors, or produced by their officially released codes.
\vspace{-1.0em}
\paragraph{Implementation Details.}
Our model is trained by Adam optimizer with $\beta_{1}=0.9$, $\beta_{2}=0.999$, and $\epsilon=10^{-8}$. We train the model for 120 epochs with input images resized to $256 \times 256$ and batch size set to 12. The initial learning rate is set to $10^{-3}$ and multiply by 0.1 in the 100th and 110th epoch. We use PyTorch to implement our models with Nvidia 2080Ti GPUs. Due to the powerful style representation ability of the VGG network \cite{johnson2016perceptual}, we choose the fixed pretrained VGG-16 \cite{simonyan2014very} as the style representation extrector and use the latent feature at layer \textit{relu4-3}. We set $\lambda=0.01$ in Eq. (\ref{equ:loss}). For the number of negative samples, we set $K=5$, and will be further explored in Section \ref{sec:Ablation}.

\subsection{Comparison with Existing Methods}
\paragraph{Performance on Synthesized iHarmony4 Dataset.}
Table \ref{tab:metric-result} shows the quantitative results of previous state-of-the-art methods as well as our method. From Table \ref{tab:metric-result}, we can observe that our method outperforms other compared methods on all datasets, except for MSE and fMSE value on Hday2night. Moreover, compared with the second best method, our method achieves a huge average performance gain of 0.56 dB in PSNR, 3.11 in MSE, and 16.1 in fMSE.

\begin{figure}[htbp]
	\centering
	\begin{subfigure}{0.19\linewidth}
		% include first image
		\centering
		\includegraphics[width=\linewidth]{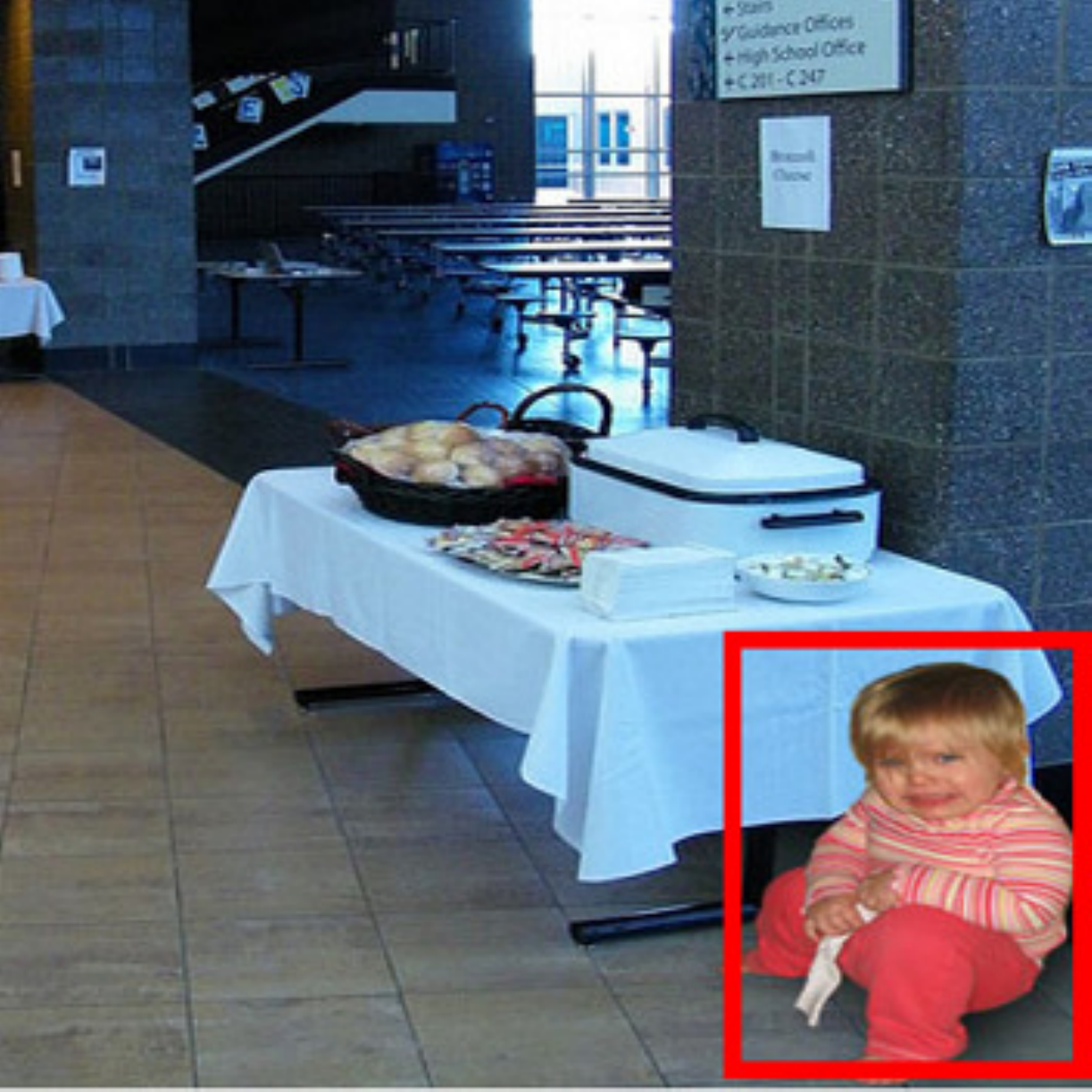}
	\end{subfigure}
	\hfil
	\begin{subfigure}{0.19\linewidth}
		% include second image
		\centering
		\includegraphics[width=\linewidth]{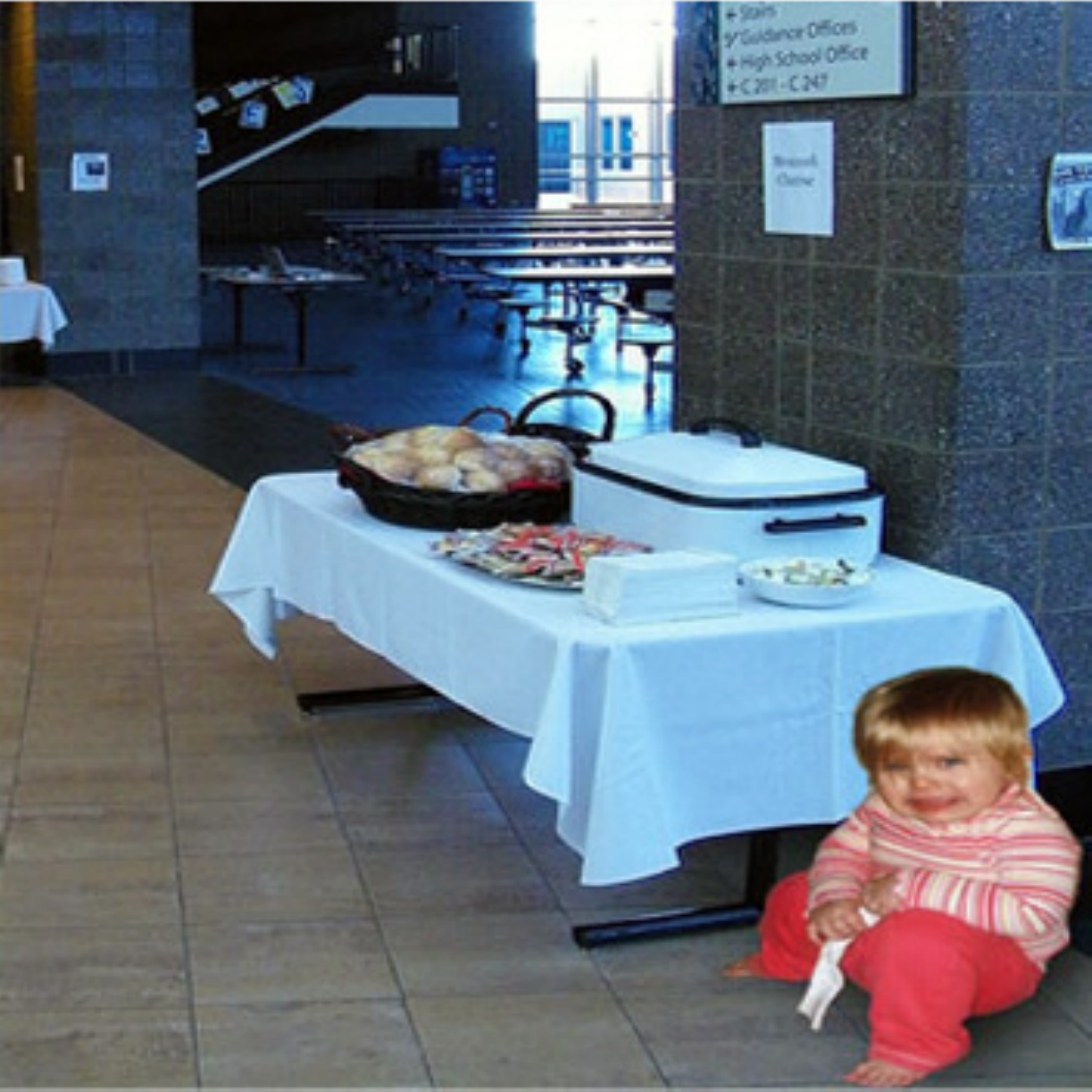}
	\end{subfigure}
	\hfil
	\begin{subfigure}{0.19\linewidth}
		% include second image
		\centering
		\includegraphics[width=\linewidth]{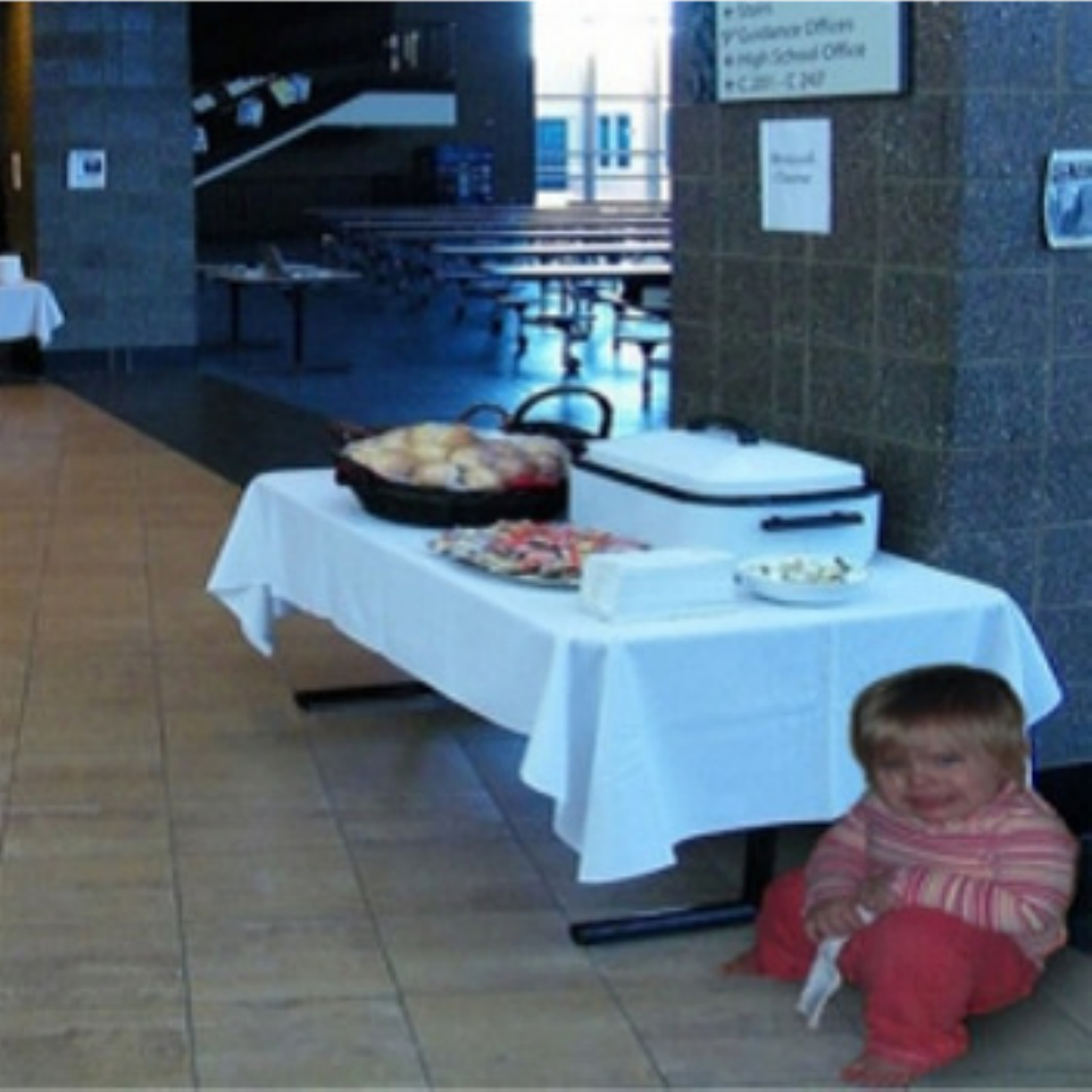}
	\end{subfigure}
	\hfil
	\begin{subfigure}{0.19\linewidth}
		% include first image
		\centering
		\includegraphics[width=\linewidth]{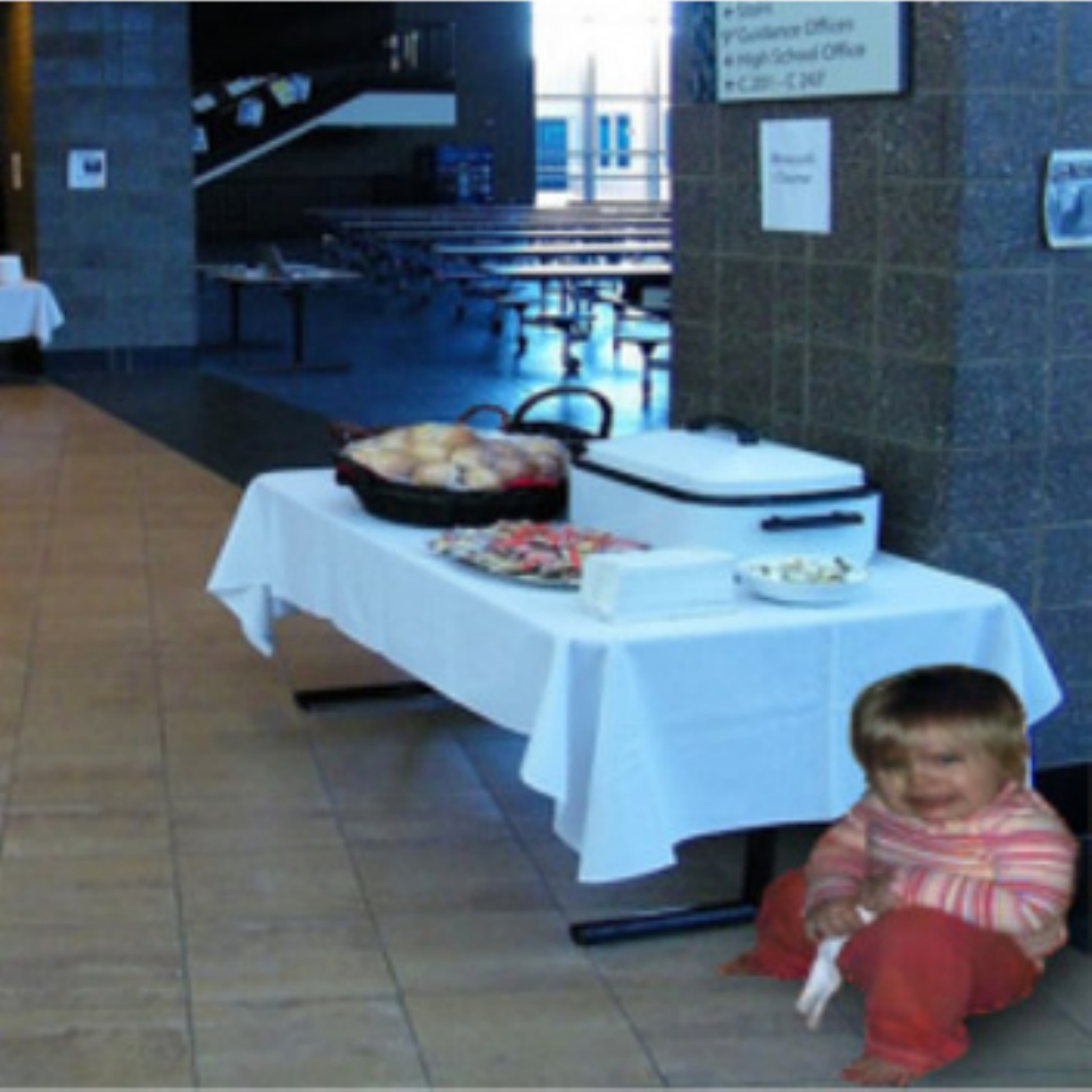}
	\end{subfigure}
	\hfil
	\begin{subfigure}{0.19\linewidth}
		% include second image
		\centering
		\includegraphics[width=\linewidth]{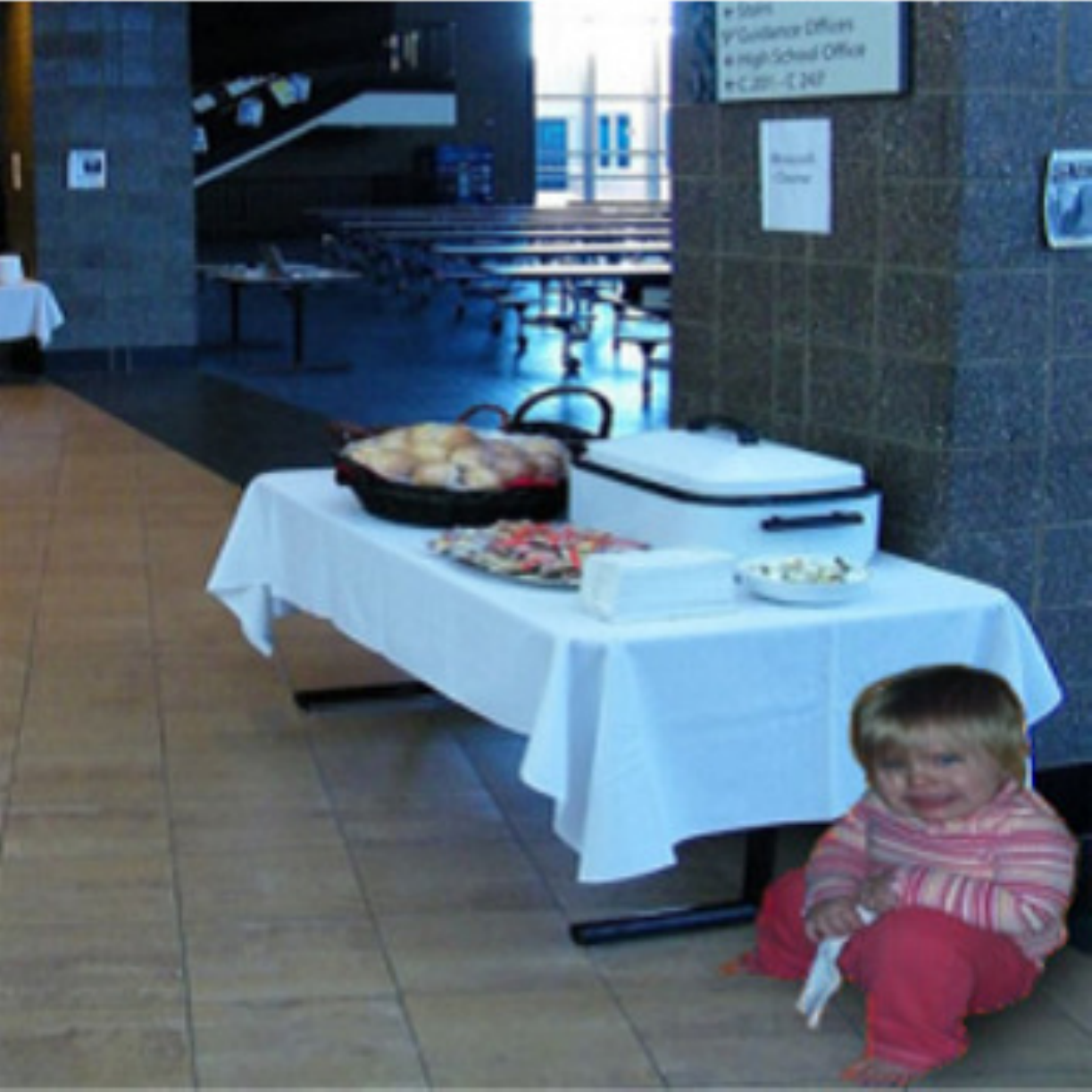}
	\end{subfigure}
	\quad
	\begin{subfigure}{0.19\linewidth}
		% include second image
		\centering
		\includegraphics[width=\linewidth]{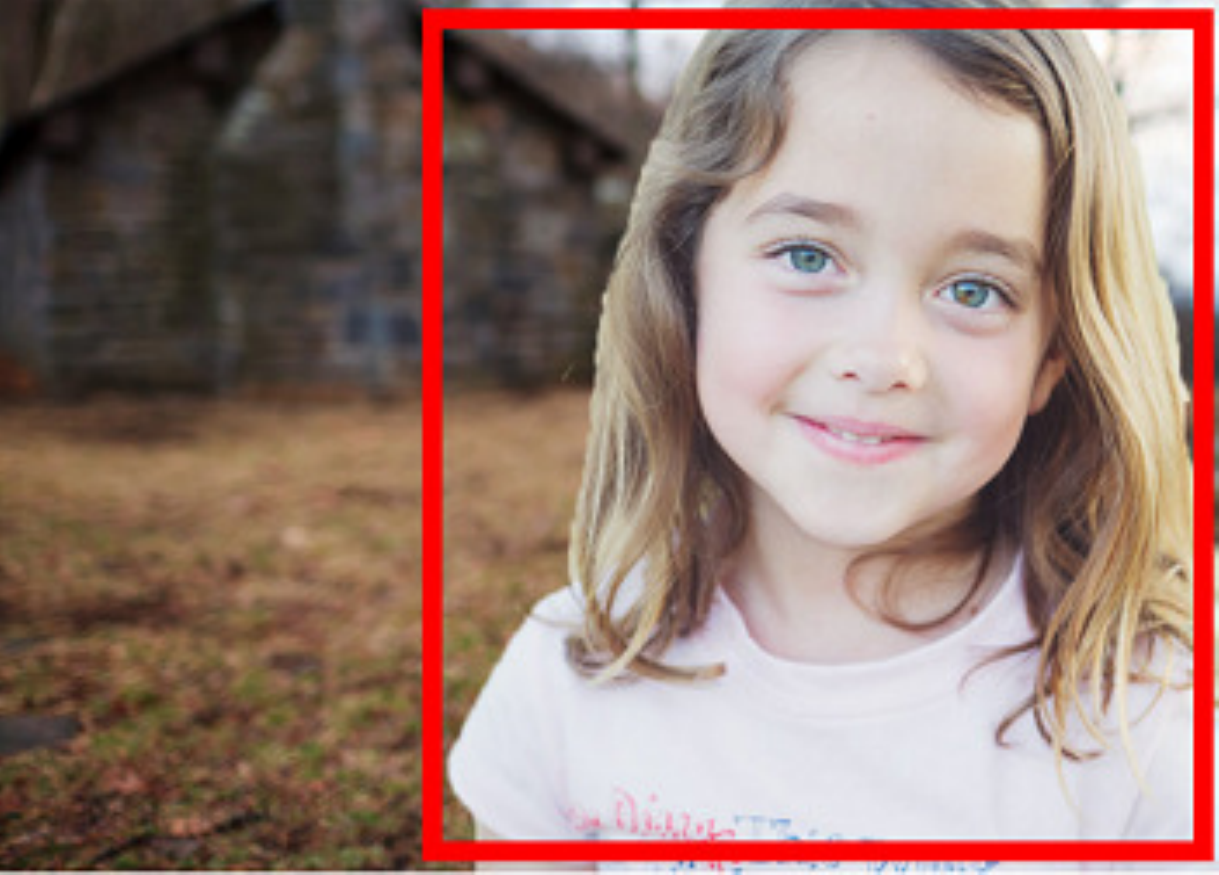}
	\end{subfigure}
	\hfil
	\begin{subfigure}{0.19\linewidth}
		% include first image
		\centering
		\includegraphics[width=\linewidth]{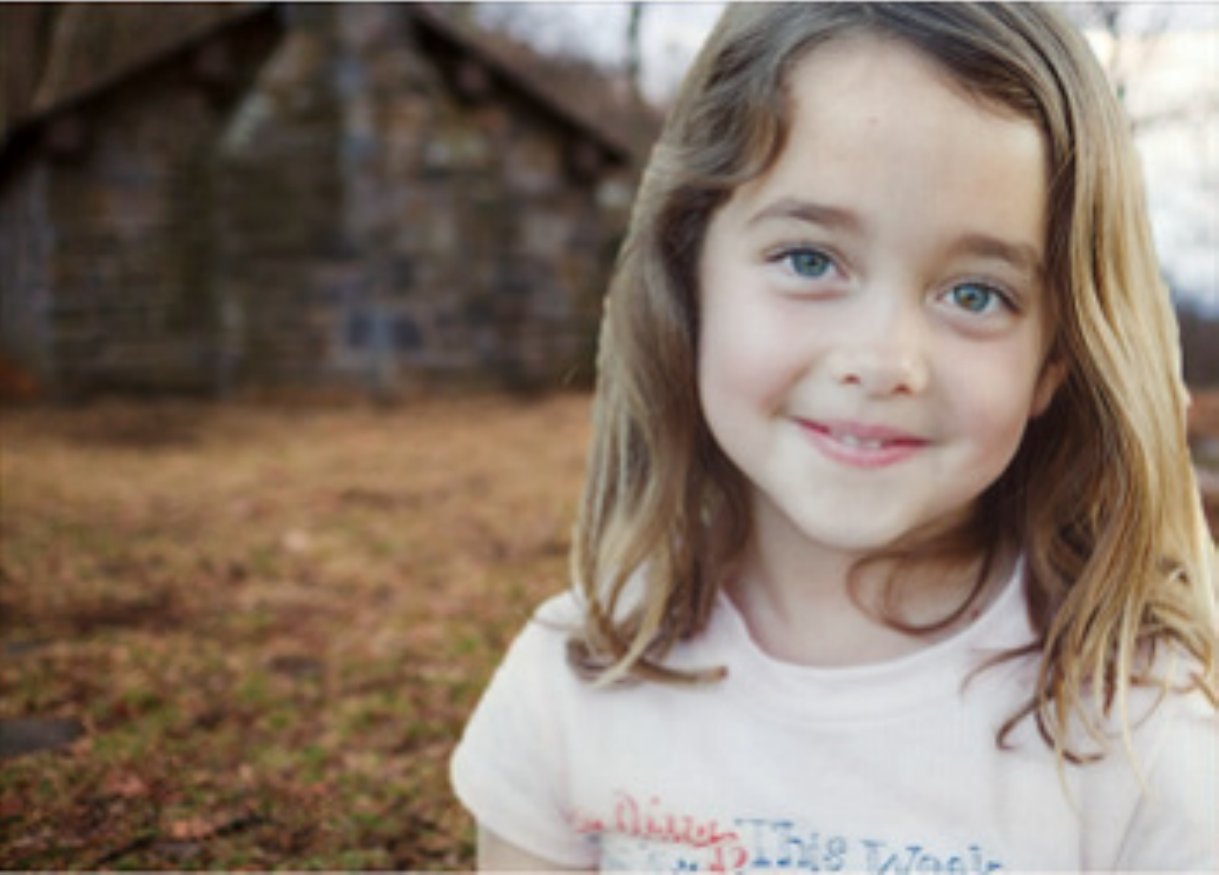}
	\end{subfigure}
	\hfil
	\begin{subfigure}{0.19\linewidth}
		% include second image
		\centering
		\includegraphics[width=\linewidth]{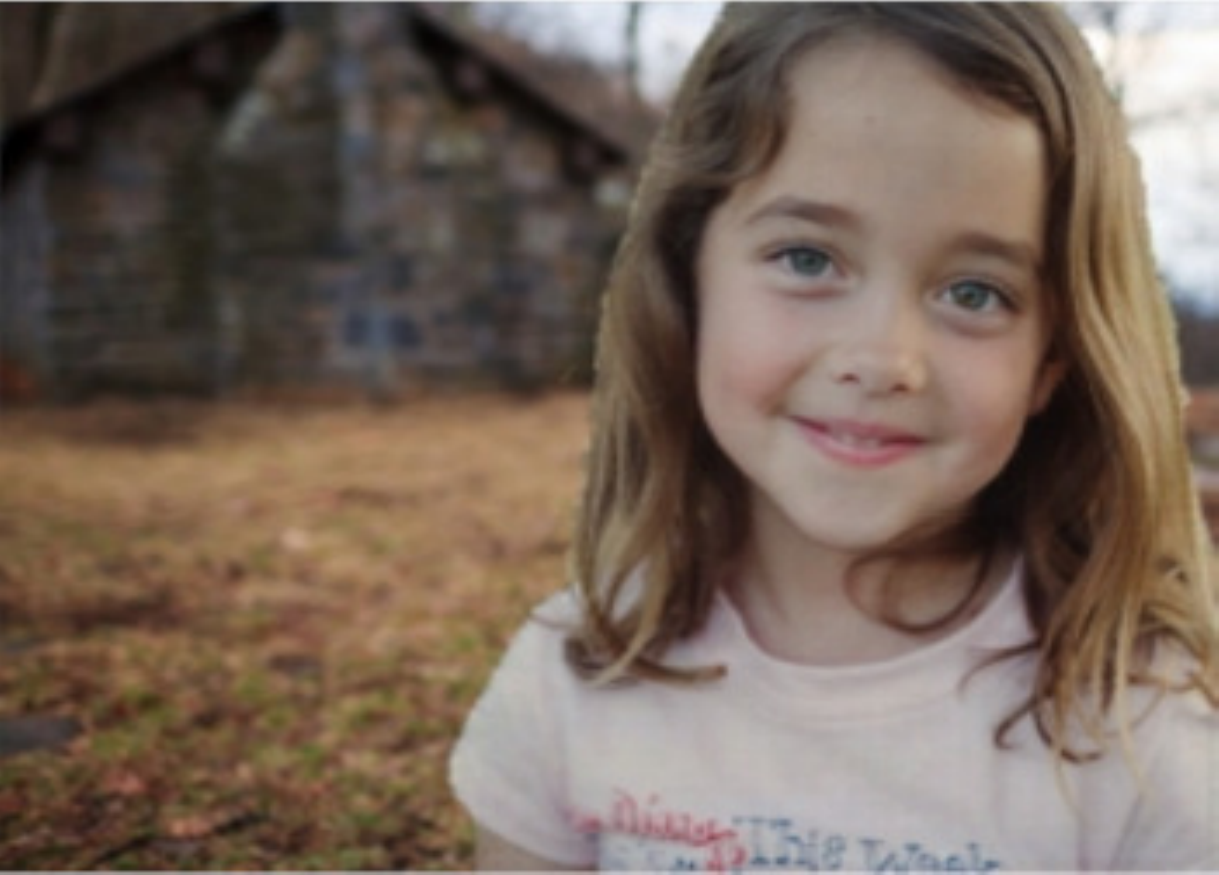}
	\end{subfigure}
	\hfil
	\begin{subfigure}{0.19\linewidth}
		% include second image
		\centering
		\includegraphics[width=\linewidth]{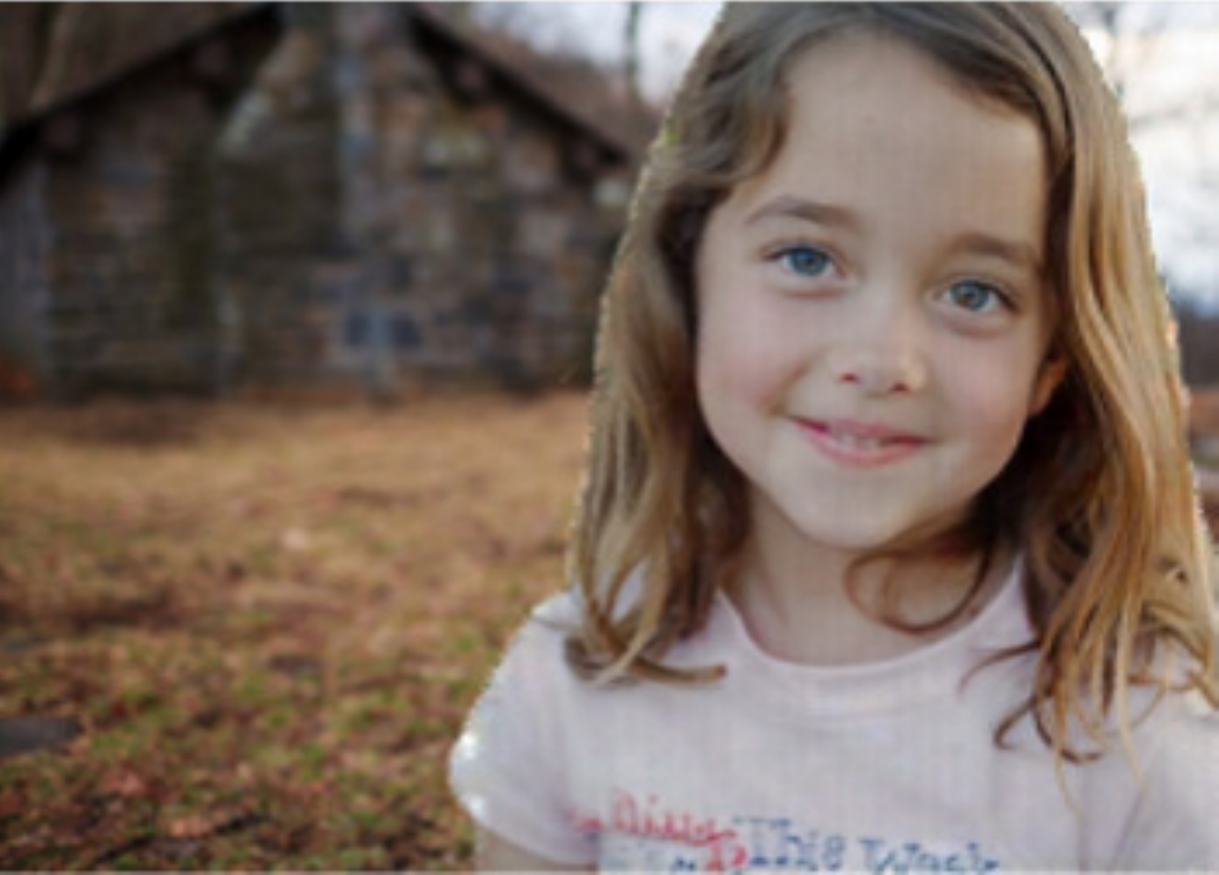}
	\end{subfigure}
	\hfil
	\begin{subfigure}{0.19\linewidth}
		% include first image
		\centering
		\includegraphics[width=\linewidth]{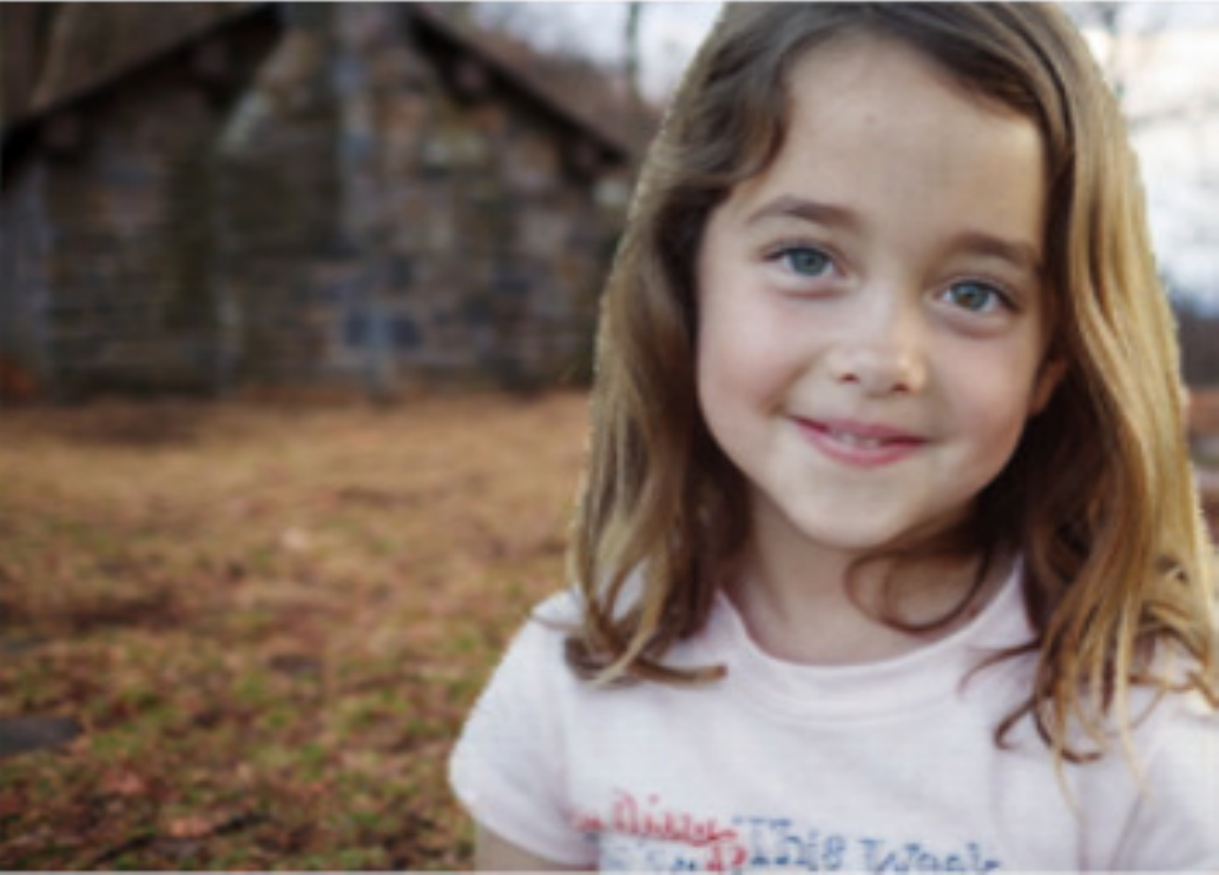}
	\end{subfigure}
	\quad
	\begin{subfigure}{0.19\linewidth}
		% include second image
		\centering
		\includegraphics[width=\linewidth]{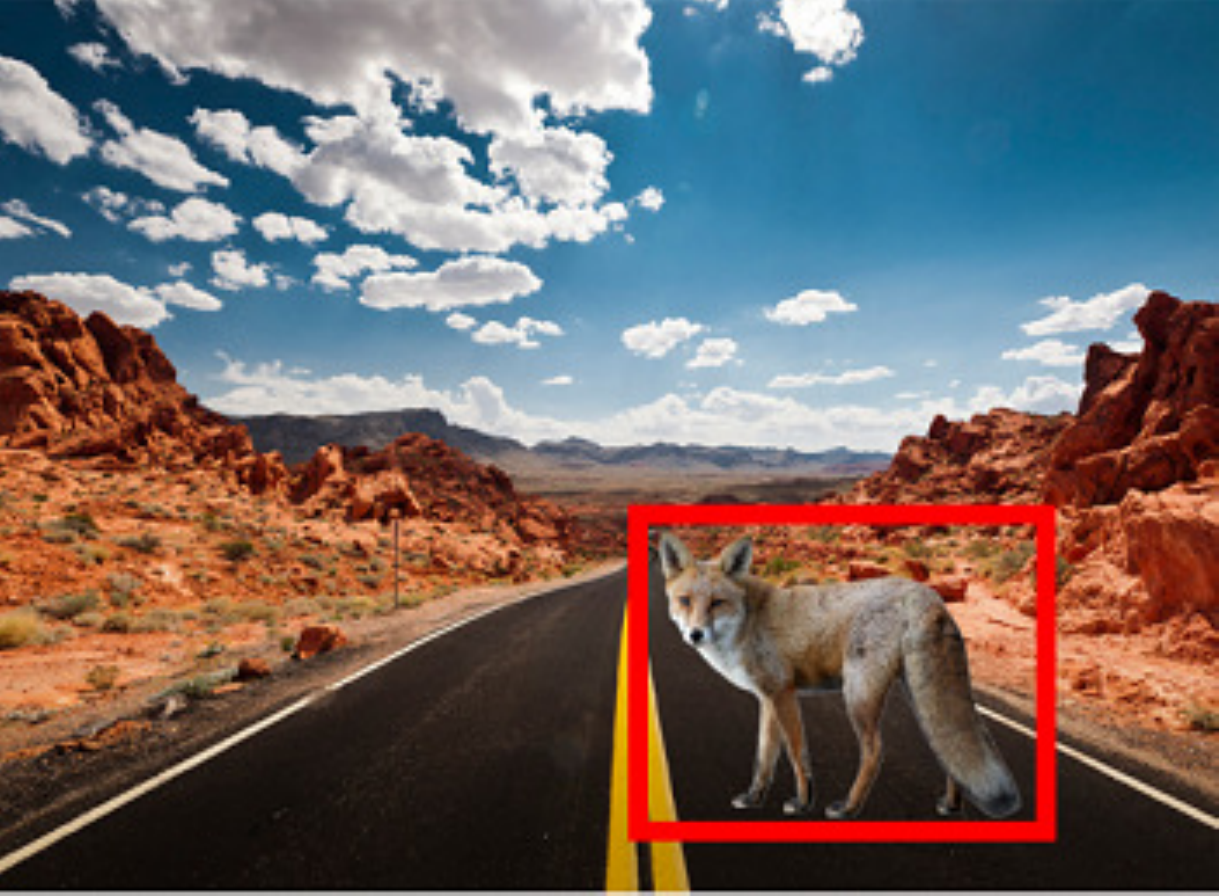}
		\caption*{Composite}
	\end{subfigure}
	\hfil
	\begin{subfigure}{0.19\linewidth}
		% include second image
		\centering
		\includegraphics[width=\linewidth]{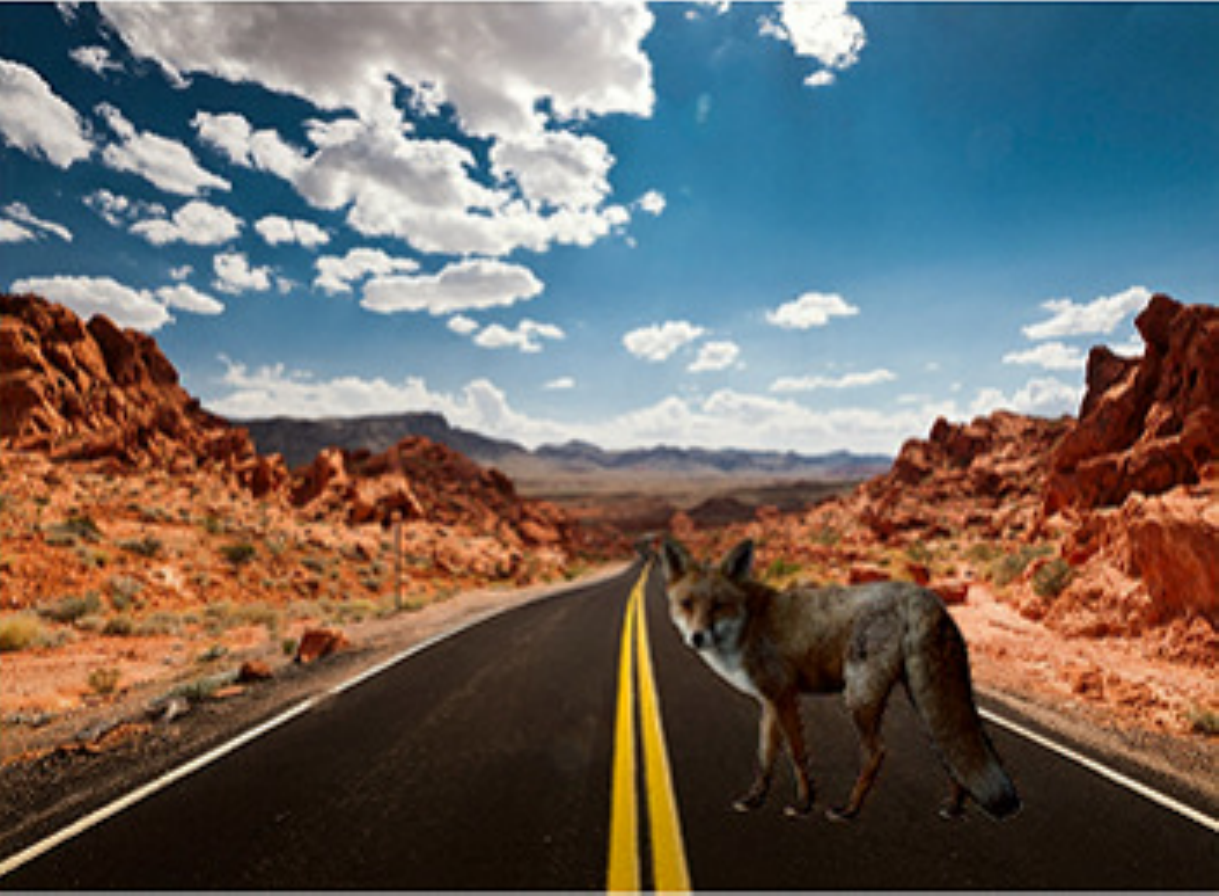}
		\caption*{DIH \cite{tsai2017deep}}
	\end{subfigure}
	\hfil
	\begin{subfigure}{0.19\linewidth}
		% include first image
		\centering
		\includegraphics[width=\linewidth]{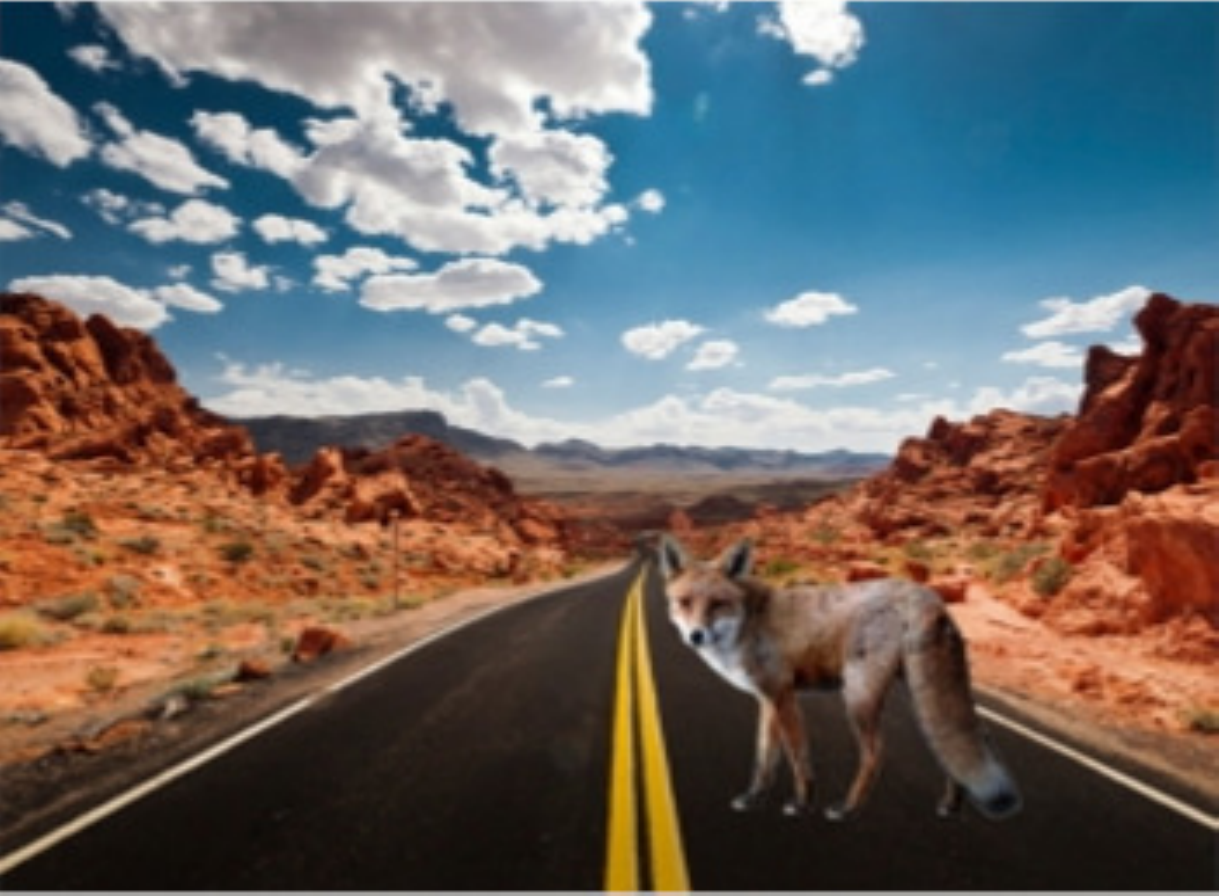}
		\caption*{DoveNet \cite{cong2020dovenet}}
	\end{subfigure}
	\hfil
	\begin{subfigure}{0.19\linewidth}
		% include second image
		\centering
		\includegraphics[width=\linewidth]{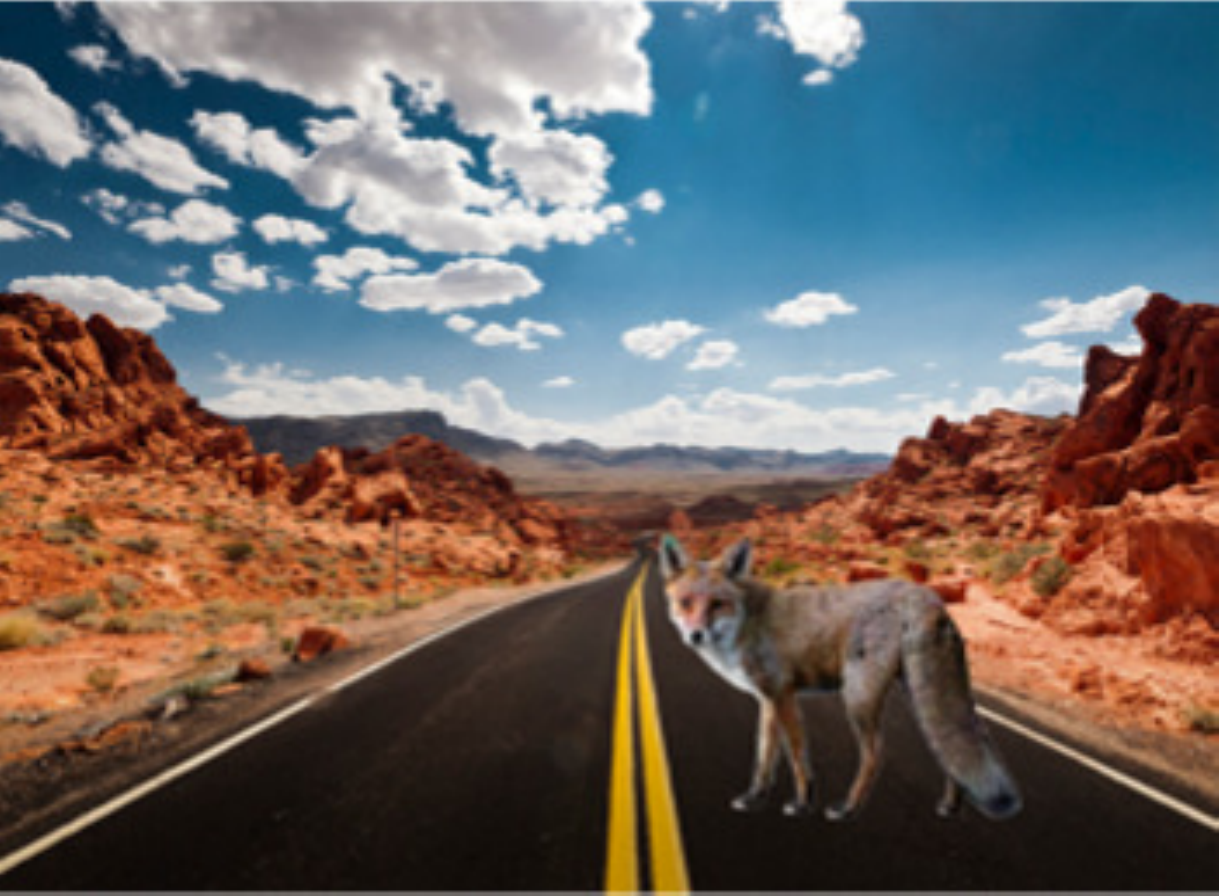}
		\caption*{RainNet \cite{ling2021region}}
	\end{subfigure}
	\hfil
	\begin{subfigure}{0.19\linewidth}
		% include second image
		\centering
		\includegraphics[width=\linewidth]{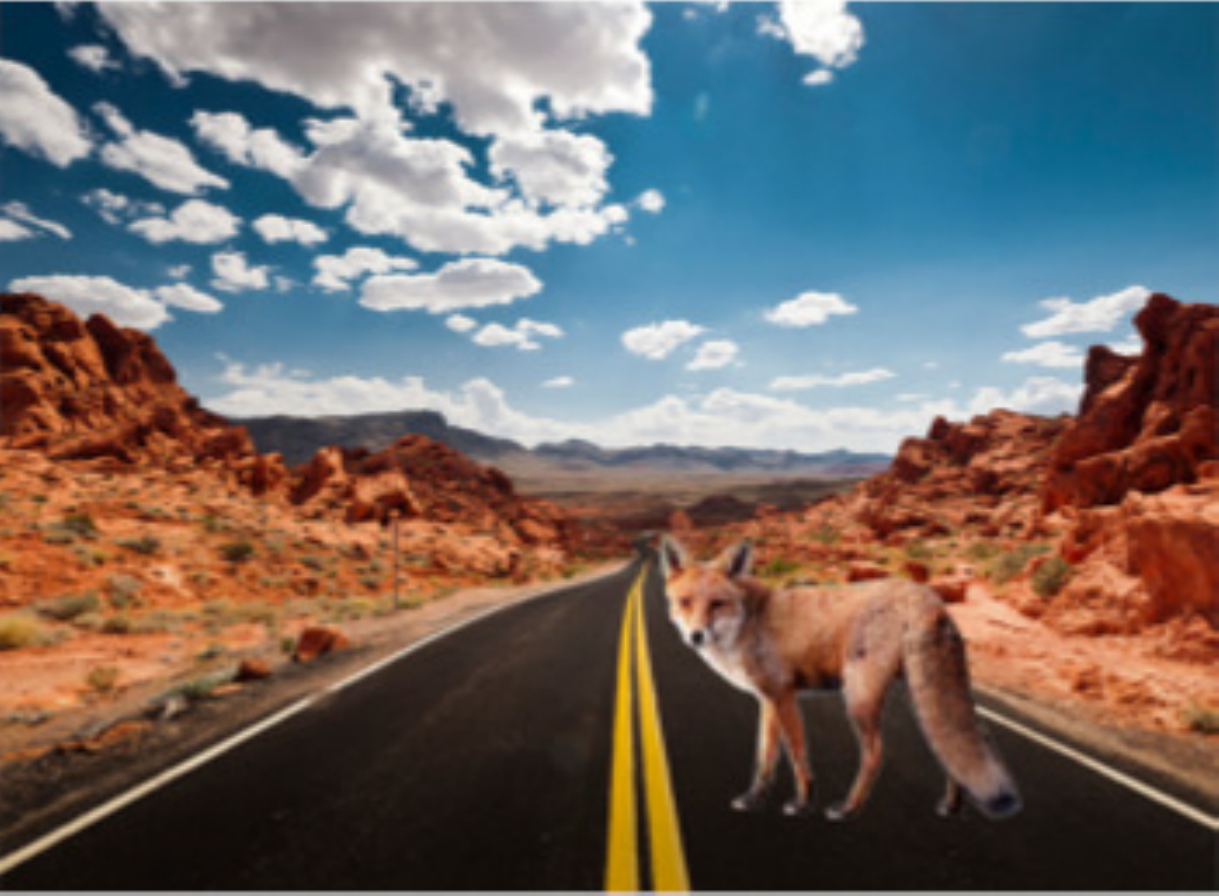}
		\caption*{Ours}
	\end{subfigure}
	\caption{Example results on real composite images taken from \cite{tsai2017deep}. Red boxes in composite images mark foreground.}
	\label{result2}
	\vspace{-1.0em}
\end{figure}

In Figure \ref{result1}, we further present qualitative comparison results on iHarmony4. It can be easily observed that our method obtains a more consistent visual style in the whole composite image, achieving a more photorealistic output. For example, as shown in the third row of Figure \ref{result1}, the visual style of the foreground and the background are quite different, resulting in obvious image distortion. The other three methods cannot adjust the style of the foreground, especially the overall tone and the contrast of lighting and shadows. Unlike them, our method produces a more photorealistic result and is closer to the ground-truth real image.
\vspace{-2.3em}
\paragraph{Performance on Real Composite Images.}
Figure \ref{result2} presents some results on real composite images released by DIH \cite{tsai2017deep}. Because there is no ground truth image as a reference, it is impossible to compare different methods quantitatively using PSNR, MSE or fMSE. However, we can still find that our method achieves the best visual effect. Please refer to supplementary for more visual comparison.

We further conduct a user study. Following \cite{cong2020dovenet,cong2021bargainnet,guo2021intrinsic}, we invite 60 volunteers and acquire a total of 29700 pairwise results for all 99 images, with 30 results for each pair of different methods on average. Then, we use the Bradley-Terry model (B-T model) \cite{bradley1952rank,lai2016comparative} to calculate the global ranking score for each method. Table \ref{tab:BT} demonstrates that our method achieves the highest B-T score, which proves its effectiveness in real-world applications.
\begin{table}[htbp]
	\centering
	\caption{B-T scores comparison on 99 real composite images.}
	\resizebox{\linewidth}{!}{
		\begin{tabular}{cccccc}
			\toprule
			Method & Composite & DIH \cite{tsai2017deep}   & DoveNet \cite{cong2020dovenet} & RainNet \cite{ling2021region} & Ours \\
			\midrule
			B-T score$\uparrow$ & 0.574 & 0.889 & 1.075 & 1.213 & \textbf{1.841} \\
			\bottomrule
	\end{tabular}}%
	\label{tab:BT}%
\end{table}%

\subsection{Ablation Studies}
\label{sec:Ablation}
\begin{figure}[htbp]
	\centering
	\begin{subfigure}{0.19\linewidth}
		% include first image
		\centering
		\includegraphics[width=\linewidth]{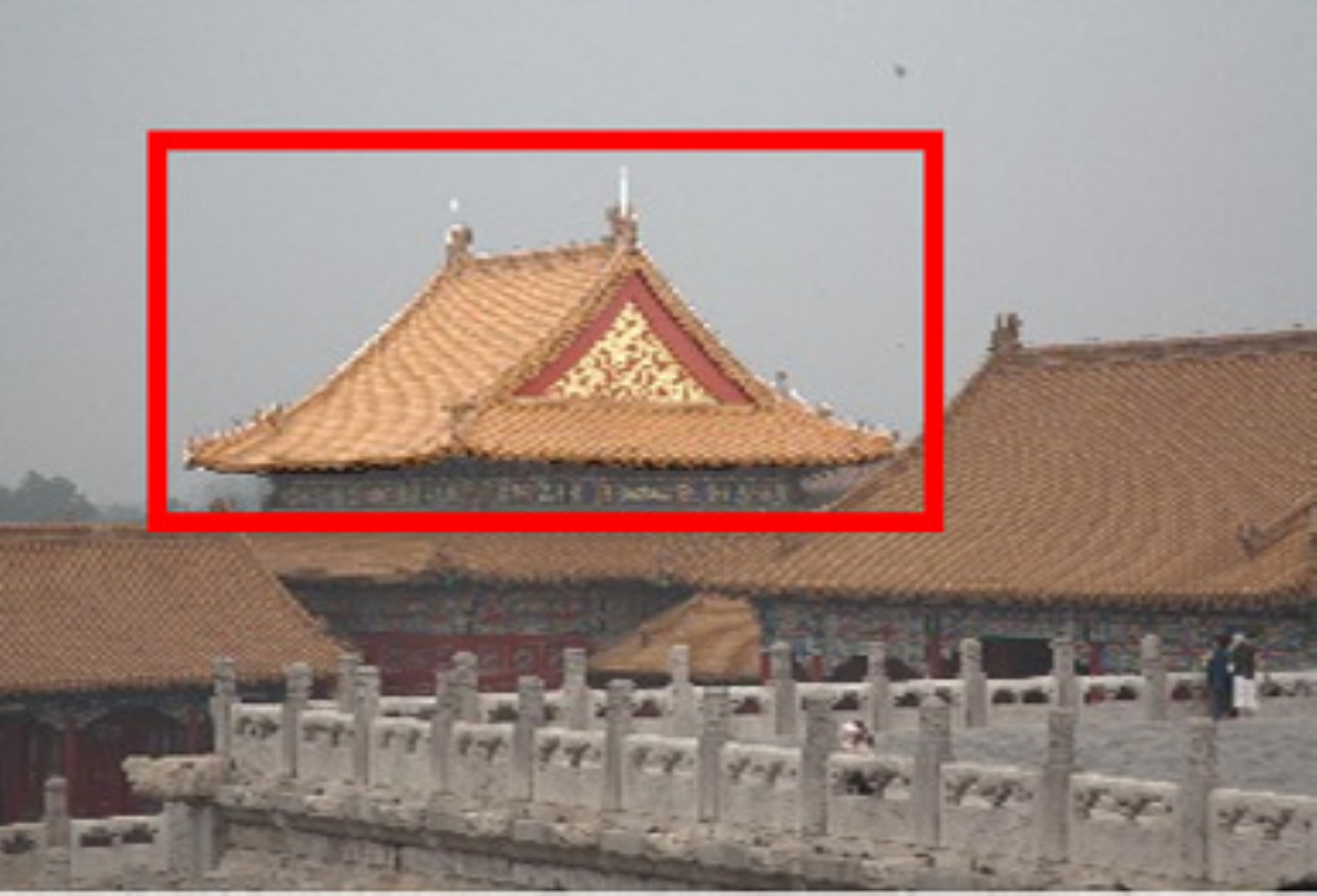}
	\end{subfigure}
	\hfil
	\begin{subfigure}{0.19\linewidth}
		% include second image
		\centering
		\includegraphics[width=\linewidth]{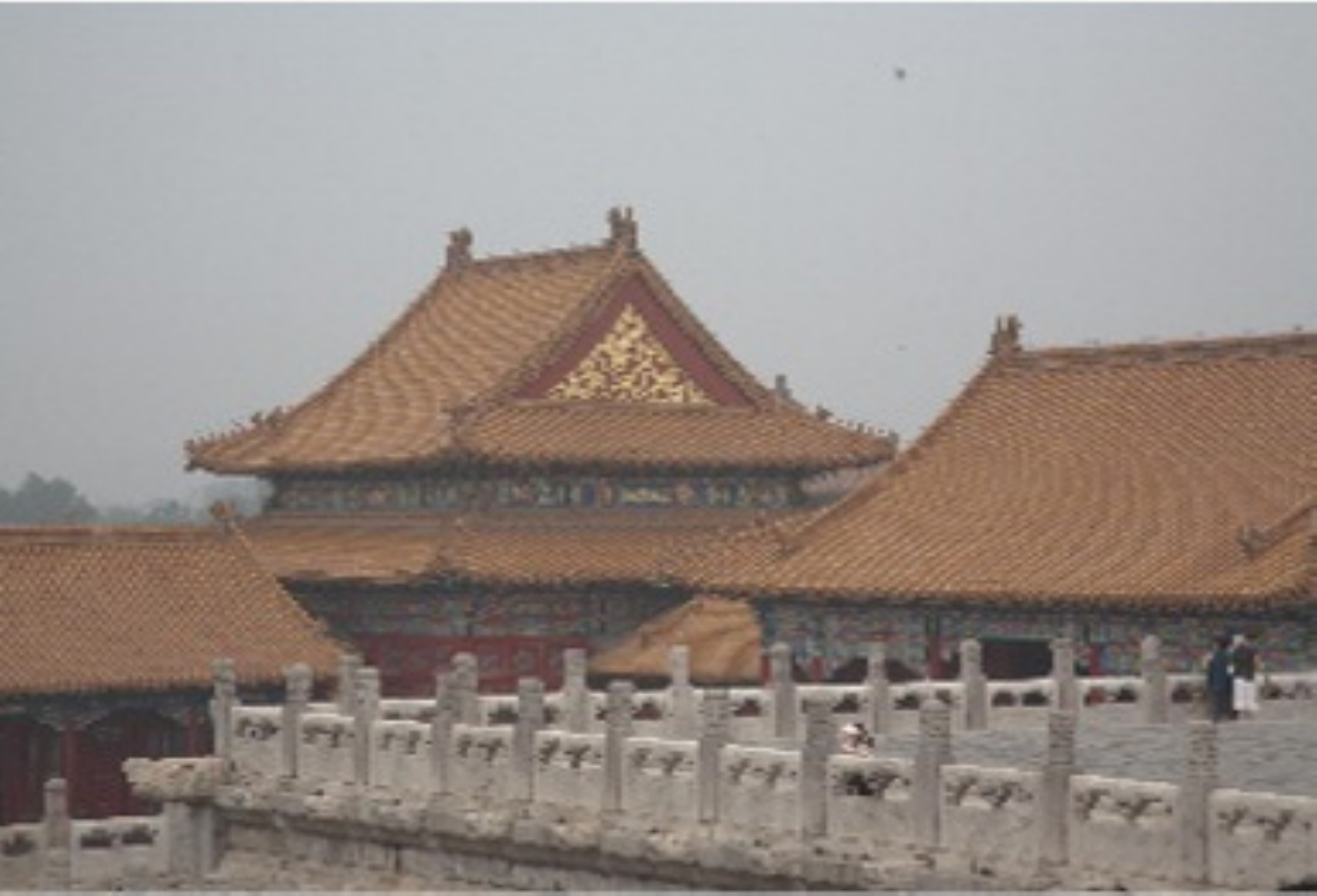}
	\end{subfigure}
	\hfil
	\begin{subfigure}{0.19\linewidth}
		% include second image
		\centering
		\includegraphics[width=\linewidth]{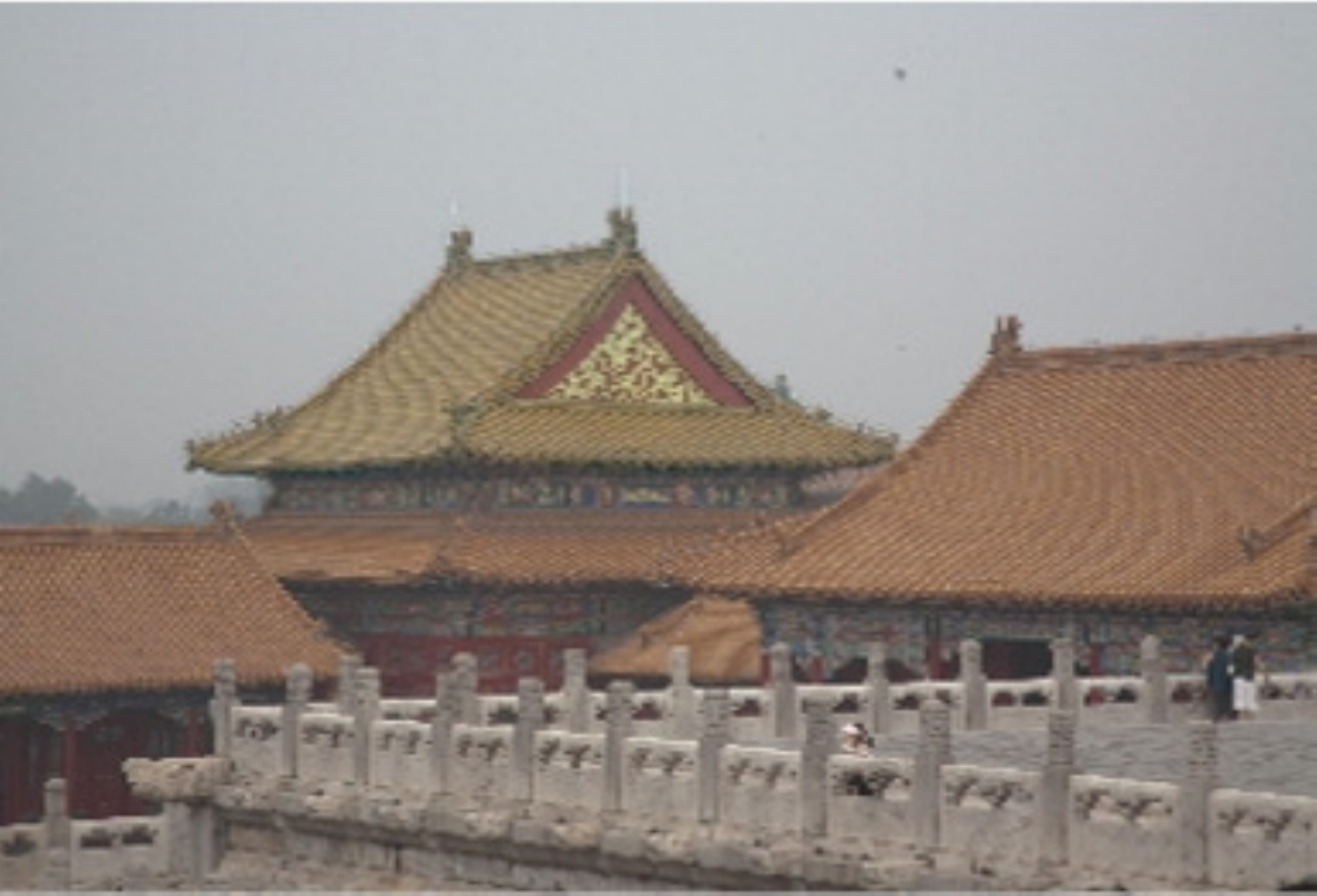}
	\end{subfigure}
	\hfil
	\begin{subfigure}{0.19\linewidth}
		% include first image
		\centering
		\includegraphics[width=\linewidth]{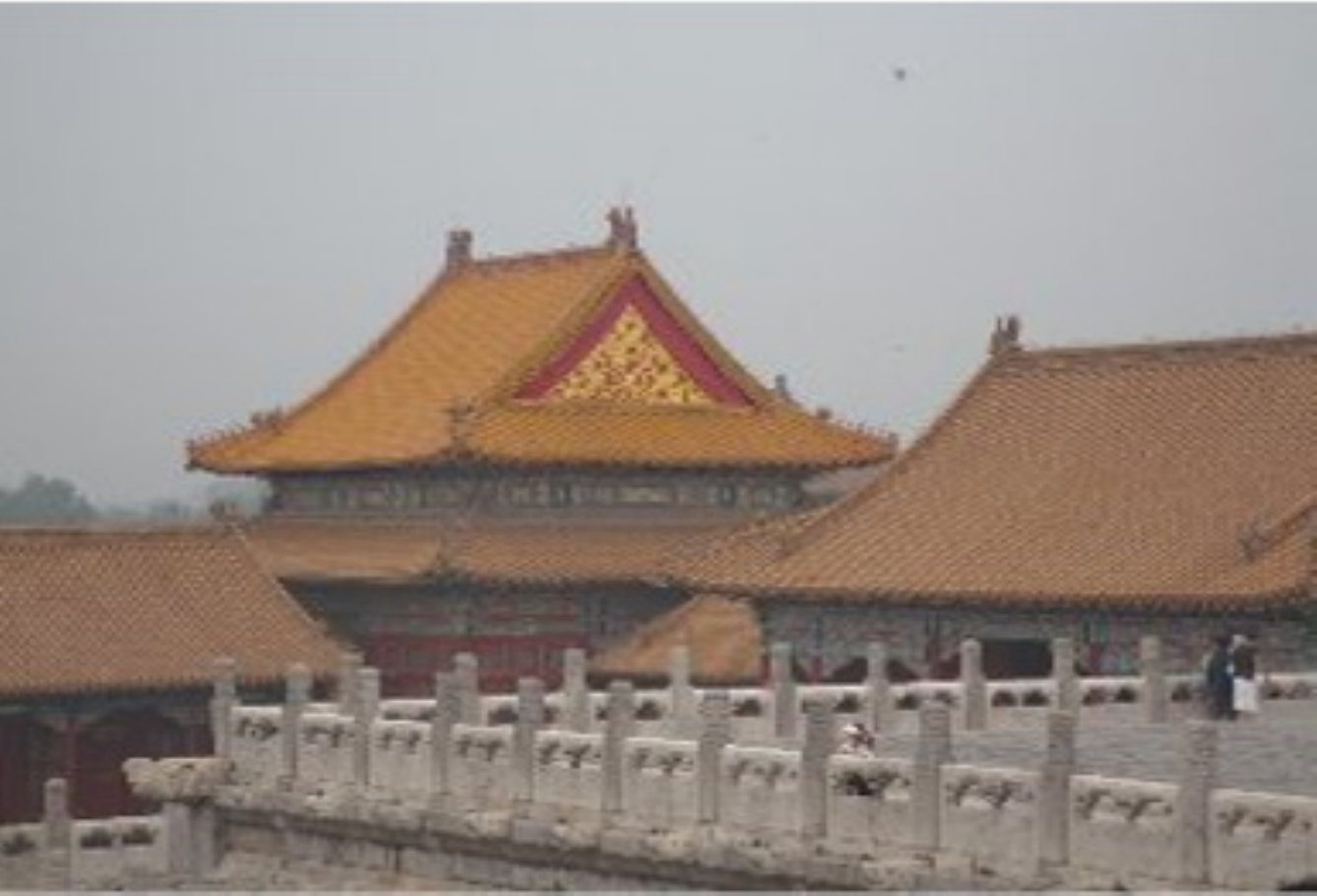}
	\end{subfigure}
	\hfil
	\begin{subfigure}{0.19\linewidth}
		% include second image
		\centering
		\includegraphics[width=\linewidth]{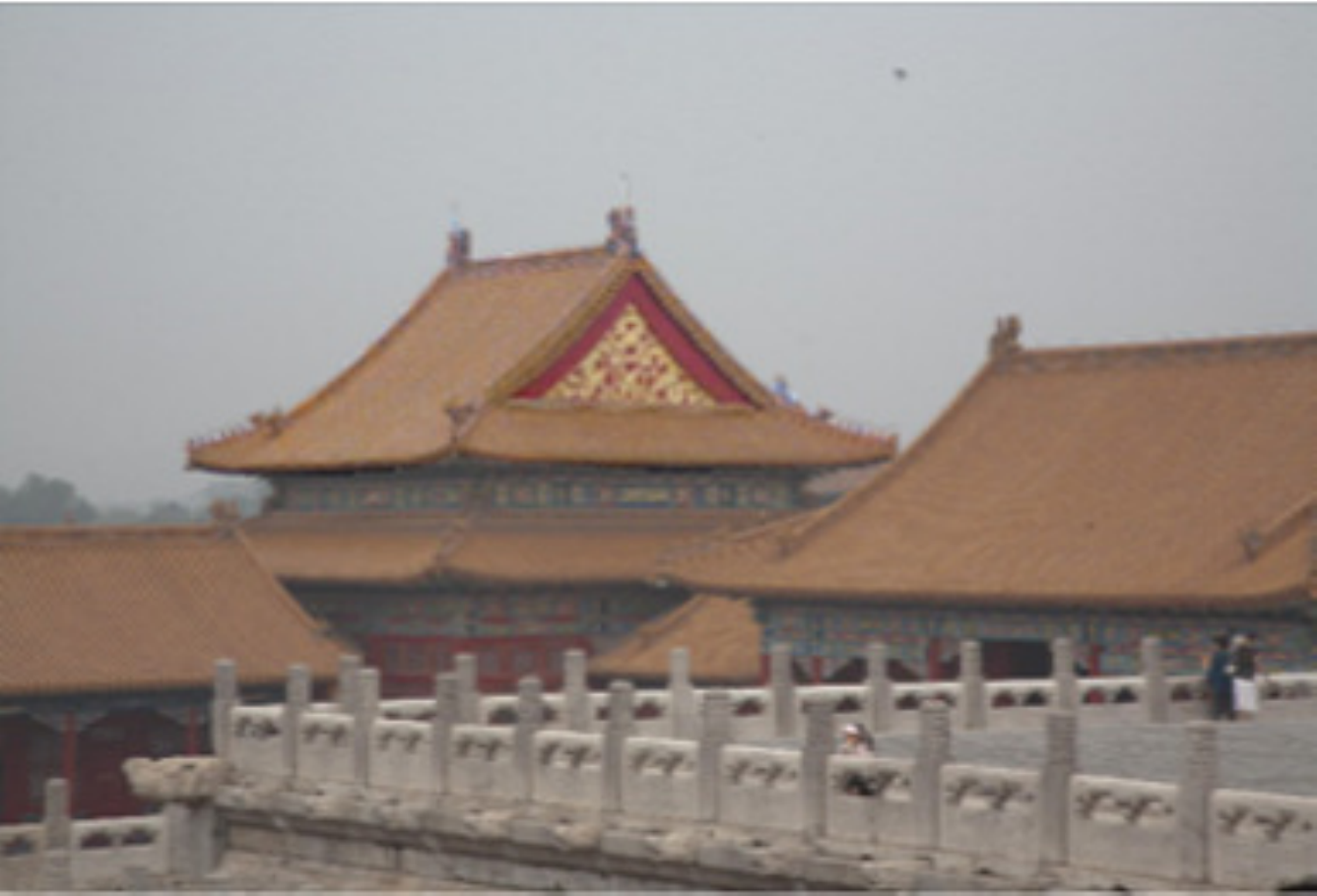}
	\end{subfigure}
	\quad
	\begin{subfigure}{0.19\linewidth}
		% include second image
		\centering
		\includegraphics[width=\linewidth]{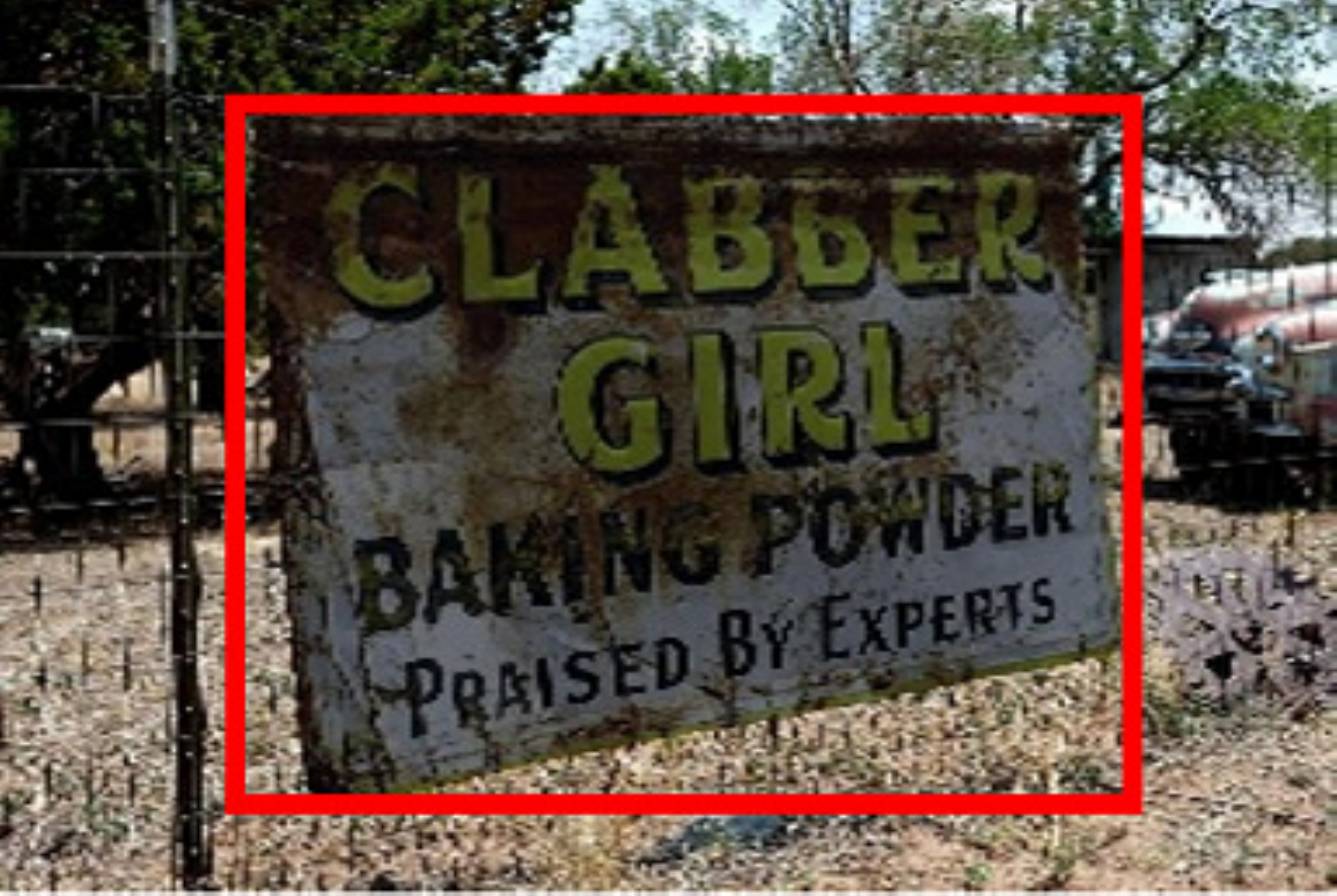}
		\caption*{Composite}
	\end{subfigure}
	\hfil
	\begin{subfigure}{0.19\linewidth}
		% include first image
		\centering
		\includegraphics[width=\linewidth]{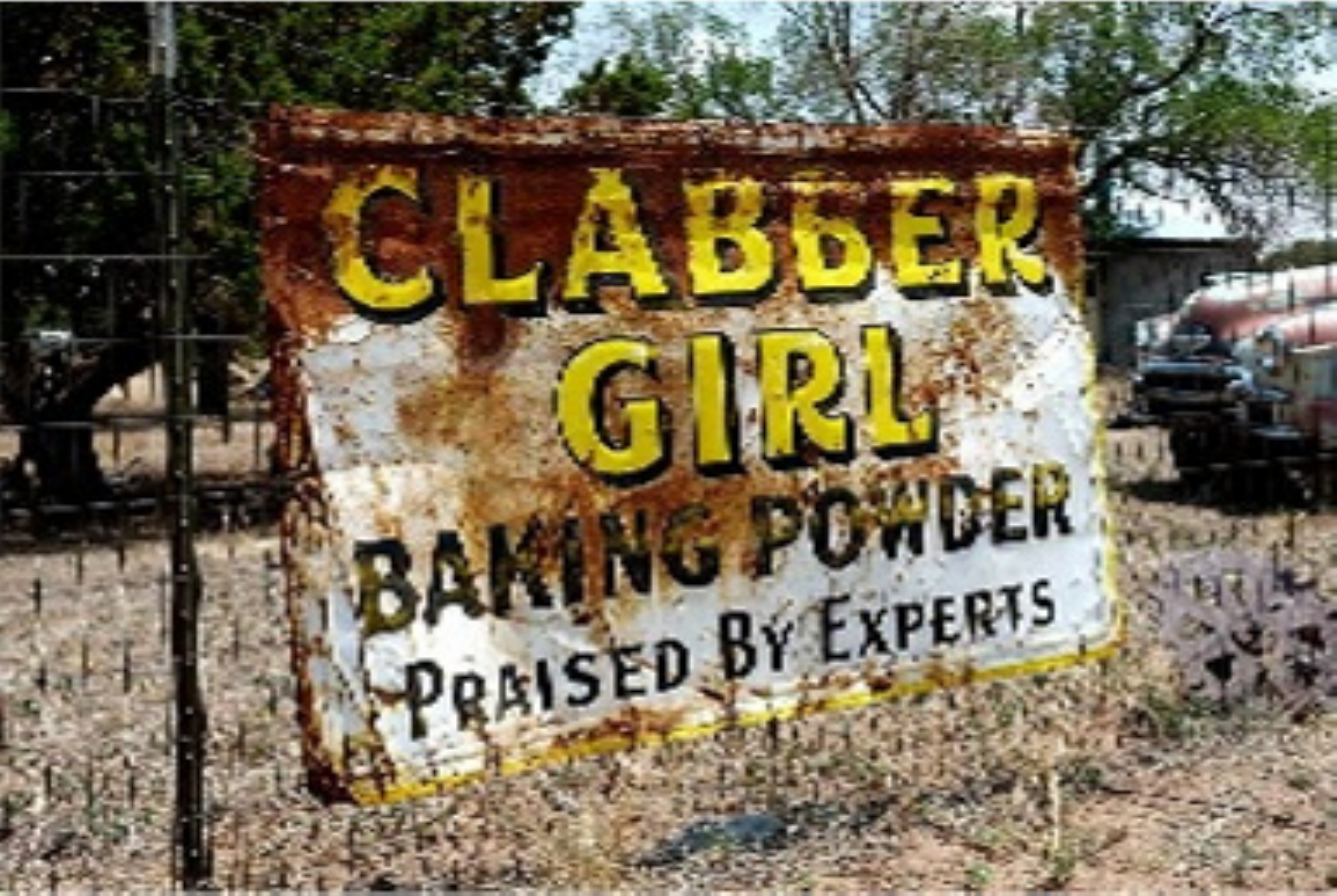}
		\caption*{Real}
	\end{subfigure}
	\hfil
	\begin{subfigure}{0.19\linewidth}
		% include second image
		\centering
		\includegraphics[width=\linewidth]{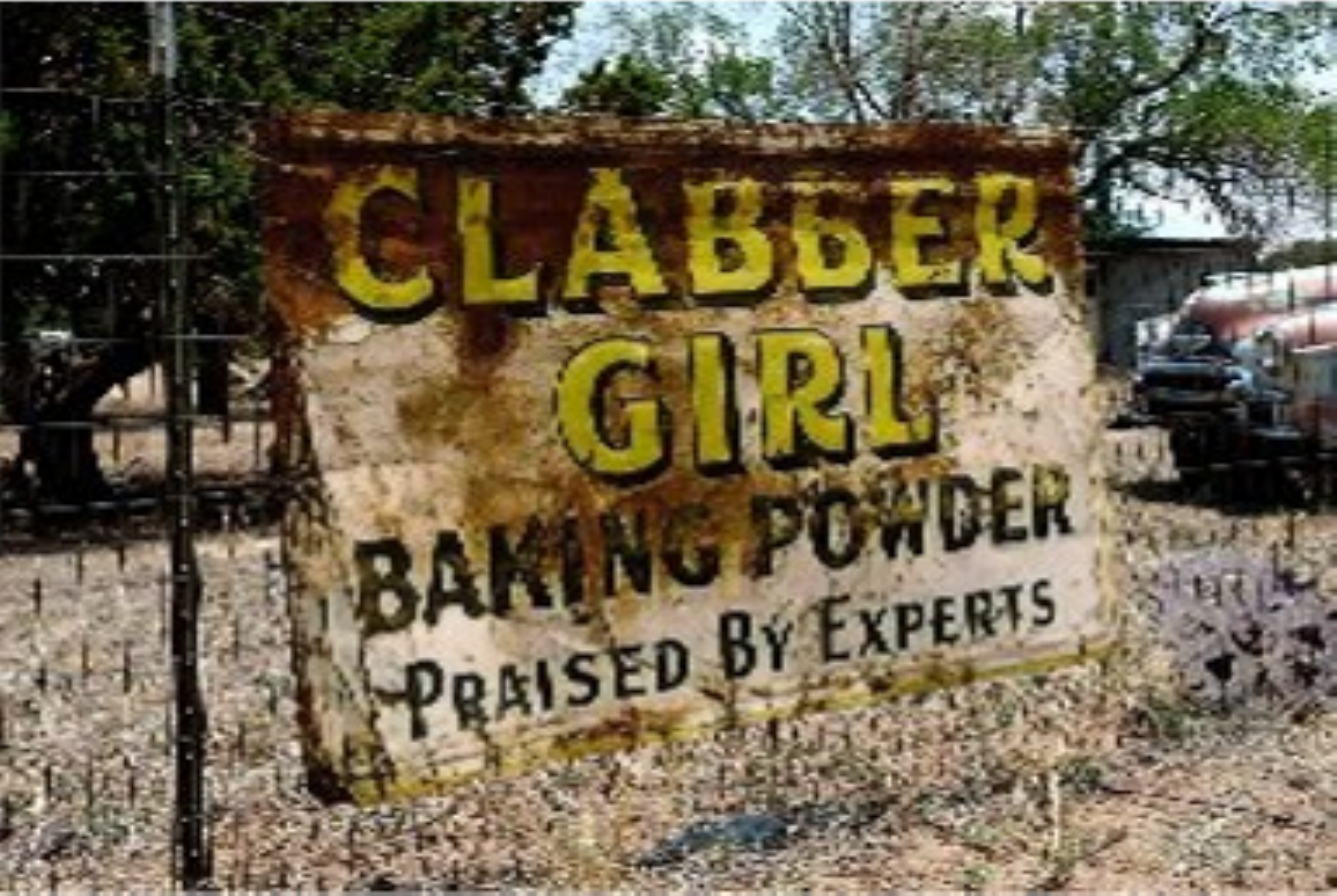}
		\caption*{Baseline}
	\end{subfigure}
	\hfil
	\begin{subfigure}{0.19\linewidth}
		% include second image
		\centering
		\includegraphics[width=\linewidth]{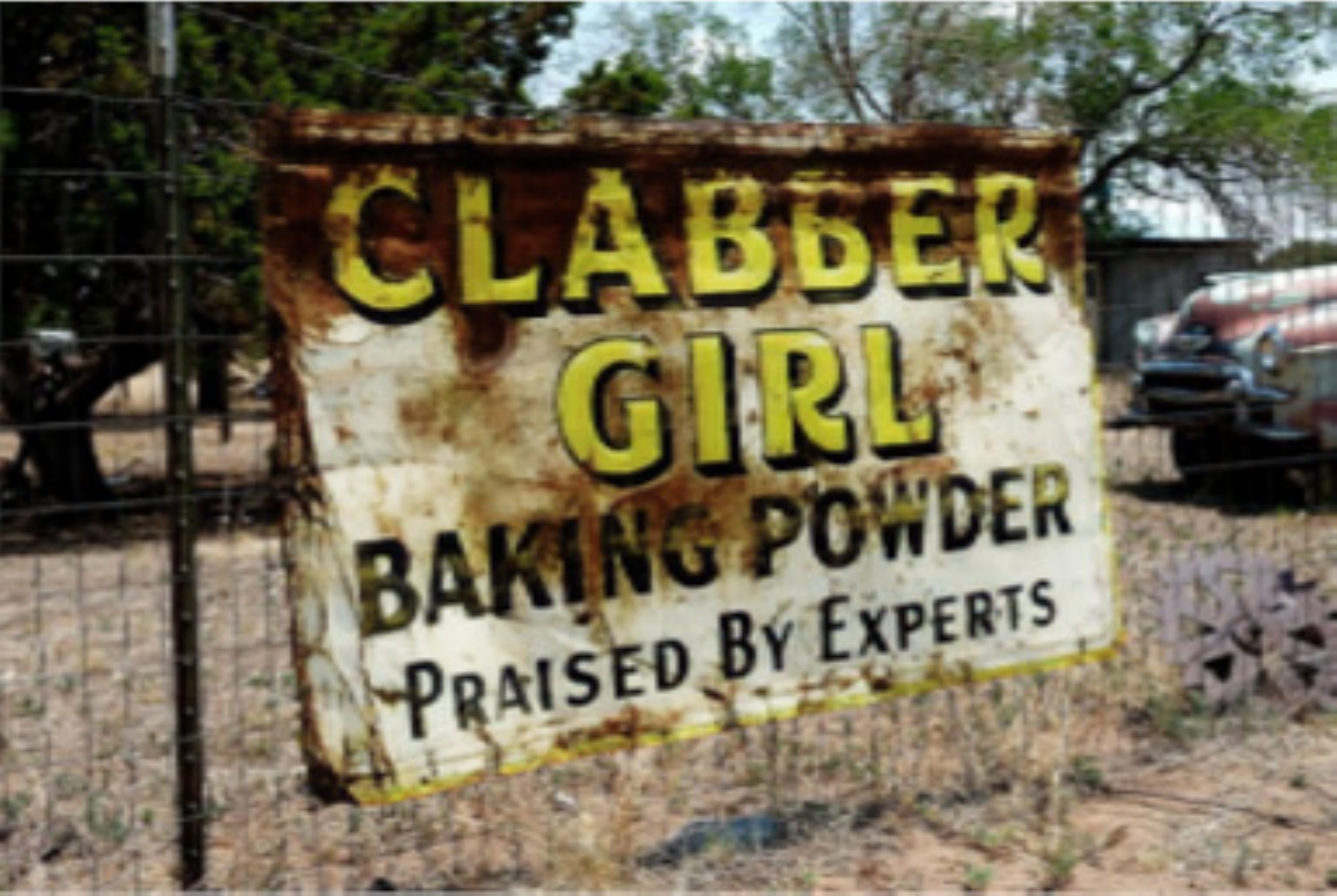}
		\caption*{w/ BAIN}
	\end{subfigure}
	\hfil
	\begin{subfigure}{0.19\linewidth}
		% include first image
		\centering
		\includegraphics[width=\linewidth]{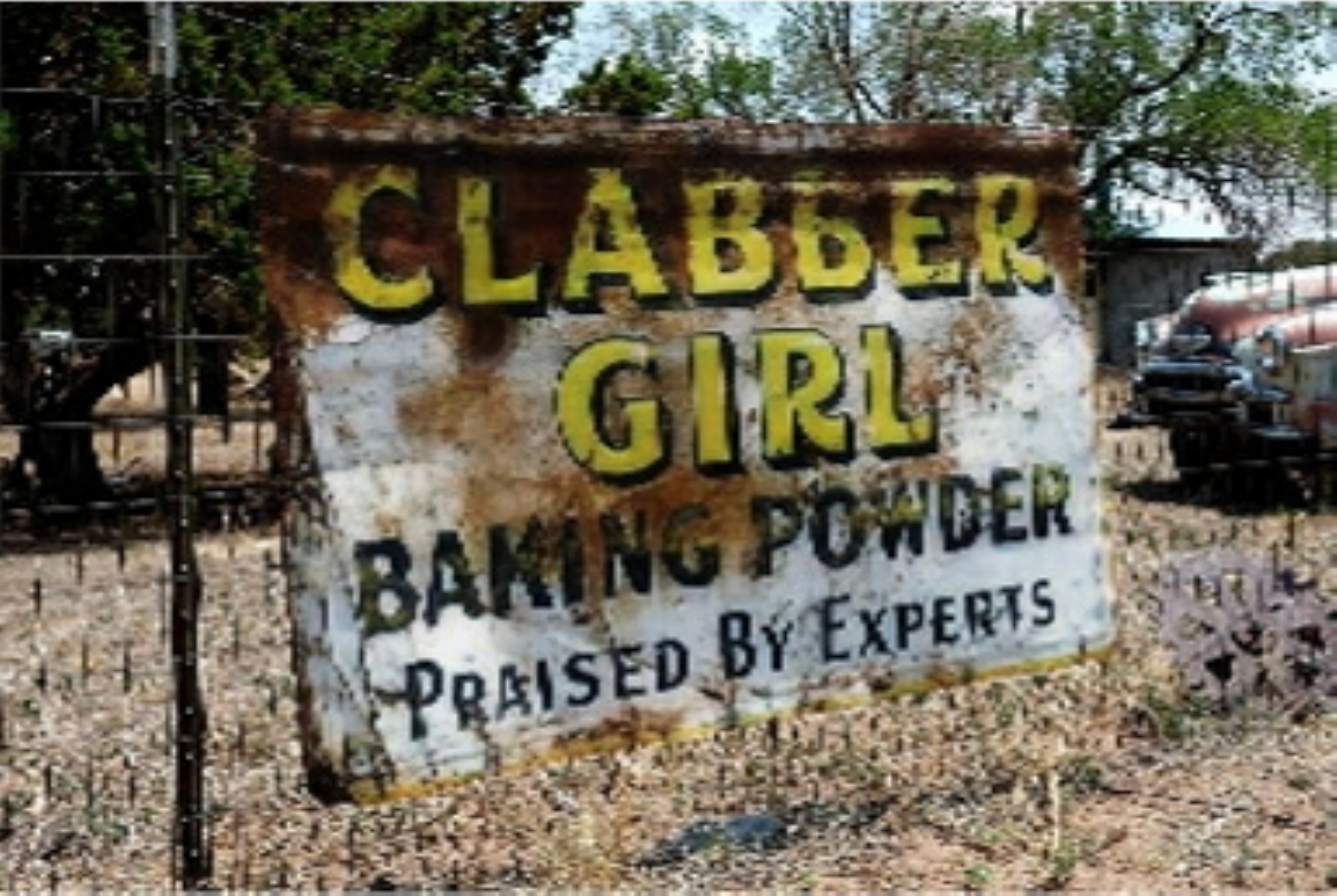}
		\caption*{Full model}
	\end{subfigure}
	\caption{Ablation study on BAIN and SCS-Co. Full model means baseline with both BAIN and SCS-Co.}
	\label{fig:ab}
\end{figure}

\begin{figure}[htbp]
	\centering
	\begin{subfigure}{0.49\linewidth}
		% include first image
		\centering
		\includegraphics[width=.9\linewidth]{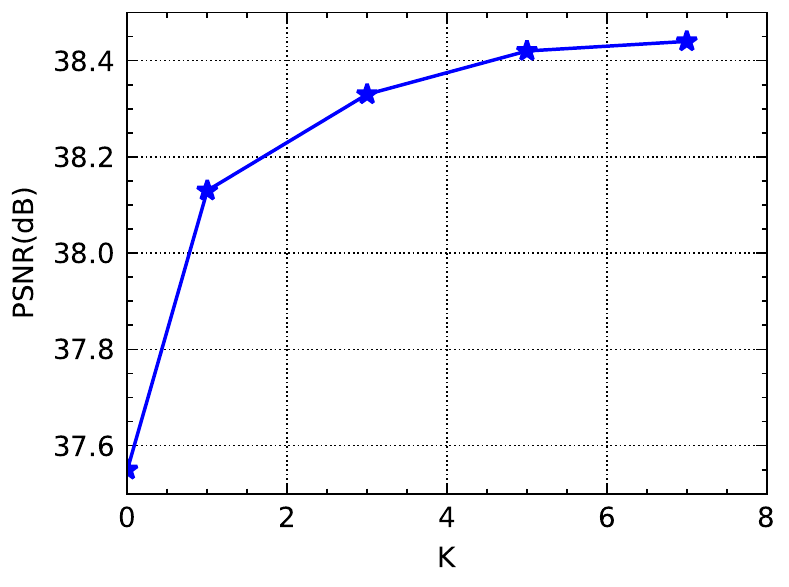}
		\caption{PSNR}
	\end{subfigure}
	\hfil
	\begin{subfigure}{0.49\linewidth}
		% include second image
		\centering
		\includegraphics[width=.9\linewidth]{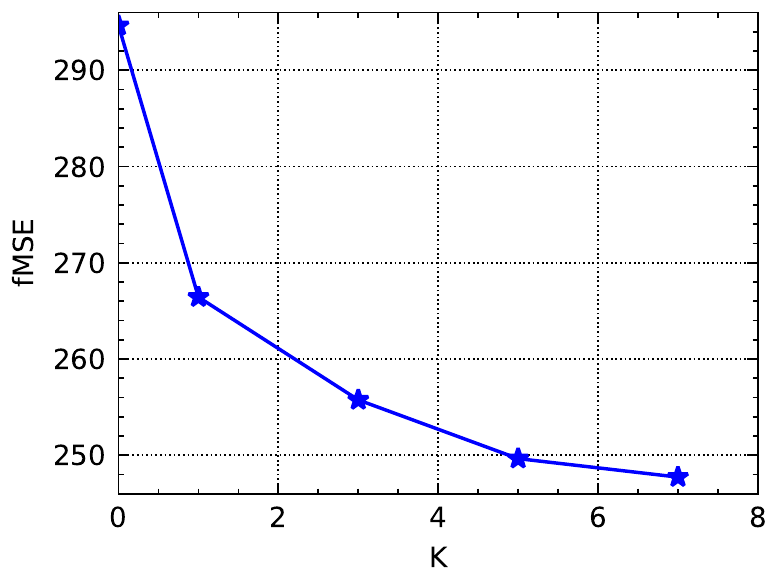}
		\caption{fMSE}
	\end{subfigure}
	\caption{Performance of using different numbers of negative samples in SCS-Co. We report PSNR and fMSE.}
	\label{fig:number-of-negative-samples}
\vspace{-2.0em}
\end{figure}

\paragraph{Effects of BAIN and SCS-Co.}
In Table \ref{tab:ab-BAIN-SCS-CR}, we find that when BAIN is added, the PSNR value is improved from 37.55 dB to 37.84 dB. When SCS-Co is used, the PSNR value is improved from 37.55 dB to 38.42 dB. After adopting both of them, the PSNR value is further improved to 38.75 dB. A similar phenomenon also appears on other metrics. These comparisons demonstrate the effectiveness of our BAIN and SCS-Co, and they cooperate very well to further improve the performance. In addition, to further illustrate the effectiveness of our BAIN and SCS-Co, we show some output results of ablation experiments in Figure \ref{fig:ab}. It can be found that compared with the distortion results produced by the baseline, after adding BAIN, the color and lighting of the output results are close to the real images, but there is still a certain degree of deviation. After the introduction of SCS-Co, the deviation is further corrected, the output results are very close to the real images. More ablation studies on BAIN can be found in supplementary.
\vspace{-1.6em}
\paragraph{SS-CR and CS-CR in SCS-CR.}
SCS-CR is a key component of our SCS-Co and it consists of SS-CR and CS-CR. Therefore, we investigate SS-CR and CS-CR in SCS-CR. As shown in Table \ref{tab:ab-SS-CR-CS-CR}, we find that both SS-CR and CS-CR significantly improve the performance of our model, and the best result is achieved by using them all. The combination of them can strictly regularize the harmonized image in the style representation space, which significantly facilitates the generation of photorealistic visual results.
\vspace{-1.6em}
\paragraph{Number of Negative Samples.}
We further study the effect of the number of negative samples. As shown in Figure \ref{fig:number-of-negative-samples}, adding more negative samples achieves better performance, because the more negative samples, the more powerful constraints can be performed. However, in Figure \ref{fig:number-of-negative-samples}, we also observe that as the number of negative samples increases, the gain brought by adding negative samples decreases. Besides, it takes longer training time when increasing the number of negative samples. Therefore, for the performance-efficiency trade-off, we finally choose to use five negative samples, \ie, we set $K=5$.
\begin{table}[htbp]
	\centering
	\caption{Performance of the baseline with BAIN and/or SCS-Co. The network with both BAIN and SCS-Co performs best.}
	\begin{tabular}{ccccc}
		\toprule
		BAIN  & SCS-Co & PSNR$\uparrow$  & MSE$\downarrow$   & fMSE$\downarrow$ \\
		\midrule
		\XSolidBrush&  \XSolidBrush     & 37.55  & 27.81  & 294.64 \\
		\Checkmark     &    \XSolidBrush   & 37.84 & 25.23 & 269.05 \\
		\XSolidBrush& \Checkmark     & 38.42 & 22.98 & 249.65 \\
		\Checkmark     & \Checkmark     & 38.75 & 21.33  & 248.86 \\
		\bottomrule
	\end{tabular}%
	\label{tab:ab-BAIN-SCS-CR}%
\end{table}%

\begin{table}[htbp]
	\centering
	\caption{Ablation Study on SS-CR and CS-CR in SCS-CR.}
	\begin{tabular}{ccccc}
		\toprule
		SS-CR  & CS-CR & PSNR$\uparrow$  & MSE$\downarrow$   & fMSE$\downarrow$ \\
		\midrule
		\XSolidBrush&  \XSolidBrush     & 37.55  & 27.81  & 294.64 \\
		\Checkmark     &    \XSolidBrush   & 38.03 & 24.09 & 258.64 \\
		\XSolidBrush& \Checkmark     & 37.88 & 25.06 & 269.79 \\
		\Checkmark     & \Checkmark     & 38.42 & 22.98 & 249.65 \\
		\bottomrule
	\end{tabular}%
	\label{tab:ab-SS-CR-CS-CR}%
\end{table}%

\vspace{-1.0em}
\paragraph{Comparison of SCS-Co and Triplet Loss.}
In section \ref{cr-discussion}, we discuss the difference between our SCS-Co and the triplet loss in \cite{cong2021bargainnet}. To further prove the effectiveness of our SCS-Co, we add the triplet loss to the baseline network and compare its result with our SCS-Co. In Table \ref{tab:compare-triplet}, we can find that compared with using the triplet loss, using our SCS-Co brings much more performance improvement, increasing 0.48 dB in PSNR. A similar phenomenon also appears on other metrics. Moreover, we set $K=1$, \ie, we only use the input composite image as the negative sample, which is consistent with the triplet loss. As shown in Table \ref{tab:compare-triplet}, our SCS-Co ($K=1$) still obtains obvious improvement over the triplet loss. It proves that the improvement of our SCS-Co is not only by introducing more dynamically generated negative samples, but also by using a contrastive learning framework and constraining from the foreground self-style and foreground-background style consistency.
\begin{table}[t]
	\centering
	\caption{Comparison of SCS-Co and the triplet loss \cite{cong2021bargainnet}.}
	\resizebox{\linewidth}{!}{
		\begin{tabular}{rcccc}
			\toprule
			Method & Baseline & w/ Triplet Loss & w/ SCS-Co & w/ SCS-Co ($K=1$)  \\
			\midrule
			PSNR$\uparrow$  &   37.55    &    37.94   &  38.42 &   38.13\\
			MSE$\downarrow$  &   27.81    &    25.48   &  22.98 &  23.32\\
			fMSE$\downarrow$  &   294.64    &   274.65    & 249.65 & 266.43 \\
			\bottomrule
	\end{tabular}}%
	\label{tab:compare-triplet}%
\end{table}%
\begin{table}[t]
	\centering
	\caption{Results of integrating SCS-Co into SOTA methods.}
	\resizebox{\linewidth}{!}{
		\begin{tabular}{rccc}
			\toprule
			Method & RainNet \cite{ling2021region} & DIH \cite{tsai2017deep}  & S$^2$AM \cite{cun2020improving}\\
			\midrule
			PSNR$\uparrow$  &   37.07(\textcolor[rgb]{ 1,  0,  0}{$\uparrow$}0.95)    &    34.09(\textcolor[rgb]{ 1,  0,  0}{$\uparrow$}0.68)   & 35.13(\textcolor[rgb]{ 1,  0,  0}{$\uparrow$}0.78) \\
			MSE$\downarrow$  &   34.92(\textcolor[rgb]{ 1,  0,  0}{$\downarrow$}5.37)    &    74.72(\textcolor[rgb]{ 1,  0,  0}{$\downarrow$}2.05)   & 53.86(\textcolor[rgb]{ 1,  0,  0}{$\downarrow$}5.81) \\
			fMSE$\downarrow$  &   364.29(\textcolor[rgb]{ 1,  0,  0}{$\downarrow$}105.31)    &   707.16(\textcolor[rgb]{ 1,  0,  0}{$\downarrow$}66.02)    & 538.99(\textcolor[rgb]{ 1,  0,  0}{$\downarrow$}55.68) \\
			\bottomrule
	\end{tabular}}%
	\label{tab:universality-CR}%
	\vspace{-1.1em}
\end{table}%
\vspace{-1.3em}
\paragraph{Universality of SCS-Co.}
To evaluate the universality of our SCS-Co, we integrate it into three SOTA methods: RainNet \cite{ling2021region}, DIH \cite{tsai2017deep} and S$^2$AM \cite{cun2020improving}. As shown in Table \ref{tab:universality-CR}, after integrating SCS-Co, the performance of each method is improved. This proves the universality of our SCS-Co, which can be easily added to different models without any increase in model parameters.
\vspace{-0.3em}
\section{Conclusion}
In this paper, we propose a novel self-consistent style contrastive learning scheme (SCS-Co) with a self-consistent style contrastive regularization (SCS-CR) and a dynamic negative samples generation strategy. SCS-Co is built upon contrastive learning to ensure that the output harmonized image (anchor sample) is pulled closer to the real image (positive sample) and pushed away from the composite image (the first negative sample) and other dynamically generated negative samples in the style representation space. The constraint is jointly from two aspects of the foreground self-style and foreground-background style consistency. As a result, our SCS-Co can learn more distortion knowledge and reduce the solution space well. Moreover, we propose a background-attentional adaptive instance normalization (BAIN) to pay more attention to those areas in the background that feature-similar to the foreground, and the attention-weighted background feature distribution is calculated to locally align the foreground feature distribution. Experiments demonstrate that our method is superior to other SOTA methods on both synthetic and real datasets.
\vspace{-2.5em}
\paragraph{Acknowledgements.}
This work was partly supported by the Natural Science Foundation of
Guangdong Province (No.2020A1515010711), the Natural Science
Foundation of China (No.61771276) and the Special Foundation
for the Development of Strategic Emerging Industries of Shenzhen
(Nos.JCYJ20200109143010272 and CJGJZD20210408092804011). It is also partly supported by Overseas Cooperative Foundation.

%\vspace{-2.3em}
%%%%%%%%% REFERENCES
{\small
\bibliographystyle{ieee_fullname}
\bibliography{AAAI22}
}

\end{document}